\documentclass[journal]{IEEEtran}
\usepackage{graphicx}
\ifCLASSINFOpdf
\else
\fi
\begin{document}
\title{A Survey of Deep Learning-based Object Detection}

\author{Licheng~Jiao,~\IEEEmembership{Fellow,~IEEE,}
        Fan~Zhang,~Fang~Liu,~\IEEEmembership{Senior~Member,~IEEE,}
        Shuyuan~Yang,~\IEEEmembership{Senior~Member,~IEEE,}
        Lingling~Li,~\IEEEmembership{Member,~IEEE,}
        Zhixi~Feng,~\IEEEmembership{Member,~IEEE,}
        and~Rong~Qu,~\IEEEmembership{Senior~Member,~IEEE}% <-this % stops a space
\thanks{Key Laboratory of Intelligent Perception and Image Understanding of Ministry of Education, International Research Center for Intelligent Perception and Computation, Joint International Research Laboratory of Intelligent Perception and Computation, School of Artificial Intelligence, Xidian University, Xian, Shaanxi Province 710071, China e-mail: (lchjiao@mail.xidian.edu.cn).}}% <-this % stops a space
\maketitle

\begin{abstract}
Object detection is one of the most important and challenging branches of computer vision, which has been widely applied in people’s life, such as monitoring security, autonomous driving and so on, with the purpose of locating instances of semantic objects of a certain class. With the rapid development of deep learning networks for detection tasks, the performance of object detectors has been greatly improved. In order to understand the main development status of object detection pipeline, thoroughly and deeply, in this survey, we first analyze the methods of existing typical detection models and describe the benchmark datasets. Afterwards and primarily, we provide a comprehensive overview of a variety of object detection methods in a systematic manner, covering the one-stage and two-stage detectors. Moreover, we list the traditional and new applications. Some representative branches of object detection are analyzed as well. Finally, we discuss the architecture of exploiting these object detection methods to build an effective and efficient system and point out a set of development trends to better follow the state-of-the-art algorithms and further research. 
\end{abstract}


% \fi
\IEEEpeerreviewmaketitle

\section{Introduction}
\IEEEPARstart{O}{bject} detection has been attracting increasing amounts of attention in recent years due to its wide range of applications and recent technological breakthroughs. This task is under extensive investigation in both academia and real world applications, such as monitoring security, autonomous driving, transportation surveillance, drone scene analysis, and robotic vision. Among many factors and efforts that lead to the fast evolution of object detection techniques, notable contributions should be attributed to the development of deep convolution neural networks and GPUs computing power. At present, deep learning model has been widely adopted in the whole field of computer vision, including general object detection and domain-specific object detection. Most of the state-of-the-art object detectors utilize deep learning networks as their backbone and detection network to extract features from input images (or videos), classification and localization respectively.
Object detection is a computer technology related to computer vision and image processing which deals with detecting instances of semantic objects of a certain class (such as humans, buildings, or cars) in digital images and videos. Well-researched domains of object detection include multi-categories detection, edge detection, salient object detection, pose detection, scene text detection, face detection, and pedestrian detection etc. As an important part of scene understanding, object detection has been widely used in many fields of modern life, such as security field, military field, transportation field, medical field and life field. Furthermore, many benchmarks have played an important role in object detection field so far, such as Caltech \cite{1}, KITTI \cite{2}, ImageNet \cite{3Russakovsky2015}, PASCAL VOC \cite{4Everingham2010}, MS COCO \cite{5_10.1007/978-3-319-10602-1_48}, and Open Images V5 \cite{OpenImages}. In ECCV VisDrone 2018 contest, organizers have released a novel drone platform-based dataset \cite{11visdrone_zhu2018vision} which contains a large amount of images and videos.

$\bullet$ \textbf{Two kinds of object detectors}

Pre-existing domain-specific image object detectors usually can be divided into two categories, the one is two-stage detector, the most representative one, Faster R-CNN \cite{6faster_rcnn}. The other is one-stage detector, such as YOLO \cite{7yolo}, SSD \cite{8ssd}. Two-stage detectors have high localization and object recognition accuracy, whereas the one-stage detectors achieve high inference speed. The two stages of two-stage detectors can be divided by RoI (Region of Interest) pooling layer. For instance, in Faster R-CNN, the first stage, called RPN, a Region Proposal Network, proposes candidate object bounding boxes. The second stage, features are extracted by RoIPool (RoI Pooling) operation from each candidate box for the following classification and bounding-box regression tasks \cite{9mask_rcnn}. Fig.1 (a) shows the basic architecture of two-stage detectors. Furthermore, the one-stage detectors propose predicted boxes from input images directly without region proposal step, thus they are time efficient and can be used for real-time devices. Fig.1 (b) exhibits the basic architecture of one-stage detectors.

$\bullet$ \textbf{Contributions}

This survey focuses on describing and analyzing deep learning based object detection task. The existing surveys always cover a series of domain of general object detection and may not contain state-of-the-art methods which provide some novel solutions and newly directions of these tasks due of the rapid development of computer vision research. 

(1) This paper lists very novel solutions proposed recently but neglects to discuss the basics so that readers can see the cutting edge of the field more easily. 

(2) Moreover, different from previous object detection surveys, this paper systematically and comprehensively reviews deep learning based object detection methods and most importantly the up to date detection solutions and a set of significant research trends as well. 

(3) This survey is featured by in-depth analysis and discussion in various aspects, many of which, to the best of our knowledge, are the first time in this field. 

Above all, it is our intention to provide an overview how different deep learning methods are used rather than a full summary of all related papers. To get into this field, we recommend readers refer to \cite{khan2019survey} \cite{DBLP:journals/corr/abs-1905-05055} \cite{liu2018deep} for more details of early methods.

The rest of this paper is organized as follows. Object detectors need a powerful backbone network to extract rich features. This paper discusses backbone networks in section 2 below. As is known to all, the typical pipelines of domain-specific image detectors act as basics and milestone of the task. In section 3, this paper elaborates the most representative and pioneering deep learning-based approaches proposed before June 2019. Section 4 describes common used datasets and metrics. Section 5 systematically explains the analysis of general object detection methods. Section 6 details five typical fields and several popular branches of object detection. The development trend is summarized in section 7.
\begin{figure*}[!t]
\centering
\includegraphics[width=8in]{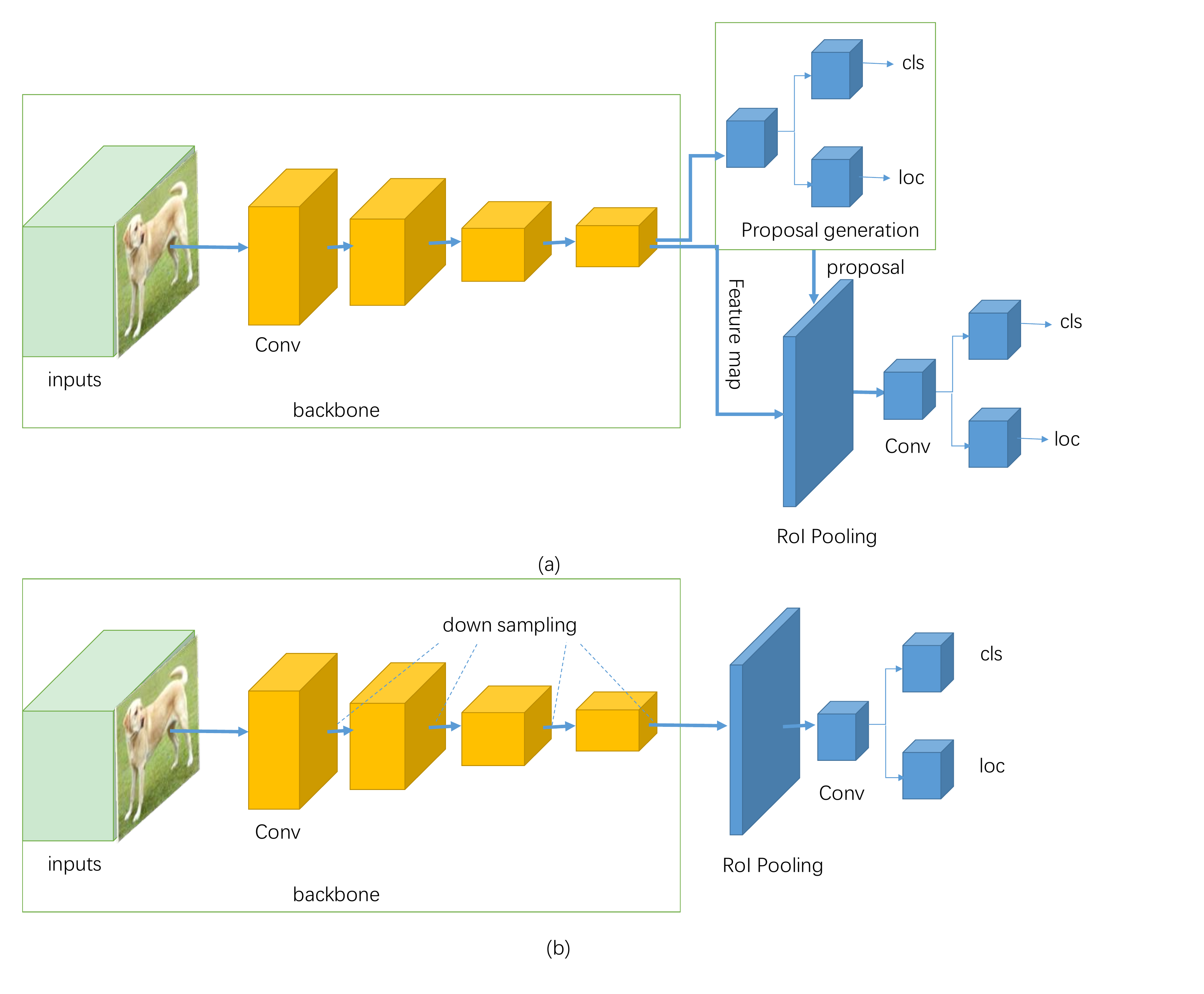}
% where an .eps filename suffix will be assumed under latex,
% and a .pdf suffix will be assumed for pdflatex; or what has been declared
% via \DeclareGraphicsExtensions.
\caption{(a) exhibits the basic architecture of two-stage detectors, which consists of region proposal network to feed region proposals into classifier and regressor. (b) shows the basic architecture of one-stage detectors, which predicts bounding boxes from input images directly. Yellow cubes are a series of convolutional layers (called a block) with the same resolution in backbone network, because of down-sampling operation after one block, the size of the following cubes gradually becoming small. Thick blue cubes are a series of convolutional layers contain one or more convolutional layers. The flat blue cube demonstrates the RoI pooling layer which generates feature maps for objects of the same size.
}
\label{fig_sim}
\end{figure*}
\section{Backbone networks}

Backbone network is acting as the basic feature extractor for object detection task which takes images as input and outputs feature maps of the corresponding input image. Most of backbone networks for detection are the network for classification task taking out the last fully connected layers. The improved version of basic classification network is also available. For instance, Lin et al. \cite{14fpn} add or subtract layers or replace some layers with special designed layers. To better meet specific requirements, some works \cite{7yolo} \cite{li2018detnet} utilize the newly designed backbone for feature extraction. 

%\IEEEpubidadjcol

Towards different requirements about accuracy vs. efficiency, people can choose deeper and densely connected backbones, like ResNet \cite{9mask_rcnn}, ResNeXt \cite{xie2017aggregated}, AmoebaNet \cite{ghiasi2019fpn} or lightweight backbones like MobileNet \cite{howard2017mobilenets}, ShuffleNet \cite{zhang2018shufflenet}, SqueezeNet \cite{iandola2016squeezenet}, Xception \cite{chollet2017xception}, MobileNetV2 \cite{sandler2018mobilenetv2}. When applied to mobile devices, lightweight backbones can meet the requirements. Wang et al. \cite{wang2018pelee} propose a novel real-time object detection system by combining PeleeNet with SSD \cite{8ssd} and optimizing the architecture for fast processing speed. In order to meet the needs of high precision and more accurate applications, complex backbones are needed. On the other hand, real-time acquirements like video or webcam require not only high processing speed but high accuracy \cite{7yolo}, which need well-designed backbone to adapt to the detection architecture and make a trade-off between speed and accuracy.

To explore more competitive detecting accuracy, deeper and densely connected backbone is adopted to replace the shallower and sparse connected counterpart. He et al. \cite{9mask_rcnn} utilize ResNet \cite{13resnet} rather than VGG \cite{19vgg_simonyan2014very} to capture rich features which is adopted in Faster R-CNN \cite{6faster_rcnn} for further accuracy gain because of its high capacity.

The newly high performance classification networks can improve precision and reduce the complexity of object detection task. This is an effective way to further improve network performance because the backbone network acts as a feature extractor. As is known to all, the quality of features determines the upper bound of network performance, thus it is an important step that needs further exploration. Please refer to \cite{rawat2017deep} for more details.

\section{Typical baselines}

With the development of deep learning and the continuous improvement of computing power, great progress has been made in the field of general object detection. When the first CNN-based object detector R-CNN was proposed, a series of significant contributions have been made which promote the development of general object detection by a large margin. We introduce some representative object detection architectures for beginners to get started in this domain.

\subsection{Two-stage Detectors}
\subsubsection{R-CNN}
R-CNN is a region based CNN detector. As Girshick et al. \cite{10rcnn} propose R-CNN which can be used in object detection tasks, their works are the first to show that a CNN could lead to dramatically higher object detection performance on PASCAL VOC datasets \cite{4Everingham2010} than those systems based on simpler HOG-like features. Deep learning method is verified effective and efficient in the field of object detection.

R-CNN detector consists of four modules. The first module generates category-independent region proposals. The second module extracts a fixed-length feature vector from each region proposal. The third module is a set of class-specific linear SVMs to classify the objects in one image. The last module is a bounding-box regressor for precisely bounding-box prediction. For detailed, first, to generate region proposals, the authors adopt selective search method. Then, a CNN is used to extract a 4096-dimensional feature vector from each region proposal. Because the fully connected layer needs input vectors of fixed length, the region proposal features should have the same size. The authors adopt a fixed ${227}\times{227}$ pixel as the input size of CNN. As we know, the objects in various images have different size and aspect ratio, which makes the region proposals extracted by the first module different in size. Regardless of the size or aspect ratio of the candidate region, the authors warp all pixels in a tight bounding box around it to the required size ${227}\times{227}$. The feature extraction network consists of five convolutional layers and two fully connected layers. And all CNN parameters are shared across all categories. Each category trains category-independent SVM which does not share parameters between different SVMs.

Pre-training on larger dataset followed by fine-tuning on the specified dataset is a good training method for deep convolutional neural networks to achieve fast convergence. First, Girshick et al. \cite{10rcnn} pre-train the CNN on a large scale dataset (ImageNet classification dataset \cite{3Russakovsky2015}). The last fully connected layer is replaced by the CNN’s ImageNet specific 1000-way classification layer. The next step is to use SGD (stochastic gradient descent) to fine-tune the CNN parameters on the warped proposal windows. The last fully connected layer is a (N+1)-way classification layer (N: object classes, 1: background) which is randomly initialized. 

When setting positive examples and negative examples the authors divide into two situations. The first is to define the IoU (intersection over union) overlap threshold as 0.5 in the process of fine-tuning. Below the threshold, region proposals are defined as negatives while above it object proposals are defined as positives. As well, the object proposals whose maximum IoU overlap with a ground-truth class are assigned to the ground-truth box. Another situation is to set parameters when training SVM. In contrast, only the ground-truth boxes are taken as positive examples for their respective classes and proposals have less than 0.3 IoU overlap with all ground-truth instances of one class as a negative proposal for that class. These proposals with overlap between 0.5 and 1 and they are not ground truth, which expand the number of positive examples by approximately ${30}\times$. Therefore such a big set can avoid overfitting during fine-tuning process effectively.

\subsubsection{Fast R-CNN}
R-CNN proposed a year later, Ross Girshick \cite{12fast_rcnn} proposed a faster version of R-CNN, called Fast R-CNN \cite{12fast_rcnn}. Because R-CNN performs a ConvNet forward pass for each region proposal without sharing computation, R-CNN takes a long time on SVMs classification. Fast R-CNN extracts features from an entire input image and then passes the region of interest (RoI) pooling layer to get the fixed size features as the input of the following classification and bounding box regression fully connected layers. The features are extracted from the entire image once and are sent to CNN for classification and localization at a time. Compared to R-CNN which inputs each region proposals to CNN, a large amount of time can be saved for CNN processing and large disk storage to store a great deal of features can be saved either in Fast R-CNN. As mentioned above, training R-CNN is a multi-stage process which covers pre-training stage, fine-tuning stage, SVMs classification stage and bounding box regression stage. Fast R-CNN is a one-stage end-to-end training process using a multi-task loss on each labeled RoI to jointly train the network.

Another improvement is that Fast R-CNN uses a RoI pooling layer to extract a fixed size feature map from region proposals of different size. This operation with no need of warping regions and reserves the spatial information of features of region proposals. For fast detection, the author uses truncated SVD which accelerates the forward pass of computing the fully connected layers.

Experiment results showed that Fast R-CNN had 66.9\% mAP while R-CNN of 66.0\% on PASCAL VOC 2007 dataset \cite{4Everingham2010}. Training time dropped to 9.5 hours as compared to R-CNN with 84h, 9 times faster. For test rate (s/image), Fast R-CNN with truncated SVD (0.32s) was ${213}\times$ faster than R-CNN (47s). These experiments were carried out on an Nvidia K40 GPU, which demonstrated that Fast R-CNN did accelerate object detection process.
\subsubsection{Faster R-CNN}
Three months after Fast R-CNN was proposed, Faster R-CNN \cite{6faster_rcnn} further improves the region-based CNN baseline. Fast R-CNN uses selective search to propose RoI, which is slow and needs the same running time as the detection network. Faster R-CNN replaces it with a novel RPN (region proposal network) that is a fully convolutional network to efficiently predict region proposals with a wide range of scales and aspect ratios. RPN accelerates the generating speed of region proposals because it shares fully-image convolutional features and a common set of convolutional layers with the detection network. The procedure is simplified in Fig.3 (b). Furthermore, a novel method for different sized object detection is that multi-scale anchors are used as reference. The anchors can greatly simplify the process of generating various sized region proposals with no need of multiple scales of input images or features. On the outputs (feature maps) of the last shared convolutional layer, sliding a fixed size window (${3}\times{3}$), the center point of each feature window is relative to a point of the original input image which is the center point of k (${3}\times{3}$) anchor boxes. The authors define anchor boxes have 3 different scales and 3 aspect ratios. The region proposal is parameterized relative to a reference anchor box. Then they measure the distance between predicted box and its corresponding ground truth box to optimize the location of the predicted box.

Experiments indicated that Faster R-CNN has greatly improved both precision and detection efficiency. On PASCAL VOC 2007 test set, Faster R-CNN achieved mAP of 69.9\% as compared to Fast R-CNN of 66.9\% with shared convolutional computations. As well, total running time of Faster R-CNN (198ms) was nearly 10 times lower than Fast R-CNN (1830ms) with the same VGG \cite{19vgg_simonyan2014very} backbone, and processing rate was 5fps vs. 0.5fps.

\subsubsection{Mask R-CNN}
Mask R-CNN \cite{9mask_rcnn} is an extending work to Faster R-CNN mainly for instance segmentation task. Regardless of the adding parallel mask branch, Mask R-CNN can be seen a more accurate object detector. He et al. use Faster R-CNN with a ResNet \cite{13resnet}-FPN \cite{14fpn} (feature pyramid network, a backbone extracts RoI features from different levels of the feature pyramid according to their scale) backbone to extract features achieves excellent accuracy and processing speed. FPN contains a bottom-up pathway and a top-down pathway with lateral connections. The bottom-up pathway is a backbone ConvNet which computes a feature hierarchy consisting of feature maps at several scales with a scaling step of 2. The top-down pathway produces higher resolution features by upsampling spatially coarser, but semantically stronger, feature maps from higher pyramid levels. At the beginning, the top pyramid feature maps are captured by the output of the last convolutional layer of the bottom-up pathway. Each lateral connection merges feature maps of the same spatial size from the bottom-up pathway and the top-down pathway. While the dimensions of feature maps are different, the ${1}\times{1}$ convolutional layer can change the dimension. Once undergoing a lateral connection operation, there will form a new pyramid level and predictions are made independently on each level. Because higher-resolution feature maps are important for detecting small objects while lower-resolution feature maps are rich in semantic information, feature pyramid network extracts significant features.

Another way to improve accuracy is to replace RoI pooling with RoIAlign to extract a small feature map from each RoI, as shown in Fig. 2. Traditional RoI pooling quantizes floating-number in two steps to get approximate feature values in each bin. First, quantization is applied to calculate the coordinates of each RoI on feature maps, given the coordinates of RoIs in the input images and down sampling stride. Then RoI feature maps are divided into bins to generate feature maps at the same size, which is also quantized during the process. These two quantization operations cause misalignments between the RoI and the extracted features. To address this, at those two steps, RoIAlign avoids any quantization of the RoI boundaries or bins. First it computes the floating-number of the coordinates of each RoI feature map followed by a bilinear interpolation operation to compute the exact values of the features at four regularly sampled locations in each RoI bin. Then it aggregates the results using max or average pooling to get values of each bin. Fig. 2 is an example of RoIAlign operation. 

Experiments showed that with the above two improvements the precision got promotion. Using ResNet-FPN backbone improved 1.7 points box AP and RoIAlign operation improved 1.1 points box AP on MS COCO detection dataset.

\begin{figure}[!t]
\centering
\includegraphics[width=3.6in]{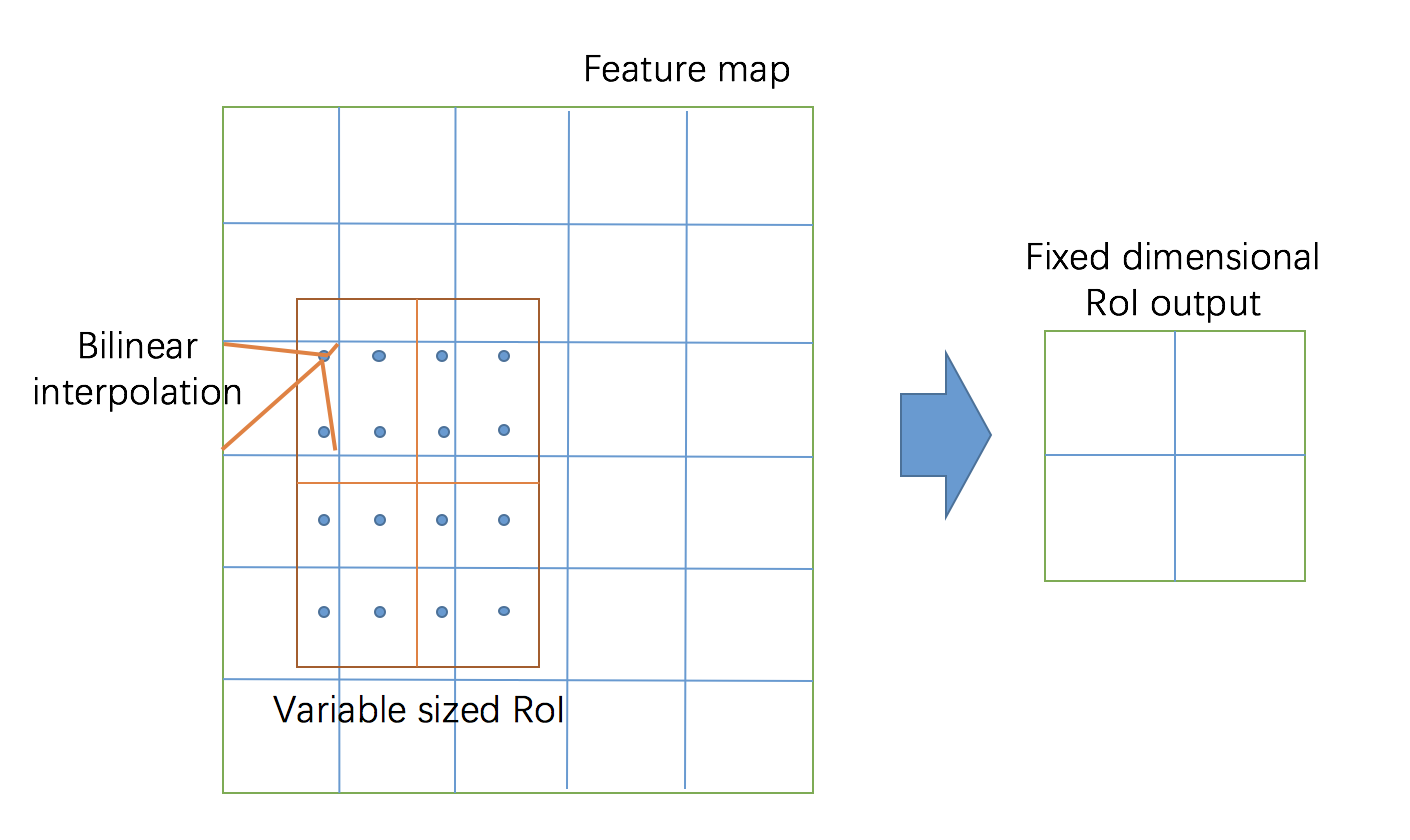}
% where an .eps filename suffix will be assumed under latex,
% and a .pdf suffix will be assumed for pdflatex; or what has been declared
% via \DeclareGraphicsExtensions.
\caption{RoIAlign operation. The first step calculates floating number coordinates of an object in the feature map. Next step utilizes bilinear interpolation to compute the exact values of the features at four regularly sampled locations in the separated bin.}
\label{fig_sim}
\end{figure}

\subsection{One-stage Detectors}
\subsubsection{YOLO}
YOLO \cite{7yolo} (you only look once) is a one-stage object detector proposed by Redmon et al. after Faster R-CNN \cite{6faster_rcnn}. The main contribution is real-time detection of full images and webcam. Firstly, it is due to this pipeline only predicts less than 100 bounding boxes per image while Fast R-CNN using selective search predicts 2000 region proposals per image. Secondly, YOLO frames detection as a regression problem, so a unified architecture can extract features from input images straightly to predict bounding boxes and class probabilities. YOLO network runs at 45 frames per second with no batch processing on a Titan X GPU as compared to Fast R-CNN at 0.5fps and Faster R-CNN at 7fps.

YOLO pipeline first divides the input image into an ${S}\times{S}$ grid, where a grid cell is responsible to detect the object whose center falls into. The confidence score is obtained by multiplying two parts, where $P(object)$ denotes the probability of the box containing an object and IOU (intersection over union) shows how accurate the box containing that object. Each grid cell predicts B bounding boxes $(x, y, w, h)$ and confidence scores for them and C-dimension conditional class probabilities for C categories. The feature extraction network contains 24 convolutional layers followed by 2 fully connected layers. When pre-training on ImageNet dataset, the authors use the first 20 convolutional layers and an average pooling layer followed by a fully connected layer. For detection, the whole network is used for better performance. In order to get fine-grained visual information to improve detection precision, in detection stage double the input resolution of ${224}\times{224}$ in pre-training stage.

The experiments showed that YOLO was not good at accurate localization and localization error was the main component of prediction error. Fast R-CNN makes many background false positives mistakes while YOLO is 3 times less than it. Training and testing on PASCAL VOC dataset, YOLO achieved 63.4\% mAP with 45 fps as compared to Fast R-CNN (70.0\% mAP, 0.5fps) and Faster R-CNN (73.2\% mAP, 7fps).

\subsubsection{YOLOv2}
YOLOv2 \cite{15yolov2} is a second version of YOLO \cite{7yolo}, which adopts a series of design decisions from past works with novel concepts to improve YOLO’s speed and precision.

\textbf{Batch Normalization.} Fixed distribution of inputs to a ConvNet layer would have positive consequences for the layers. It is impractical to normalize the entire training set because the optimization step uses stochastic gradient descent. Since SGD uses mini-batches during training, each mini-batch produces estimates of the mean and variance of each activation. Computing the mean and variance value of the mini-batch of size $m$, then normalize the activations of number $m$ to have mean zero and variance 1. Finally, the elements of each mini-batch are sampled from the same distribution. This operation can be seen as a BN layer \cite{23batchnormalization_ioffe2015batch} which outputs activations with the same distribution. YOLOv2 adds a BN layer ahead of each convolutional layer which accelerates the network to get convergence and helps regularize the model. Batch normalization gets more than 2\% improvement in mAP.

\textbf{High Resolution Classifier.} In YOLO backbone, the classifier adopts an input resolution of ${224}\times{224}$ then increases the resolution to 448 for detection. This process needs the network adjust to a new resolution inputs when switches to object detection task. To address this, YOLOv2 adds a fine-tuning process to the classification network at ${448}\times{448}$ for 10 epochs on ImageNet dataset which increases the mAP at 4\%.

\textbf{Convolutional with Anchor Boxes.} In original YOLO networks, coordinates of predicted boxes are directly generated by fully connected layers. Faster R-CNN uses anchor boxes as reference to generate offsets with predicted boxes. YOLOv2 adopts this prediction mechanism and firstly removes fully connected layers. Then it predicts class and objectness for every anchor box. This operation increases 7\% recall while mAP decreases 0.3\%.

\textbf{Predicting the size and aspect ratio of anchor boxes using dimension clusters.}  In Faster R-CNN, the size and aspect ratio of anchor boxes is identified empirically. For easier learning to predict good detections, YOLOv2 uses K-means clustering on the training set bounding boxes to automatically get good priors. Using dimension clusters along with directly predicting the bounding box center location improves YOLO by almost 5\% over the above version with anchor boxes.

\textbf{Fine-Grained Features.} For localizing smaller objects, high-resolution feature maps can provide useful information. Similar to the identity mappings in ResNet, YOLOv2 concatenates the higher resolution features with the low resolution features by stacking adjacent features into different channels which gives a modest 1\% performance increase.

\textbf{Multi-Scale Training.} For networks to be robust to run on images of different sizes, every 10 batches the network randomly chooses a new image dimension size from $\{320, 352, ..., 608\}$. This means the same network can predict detections at different resolutions. At high resolution detection, YOLOv2 achieves 78.6\% mAP and 40fps as compared to YOLO with 63.4\% mAP and 45fps on VOC 2007.

As well, YOLOv2 proposes a new classification backbone namely Darknet-19 with 19 convolutional layers and 5 max-pooling layers which requires less operations to process an image yet achieves high accuracy. The more competitive YOLOv2 version has 78.6\% mAP and 40fps as compared to Faster R-CNN with ResNet backbone of 76.4\% mAP and 5fps, and SSD500 has 76.8\% mAP and 19fps. As mentioned above, YOLOv2 can achieve high detecting precision while high processing rate which benefit from 7 main improvements and a new backbone.
\subsubsection{YOLOv3}
YOLOv3 \cite{16Redmon2018YOLOv3AI} is an improved version of YOLOv2. First, YOLOv3 uses multi-label classification (independent logistic classifiers) to adapt to more complex datasets containing many overlapping labels. Second, YOLOv3 utilizes three different scale feature maps to predict the bounding box. The last convolutional layer predicts a 3-d tensor encoding class predictions, objectness, and bounding box. Third, YOLOv3 proposes a deeper and robust feature extractor, called Darknet-53, inspired by ResNet.

According to results of experiments on MS COCO dataset, YOLOv3 (AP:33\%) performs on par with the SSD variant (DSSD513:AP:33.2\%) under MS COCO metrics yet 3 times faster than DSSD while quite a bit behind RetinaNet \cite{17retinanet} (AP:40.8\%). But uses the “old” detection metric of mAP at IOU= 0.5 (or ${AP}_{50}$), YOLOv3 can achieve 57.9\% mAP as compared to DSSD513 of 53.3\% and RetinaNet of 61.1\%. Due to the advantages of multi-scale predictions, YOLOv3 can detect small objects even more but has comparatively worse performance on medium and larger sized objects.
\begin{table}[!t]
\renewcommand{\arraystretch}{1.3}
\caption{AP scores (\%) on the MS COCO dataset,${AP}_S$:AP of small objects, ${AP}_M$:AP of medium objects, ${AP}_L$:AP of large objects}
\label{table_example}
\centering
%% Some packages, such as MDW tools, offer better commands for making tables
%% than the plain LaTeX2e tabular which is used here.
\begin{tabular}{|c|c|c|c|}
\hline
Model & ${AP}_S$ & ${AP}_M$ & ${AP}_L$\\
\hline
DSSD513 & 13.0 & 35.4 & 51.1\\
\hline
RetinaNet & 24.1 & 44.2 & 51.2\\
\hline
\end{tabular}
\end{table}

\subsubsection{SSD}
SSD \cite{8ssd}, a single-shot detector for multiple categories within one-stage which directly predicts category scores and box offsets for a fixed set of default bounding boxes of different scales at each location in several feature maps with different scales, as shown in Fig.4 (a). The default bounding boxes have different aspect ratios and scales in each feature map. In different feature maps, the scale of default bounding boxes is computed with regularly space between the highest layer and the lowest layer where each specific feature map learns to be responsive to the particular scale of the objects. For each default box, it predicts both the offsets and the confidences for all object categories. Fig.3 (c) shows the method. At training time, matching these default bounding boxes to ground truth boxes where the matched default boxes as positive examples and the rest as negatives. For the large amount of default boxes are negatives, the authors adopt hard negative mining using the highest confidence loss for each default box then pick the top ones to make the ratio between the negatives and positives at most 3:1. As well, the authors implement data augmentation which is proved an effective way to enhance precision by a large margin.

Experiments showed that SSD512 had a competitive result on both mAP and speed with VGG-16 \cite{19vgg_simonyan2014very} backbone. SSD512 (input image size: ${512}\times{512}$) achieved mAP of 81.6\% on PASCAL VOC 2007 test set and 80.0\% on PASCAL VOC 2012 test set as compared to Faster R-CNN (78.8\%, 75.9\%) and YOLO (VOC2012: 57.9\%). On MS COCO DET dataset, SSD512 was better than Faster R-CNN under all evaluation criteria.

\begin{figure*}[!t]
\centering
\includegraphics[width=7in]{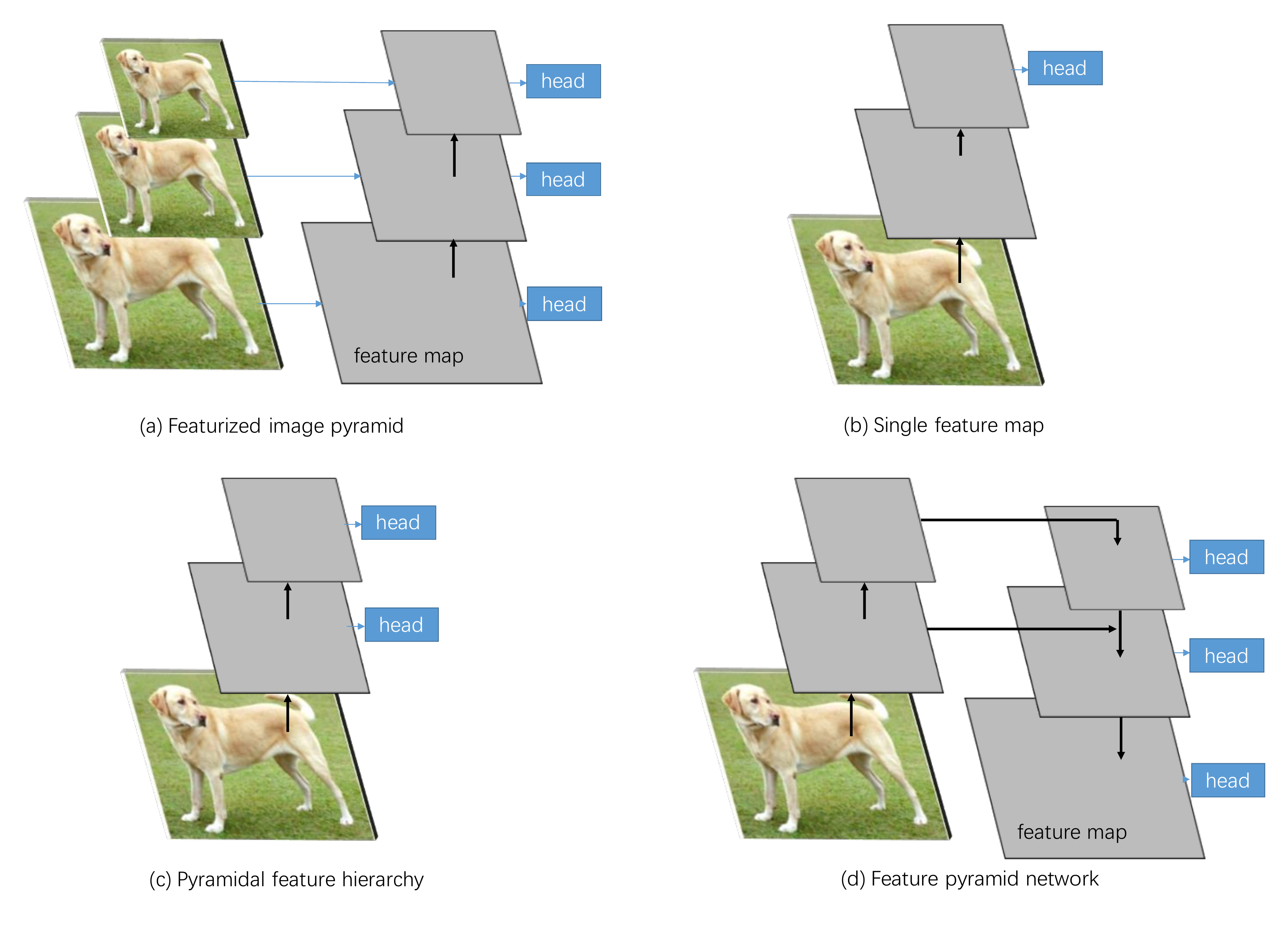}
% where an .eps filename suffix will be assumed under latex,
% and a .pdf suffix will be assumed for pdflatex; or what has been declared
% via \DeclareGraphicsExtensions.
\caption{Four methods utilize features for different sized object prediction. (a) Using an image pyramid to build a feature pyramid. Features are computed on each of the image scales independently, which is slow. (b) Detection systems \cite{6faster_rcnn} \cite{12fast_rcnn} use only single scale features (the outputs of the last convolutional layer) for faster detection. (c) Predicting each of the pyramidal feature hierarchy from a ConvNet as if it is a image pyramid like SSD \cite{8ssd}. (d) Feature Pyramid Network (FPN) \cite{14fpn} is fast like (b) and (c), but more accurate. In this figure, the feature graph is represented by a gray-filled quadrilateral. The head network is represented by a blue rectangle.}
\label{fig_sim}
\end{figure*}

\subsubsection{DSSD}
DSSD \cite{18dssd_fu2017dssd} (Deconvolutional Single Shot Detector) is a modified version of SSD (Single Shot Detector) which adds prediction module and deconvolution module also adopts ResNet-101 as backbone. The architecture of DSSD is shown in Fig.4 (b). For prediction module, Fu et al. add a residual block to each predicting layer, then do element-wise addition of the outputs of prediction layer and residual block. Deconvolution module increases the resolution of feature maps to strengthen features. Each deconvolution layer followed by a prediction module is to predict a variety of objects with different sizes. At training process, first the authors pre-train ResNet-101 based backbone network on the ILSVRC CLS-LOC dataset, then use ${321}\times{321}$ inputs or ${513}\times{513}$ inputs training the original SSD model on detection dataset. Finally, they train the deconvolution module freezing all the weights of SSD module.

Experiments on both PASCAL VOC dataset and MS COCO dataset showed the effectiveness of DSSD513 model, while the added prediction module and deconvolution module brought 2.2\% enhancement on PASCAL VOC 2007 test dataset.

\begin{figure*}[!t]
\centering
\includegraphics[width=7in]{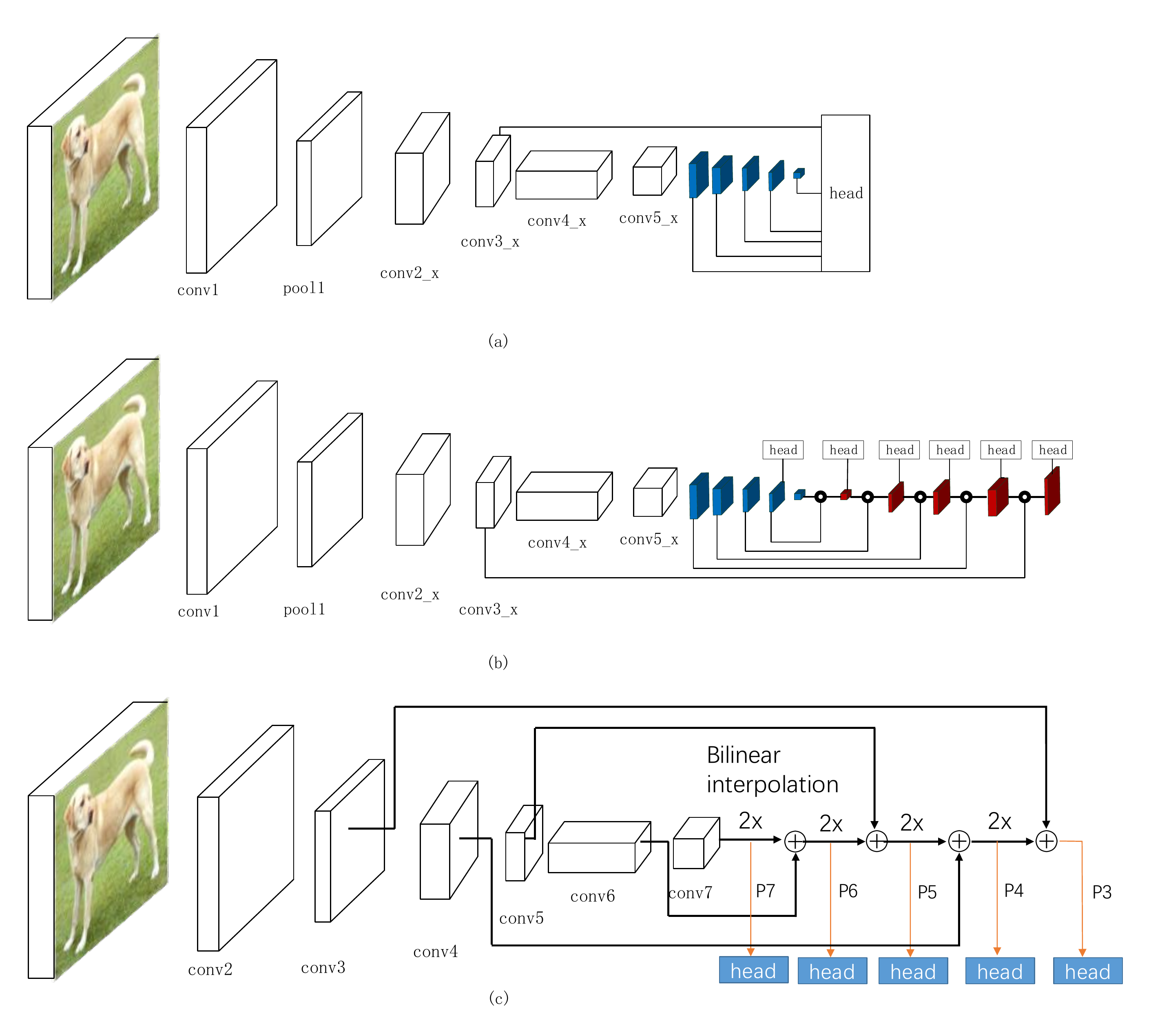}
% where an .eps filename suffix will be assumed under latex,
% and a .pdf suffix will be assumed for pdflatex; or what has been declared
% via \DeclareGraphicsExtensions.
\caption{Networks of SSD, DSSD and RetinaNet on residual network. (a) The blue modules are the layers added in SSD framework whose resolution gradually drop because of down sampling. In SSD the prediction layer is acting on fused features of different levels. Head module consists of a series of convolutional layers followed by several classification layers and localization layers. (b) The red modules are the layers added in DSSD framework denoting deconvolution operation. In DSSD, the prediction layer is following every deconvolution module. (c) RetinaNet utilizes ResNet-FPN as its backbone network, which generates 5 level feature pyramid (P3-P7) corresponding to C3-C7 (the feature map of conv3-conv7 respectively) to predict different sized objects.}
\label{fig_sim}
\end{figure*}

\subsubsection{RetinaNet}
RetinaNet \cite{17retinanet} is a one-stage object detector with focal loss as classification loss function proposed by Lin et al. \cite{17retinanet} in February 2018. The architecture of RetinaNet is shown in Fig.4 (c). R-CNN is a typical two-stage object detector. The first stage generates a sparse set of region proposals and the second stage classifies each candidate location. Owing to the first stage filters out the majority of negative locations, two-stage object detectors can achieve higher precision than one-stage detectors which propose a dense set of candidate locations. The main reason is the extreme foreground-background class imbalance when one-stage detectors train networks to get convergence. So the authors propose a loss function, called focal loss, which can down-weight the loss assigned to well-classified or easy examples. Focal loss concentrates on the hard training examples and avoids the vast number of easy negative examples overwhelming the detector during training. RetinaNet inherits the fast speed of previous one-stage detectors while greatly overcomes the disadvantage of one-stage detectors difficult to train unbalanced positive and negative examples.

Experiments showed that RetinaNet with ResNet-101-FPN backbone got 39.1\% AP as compared to DSSD513 of 33.2\% AP on MS COCO test-dev dataset. With ResNeXt-101-FPN, it made 40.8\% AP far surpassing DSSD513. RetinaNet improved the detection precision on small and medium objects by a large margin.

\subsubsection{M2Det}
To meet a large variety of scale variation across object instances, Zhao et al.  \cite{zhao2018m2det} propose a multi-level feature pyramid network (MLFPN) constructing more effective feature pyramids. The authors adopt three steps to obtain final enhanced feature pyramids. First, like FPN, multi-level features extracted from multiple layers in the backbone are fused as the base feature. Second, the base feature is fed into a block, composing of alternating joint Thinned U-shape Modules and Feature Fusion Modules, and obtains the decoder layers of TUM as the features for next step. Finally, a feature pyramid containing multi-level features is constructed by integrating the decoder layers of equivalent scale. So far, features with multi-scale and multi-level are prepared. The remaining part is to follow the SSD architecture to obtain bounding box localization and classification results in an end-to-end manner. 

For M2Det is a one-stage detector, it achieves AP of 41.0 at speed of 11.8 FPS with single-scale inference strategy and AP of 44.2 with multi-scale inference strategy utilizing VGG-16 on COCO test-dev set.  It outperforms RetinaNet800 (Res101-FPN as backbone) by 0.9\% with single-scale inference strategy, but is twice slower than RetinaNet800.

\subsubsection{RefineDet}
The whole network of RefineDet \cite{zhang2018single} contains two inter-connected modules, the anchors refinement module and the object detection module. These two modules are connected by a transfer connection block to transfer and enhance features from the former module to better predict objects in the latter module. The training process is in an end-to-end way, conducted by three stages, preprocessing, detection (two inter-connected modules), and NMS.

Classical one-stage detectors such as SSD, YOLO, RetinaNet all use one-step regression method to obtain the final results. The authors find that use two-step cascaded regression method can better predict hard detected objects, especially for small objects and provide more accurate locations of objects.

\subsection{Latest Detectors}
\subsubsection{Relation Networks for Object Detection}
Hu et al. \cite{hu2018relation} propose an adapted attention module for object detection called object relation module which considers the interaction between different targets in an image including their appearance feature and geometry information. This object relation module is added in the head of detector before two fully connected layers to get enhanced features for accurate classification and localization of objects. The relation module not only feeds enhanced features into classifier and regressor, but replaces NMS post-processing step which gains higher accuracy than NMS. By using Faster R-CNN, FPN and DCN as the backbone network on the COCO test-dev dataset, adding the relationship module increases the accuracy by 0.2, 0.6, and 0.2, respectively.
\subsubsection{DCNv2}
For learning to adapt to geometric variation reflected in the effective spatial support region of targets, deformable convolutional networks (DCN) \cite{dai2017} was proposed by Dai et al. Regular ConvNets can only focus on features of fixed square size (according to the kernel), thus the receptive field does not properly cover the entire pixel of a target object to represent it. The deformable ConvNets can produce deformable kernel and the offset from the initial convolution kernel (of fixed size) are learned from the networks. Deformable RoI Pooling can also adapt to part location for objects with different shapes. On COCO test-dev set, DCNv1 achieves significant accuracy improvement, which is almost 4\% higher than three plain ConvNets. The best mean average-precision result under the strict COCO evaluation criteria (mAP @[0.5:0.95] ) is 37.5\%.

Deformable ConvNets v2 \cite{zhu2018deformable} utilizes more deformable convolutional layers than DCNv1 (from only the convolutional layers in the conv5 stage to all the convolutional layers in the conv3-conv5 stages) to replace the regular convolutional layers. All the deformable layers are modulated by a learnable scalar, which obviously enhance the deformable effect and accuracy. The authors adopt feature mimicking to further improve detection accuracy by incorporating a feature mimic loss on the per-RoI features of DCN to be similar to good features extracted from cropped images. DCNv2 achieves 45.3\% mAP under COCO evaluation criteria on the COCO 2017 test-dev set, while DCNv1 with 41.7\% and regular Faster R-CNN with 40.1\% on ResNext-101 backbone. On other strong backbones, DCNv2 surpasses DCNv1 by $3\%-5\%$ mAP and regular Faster R-CNN by $5\%-8\%$.

\subsubsection{NAS-FPN}
In recent days, the authors from Google Brain adopt neural architecture search to find some new feature pyramid architecture, named NAS-FPN \cite{ghiasi2019fpn}, consisting of both top-down and bottom-up connections to fuse features with a variety of different scales. By repeating FPN architecture N times then concatenating them into a large architecture during the search, the high level feature layers pick arbitrary level features for them to imitate. All of the highest accuracy architectures have the connection between high resolution input feature maps and output feature layers, which indicate that it is necessary to generate high resolution features for small targets detection. Stacking more pyramid networks, adding feature dimension, adopting high capacity architecture all increase detection accuracy by a large margin. 

Experiments showed that adopting ResNet-50 as backbone of 256 feature dimension, on the COCO test-dev dataset, the mAP of NAS-FPN exceeded the original FPN by 2.9\%. The superlative configuration of NAS-FPN utilized AmoebaNet as backbone network and stacked 7 FPN of 384 feature dimension, which achieved 48.0\% on COCO test-dev.

In conclusion, the typical baselines enhance accuracy by extracting richer features of objects and adopting multi-level and multi-scale features for different sized object detection. To achieve higher speed and precision, the one-stage detectors utilize newly designed loss function to filter out easy samples which drops the number of proposal targets by a large margin. To address geometric variation, adopting deformable convolution layers is an effective way. Modeling the relationship between different objects in an image is also necessary to improve performance. Detection results on MS COCO test-dev dataset of the above typical baselines are listed on table 2.

\begin{table*}[!t]
\renewcommand{\arraystretch}{1.3}
\caption{ Detection results on the MS COCO test-dev dataset of some typical baselines. AP, ${AP}_{50}$ , ${AP}_{75}$ scores (\%). ${AP}_S$:AP of small objects, ${AP}_M$:AP of medium objects, ${AP}_L$:AP of large objects. *DCNv2+Faster R-CNN models are trained on the 118k images of the COCO 2017 train set.}
\label{table_example}
\centering
%% Some packages, such as MDW tools, offer better commands for making tables
%% than the plain LaTeX2e tabular which is used here.
\begin{tabular}{|c|c|c|c|c|c|c|c|c|}
\hline
Method & Data & Backbone & AP & ${AP}_{50}$ & ${AP}_{75}$ & ${AP}_S$ & ${AP}_M$ & ${AP}_L$\\
\hline
Fast R-CNN\cite{12fast_rcnn} & train & VGG-16 & 19.7 & 35.9 & ${-}$ & ${-}$ & ${-}$ & ${-}$ \\
\hline
Faster R-CNN\cite{6faster_rcnn} & trainval &VGG-16 &21.9 & 42.7 & ${-}$&${-}$&${-}$&${-}$\\
\hline
OHEM\cite{shrivastava2016training} & trainval& VGG-16 &22.6 & 42.5 & 22.2& 5.0&23.7&37.9\\
\hline
ION\cite{bell2016inside} & train & VGG-16 & 23.6 & 43.2& 23.6&6.4&24.1&38.3\\
\hline
OHEM++\cite{shrivastava2016training} & trainval &VGG-16 & 25.5 & 45.9 & 26.1&7.4&27.7&40.3 \\
\hline
R-FCN\cite{22rfcn_dai2016r} & trainval & ResNet-101& 29.9 & 51.9 & -&10.8&32.8&45.0\\
\hline
CoupleNet\cite{zhu2017couplenet} & trainval &ResNet-101&34.4 & 54.8 & 37.2&13.4&38.1&52.0\\
\hline
Faster R-CNN G-RMI\cite{huang2017speed} &  ${-}$ &Inception-ResNet-v2 &34.7 & 55.5 & 36.7&13.5&38.1&52.0\\
\hline
Faster R-CNN+++\cite{13resnet} & trainval &ResNet-101-C4&34.9 & 55.7& 37.4&15.6&38.7&50.9\\
\hline
Faster R-CNN w FPN\cite{14fpn} & trainval35k &ResNet-101-FPN&36.2 & 59.1 & 39.0&18.2&39.0&48.2\\
\hline
Faster R-CNN w TDM\cite{shrivastava2016beyond} & trainval &Inception-ResNet-v2-TDM&36.8 & 57.7 & 39.2&16.2&39.8&52.1\\
\hline
Deformable R-FCN\cite{dai2017} & trainval &Aligned-Inception-ResNet&37.5 & 58.0 & 40.8& 19.4&40.1&52.5\\
\hline
${{umd}_{-}}$det\cite{bodla2017soft} & trainval &ResNet-101&40.8 & 62.4 & 44.9 & 23.0&43.4&53.2\\
\hline
Cascade R-CNN\cite{cai2018cascade} & trainval35k &ResNet-101-FPN&42.8 & 62.1 & 46.3 & 23.7&45.5&55.2\\
\hline
SNIP\cite{singh2018analysis} & trainval35k &DPN-98&45.7 & 67.3 & 51.1 & 29.3&48.8&57.1\\
\hline
Fitness-NMS\cite{tychsen2018improving} & trainval35k &ResNet-101&41.8 & 60.9 & 44.9 & 21.5&45.0&57.5\\
\hline
Mask R-CNN\cite{9mask_rcnn} & trainval35k &ResNeXt-101&39.8 & 62.3 & 43.4 & 22.1&43.2&51.2\\
\hline
DCNv2+Faster R-CNN\cite{zhu2018deformable} & train118k* &ResNet-101&44.8 & 66.3 & 48.8 & 24.4&48.1&59.6\\
\hline
G-RMI\cite{huang2017speed} & trainval32k &Ensemble of Five Models & 41.6 & 61.9& 45.4&23.9&43.5&54.9\\
\hline
YOLOv2\cite{15yolov2} & trainval35k &DarkNet-53 & 33.0 & 57.9&34.4&18.3&35.4&41.9\\
\hline
YOLOv3\cite{16Redmon2018YOLOv3AI} & trainval35k &DarkNet-19 & 21.6 & 44.0&19.2&5.0&22.4&35.5\\
\hline
${SSD300}^{*}$\cite{8ssd}& trainval35k &VGG-16 & 25.1 & 43.1 & 25.8&6.6&22.4&35.5\\
\hline
RON384+++\cite{kong2017ron} & trainval &VGG-16 & 27.4 & 49.5 & 27.1&${-}$&${-}$&${-}$\\
\hline
SSD321\cite{18dssd_fu2017dssd} & trainval35k &ResNet-101& 28.0& 45.4 & 29.3&6.2&28.3&49.3\\
\hline
DSSD321\cite{18dssd_fu2017dssd}& trainval35k &ResNet-101&28.0 & 46.1 & 29.2&7.4&28.1&47.6\\
\hline
SSD512*\cite{8ssd} & trainval35k &VGG-16 & 28.8 & 48.5& 30.3&10.9&31.8&43.5\\
\hline
SSD513\cite{18dssd_fu2017dssd} & trainval35k &ResNet-101&31.2 & 50.4& 33.3& 10.2&34.5&49.8\\
\hline
DSSD513\cite{18dssd_fu2017dssd} & trainval35k &ResNet-101& 33.2& 53.3 & 35.2&13.0&35.4&51.1\\
\hline
RetinaNet500\cite{17retinanet} & trainval35k &ResNet-101& 34.4& 53.1 & 36.8&14.7&38.5&49.1\\
\hline
RetinaNet800\cite{17retinanet} & trainval35k &ResNet-101-FPN& 39.1& 59.1& 42.3&21.8&42.7&50.2\\
\hline
M2Det512\cite{zhao2018m2det} & trainval35k &VGG-16 & 37.6& 56.6& 40.5&18.4&43.4&51.2\\
\hline
M2Det512\cite{zhao2018m2det} & trainval35k &ResNet-101& 38.8& 59.4& 41.7&20.5&43.9&53.4\\
\hline
M2Det800\cite{zhao2018m2det} & trainval35k &VGG-16 & 41.0& 59.7& 45.0&22.1&46.5&53.8\\
\hline
RefineDet320\cite{zhang2018single} & trainval35k & VGG-16 &29.4 & 49.2 & 31.3&10.0&32.0&44.4\\
\hline
RefineDet512\cite{zhang2018single} & trainval35k & VGG-16 &33.0 & 54.5 & 35.5&16.3&36.3&44.3\\
\hline
RefineDet320\cite{zhang2018single} & trainval35k &ResNet-101& 32.0& 51.4 & 34.2&10.5&34.7&50.4\\
\hline
RefineDet512\cite{zhang2018single} & trainval35k &ResNet-101& 36.4& 57.5 & 39.5&16.6&39.9&51.4 \\
\hline
RefineDet320+\cite{zhang2018single} & trainval35k & VGG-16 &35.2 & 56.1 & 37.7&19.5&37.2&47.0\\
\hline
RefineDet512+\cite{zhang2018single} & trainval35k & VGG-16 &37.6 & 58.7 & 40.8&22.7&40.3&48.3\\
\hline
RefineDet320+\cite{zhang2018single} & trainval35k &ResNet-101& 38.6& 59.9 & 41.7&21.1&41.7&52.3\\
\hline
RefineDet512+\cite{zhang2018single} & trainval35k &ResNet-101& 41.8& 62.9 & 45.7&25.6&45.1&54.1\\
\hline
CornerNet512\cite{law2018cornernet} & trainval35k &Hourglass& 40.5& 57.8 & 45.3&20.8&44.8&56.7\\
\hline
NAS-FPN\cite{ghiasi2019fpn} & trainval35k &RetinaNet & 45.4& - & - & - & - & - \\
\hline
NAS-FPN\cite{ghiasi2019fpn} & trainval35k &AmoebaNet & 48.0& - & - & - & - & - \\

\hline
\end{tabular}
\end{table*}

\section{Datasets and metrics}
Detecting an object has to state that an object belongs to a specified class and locate it in the image. The localization of an object is typically represented by a bounding box as shown in Fig. 5. Using challenging datasets as benchmark is significant in many areas of research, because they are able to draw a standard comparison between different algorithms and set goals for solutions. Early algorithms focused on face detection using various ad hoc datasets. Later, more realistic and challenging face detection datasets were created. Another popular challenge is the detection of pedestrians for which several datasets have been created. The Caltech Pedestrian Dataset \cite{1} contains 350,000 labeled instances with bounding boxes. General object detection datasets like PASCAL VOC \cite{4Everingham2010}, MS COCO \cite{5_10.1007/978-3-319-10602-1_48}, ImageNet-loc \cite{3Russakovsky2015} are the mainstream benchmarks of object detection task. The official metrics are mainly adopted to measure the performance of detectors with corresponding dataset.

\begin{figure*}[!t]
\centering
\includegraphics[width=7in]{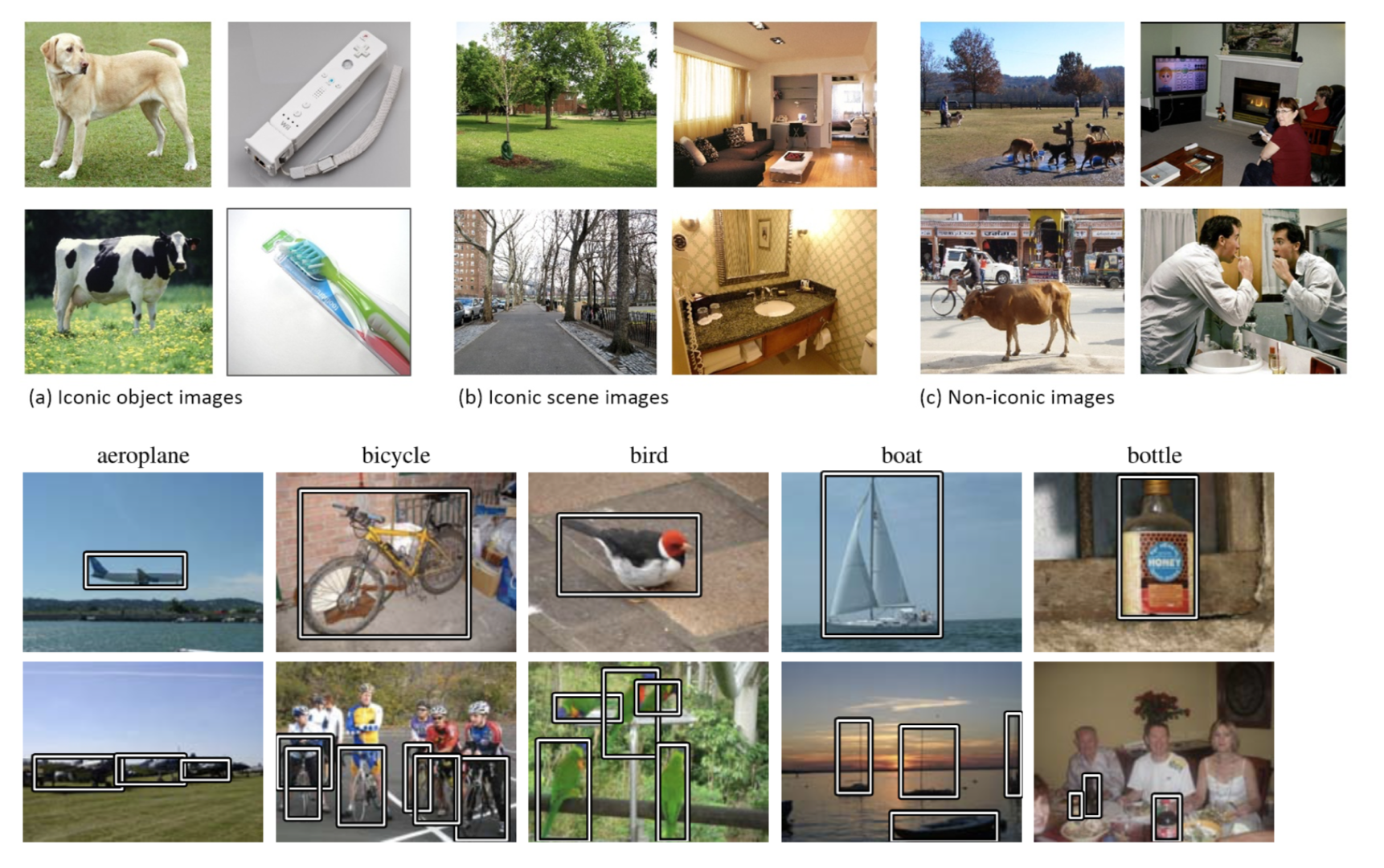}
% where an .eps filename suffix will be assumed under latex,
% and a .pdf suffix will be assumed for pdflatex; or what has been declared
% via \DeclareGraphicsExtensions.
\caption{The first two lines are examples from the MS COCO dataset \cite{5_10.1007/978-3-319-10602-1_48}. The images show three different types of images sampled in the dataset, including iconic objects, iconic scenes and non-iconic objects. In addition, the last two lines are annotated sample images from the PASCAL VOC dataset \cite{4Everingham2010}.}
\label{fig_sim}
\end{figure*}

\begin{figure}[!t]
\centering
\includegraphics[width=2.5in]{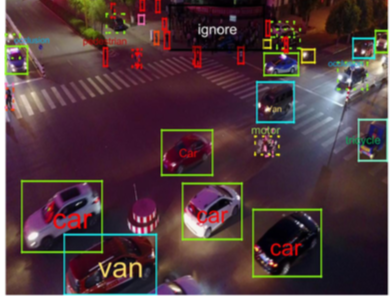}
% where an .eps filename suffix will be assumed under latex,
% and a .pdf suffix will be assumed for pdflatex; or what has been declared
% via \DeclareGraphicsExtensions.
\caption{A drone-based image with bounding box and category labels of objects. Image from VisDrone 2018 dataset \cite{11visdrone_zhu2018vision}.}
\label{fig_sim}
\end{figure}

\subsection{PASCAL VOC dataset}
\subsubsection{Dataset}
For the detection of basic object categories, a multi-year effort from 2005 to 2012 was devoted to the creation and maintenance of a series of benchmark datasets that were widely adopted. The PASCAL VOC datasets \cite{4Everingham2010} contain 20 object categories (in VOC2007, such as person, bicycle, bird, bottle, dog, etc.) spread over 11,000 images. The 20 categories can be considered as 4 main branches-vehicles, animals, household objects and people. Some of them increase semantic specificity of the output, such as car and motorbike, different types of vehicle, but not look similar. In addition, the visually similar classes increase the difficulty of detection, e.g. “dog” vs. “cat”. Over 27,000 object instance bounding boxes are labeled, of which almost 7,000 have detailed segmentations. Imbalanced datasets exist in the VOC2007 dataset, while the class “person” is definitely the biggest one, which is nearly 20 times more than the smallest class “sheep” in the training set. This problem is widespread in the surrounding scene and how can detectors solve this well? Another issue is viewpoint, such as, front, rear, left, right and unspecified, the detectors need to treat different viewpoints separately. Some annotated examples are showed in the last two lines of Fig. 5.
\subsubsection{Metric}
For the VOC2007 criteria, the interpolated average precision (Salton and McGill 1986) was used to evaluate both classification and detection. It is designed to penalize the algorithm for missing object instances, for duplicate detections of one instance, and for false positive detections. 
$$Recall(t)=\frac{\sum_{ij}{1[s_{ij}\geq{t}]}z_{ij}}{N}$$
$$Precision(t)=\frac{\sum_{ij}{1[s_{ij}\geq{t}]}z_{ij}}{\sum_{ij}{1[s_{ij}\geq{t}]}}$$
where $t$ is threshold to judge the IoU between predicted box and ground truth box. In VOC metric, $t$ is set to 0.5. $i$ is the index of the i-th image while $j$ is the index of the j-th object. $N$ is the number of predicted boxes. The indicator function  $1[s_{ij}\geq{t}]=1$ if $s_{ij}\geq{t}$ is true, 0 otherwise. If one detection is matched to a ground truth box according to the threshold criteria, it will be seen as a true positive result.

For a given task and class, the precision/recall curve is computed from a method’s ranked output. Recall is defined as the proportion of all positive examples ranked above a given rank. Precision is the proportion of all examples above that rank which are from the positive class. The mean average precision across all categories is the ultimate results.
\subsection{MS COCO benchmark}
\subsubsection{Dataset}
The Microsoft Common Objects in Context (MS COCO) dataset \cite{5_10.1007/978-3-319-10602-1_48} for detecting and segmenting objects found in everyday life in their natural environments contains 91 common object categories with 82 of them having more than 5,000 labeled instances. These categories cover the 20 categories in PASCAL VOC dataset. In total the dataset has 2,500,000 labeled instances in 328,000 images. MS COCO dataset also pays attention to varied viewpoints and all objects of it are in natural environments which gives us rich contextual information. 

In contrast to the popular ImageNet dataset \cite{3Russakovsky2015}, COCO has fewer categories but more instances per category. The dataset is also significantly larger in the number of instances per category (27k on average) than the PASCAL VOC datasets \cite{4Everingham2010} (about 10 more times less than MS COCO dataset) and ImageNet object detection dataset (1k) \cite{3Russakovsky2015}. MS COCO contains considerably more object instances per image (7.7) as compared to PASCAL VOC (2.3) and ImageNet (3.0). Furthermore, MS COCO dataset contains 3.5 categories per image as compared to PASCAL (1.4) and ImageNet (1.7) on average. In addition, 10\% images in MS COCO have only one category, while in ImageNet and PASCAL VOC all have more than 60\% of images contain a single object category. As we know, small objects need more contextual reasoning to recognize. Images among MS COCO dataset are rich in contextual information. The biggest class is also the “person”, nearly 800,000 instances, while the smallest class is “hair driver”, about 600 instances in the whole dataset. Another small class is “hair brush” whose number is nearly 800. Except for 20 classes with many or few instances, the number of instances in the remaining 71 categories is roughly the same. Three typical categories of images in MS COCO dataset are showed in the first two lines of Fig. 5.
\subsubsection{Metric}
MS COCO metric is under a strict manner and thoroughly judge the performance of detections. The threshold in PASCAL VOC is set to a single value, 0.5, but is belong to [0.5,0.95] with an interval 0.05 that is 10 values to calculate the mean average precision in MS COCO. Apart from that, the special average precision for small, medium and large objects are calculated separately to measure the performance of the detector in detecting targets of different sizes.

\subsection{ImageNet benchmark}
\subsubsection{Dataset}
Challenging datasets can encourage a step forward of vision tasks and practical applications. Another important large-scale benchmark dataset is ImageNet dataset \cite{3Russakovsky2015}. The ILSVRC task of object detection evaluates the ability of an algorithm to name and localize all instances of all target objects present in an image. ILSVRC2014 has 200 object classes and nearly 450k training images, 20k validation images and 40k test images. More comparisons with PASCAL VOC are in Table 3.

\begin{table*}[!t]
%% increase table row spacing, adjust to taste
\renewcommand{\arraystretch}{1.3}
% if using array.sty, it might be a good idea to tweak the value of
% \extrarowheight as needed to properly center the text within the cells
\caption{comparison between ILSVRC object detection dataset and PASCAL VOC dataset}
\label{table_example}
\centering
%% Some packages, such as MDW tools, offer better commands for making tables
%% than the plain LaTeX2e tabular which is used here.
\begin{tabular}{|c|c|c|c|c|c|c|}
\hline
Dataset	&  Classes & Fully annotated training images & Training objects & Val images & Val objects & Annotated obj/im\\
\hline
PASCAL VOC & 20 & 5717 & 13609 & 5823 & 15787 & 2.7\\
\hline
ILSVRC & 200 & 60658 & 478807  & 20121 & 55501 & 2.8\\
\hline
\end{tabular}
\end{table*}

\subsubsection{Metric}
The PASCAL VOC metric uses the threshold t = 0.5. However, for small objects even deviations of a few pixels would be unacceptable according to this threshold. ImageNet uses a loosen threshold calculated as:
$$t=min(0.5,\frac{wh}{(w+10)(h+10)})$$
where $w$ and $h$ are width and height of a ground truth box respectively. This threshold allows for the annotation to extend up to 5 pixels on average in each direction around the object. 
\subsection{VisDrone2018 benchmark}
Last year, a new dataset consists of images and videos captured by drones, called VisDrone2018 \cite{11visdrone_zhu2018vision}, a large-scale visual object detection and tracking benchmark dataset. This dataset aims at advancing visual understanding tasks on the drone platform. The images and video sequences in the benchmark were captured over various urban/suburban areas of 14 different cities across China from north to south. Specifically, VisDrone2018 consists of 263 video clips and 10,209 images (no overlap with video clips) with rich annotations, including object bounding boxes, object categories, occlusion, truncation ratios, etc. This benchmark has more than 2.5 million annotated instances in 179,264 images/video frames. 

Being the larger such dataset ever published, the benchmark enables extensive evaluation and investigation of visual analysis algorithms on the drone platform. VisDrone2018 has a large amount of small objects, such as dense cars, pedestrians and bicycles, which will cause difficult detection about certain categories. Moreover, a large proportion of the images in this dataset have more than 20 objects per image, 82.4\% in training set, and the average number of objects per image is 54 in 6471 images of training set. This dataset contains dark night scenes so the brightness of these images lower than those in day time, which complicates the correct detection of small and dense objects, as shown in Fig. 6. This dataset adopts MS COCO metric.

\subsection{Open Images V5}
\subsubsection{Dataset}
Open Images \cite{OpenImages} is a dataset of 9.2M images annotated with image-level labels, object bounding boxes, object segmentation masks, and visual relationships. Open Images V5 contains a total of 16M bounding boxes for 600 object classes on 1.9M images, which makes it the largest existing dataset with object location annotations. First, the boxes in this dataset have been largely manually drawn by professional annotators (Google-internal annotators) to ensure accuracy and consistency. Second, the images in it are very diverse and mostly contain complex scenes with several objects (8.3 per image on average). Third, this dataset offers visual relationship annotations, indicating pairs of objects in particular relations (e.g. "woman playing guitar", "beer on table"). In total it has 329 relationship triplets with 391,073 samples. Fourth, V5 provides segmentation masks for 2.8M object instances in 350 classes. Segmentation masks mark the outline of objects, which characterizes their spatial extent to a much higher level of detail. Finally, the dataset is annotated with 36.5M image-level labels spanning 19,969 classes.

\subsubsection{Metric}
On the basis of PASCAL VOC 2012 mAP evaluation metric, Kuznetsova et al.  propose several modifications to consider thoroughly of some important aspects of the Open Images Dataset. First, for fair evaluation, the unannotated classes are ignored to avoid wrongly counted as false negatives. Second, if an object belongs to a class and a subclass, an object detection model should give a detection result for each of the relevant classes. The absence of one of these classes would be considered a false negative in that class. Third, in Open Images Dataset, there exists group-of boxes which contain a group of (more than one which are occluding each other or physically touching) object instances but unknown a single object localization inside them. If a detection inside a group-of box and the intersection of the detection and the box divided by the area of the detection is larger than 0.5, the detection will be counted as a true positive. Multiple correct detections inside the same group-of box only count one valid true positive.

\subsection{Pedestrian detection datasets}
Table 4 and table 5 list the comparison between several people detection benchmarks and pedestrian detection datasets, respectively.

\begin{table*}[!t]
%% increase table row spacing, adjust to taste
\renewcommand{\arraystretch}{1.3}
% if using array.sty, it might be a good idea to tweak the value of
% \extrarowheight as needed to properly center the text within the cells
\caption{Comparison of person detection benchmarks,* Images in EuroCity Persons benchmark have day and night collections, which use "/" to split the number of day and night. Table information from Markus Braun et al. IEEE TPAMI2019\cite{8634919}}
\label{table_example}
\centering
%% Some packages, such as MDW tools, offer better commands for making tables
%% than the plain LaTeX2e tabular which is used here.
\begin{tabular}{|c|c|c|c|c|c|c|c|c|}
\hline
Dataset	&  countries & cities & seasons & images & pedestrians  & resolution & weather & train-cal-test-split(\%)\\
\hline
Caltech\cite{1} & 1 & 1 & 1 & 249884 & 289395 & ${640}\times{480}$ & dry & 50-0-50\\
\hline
KITTI\cite{2} & 1 & 1 & 1  & 14999 & 9400 & ${1240}\times{376}$ & dry & 50-0-50\\
\hline
CityPersons\cite{zhang2017citypersons} & 3&27&3&5000&31514&${2048}\times{1024}$ &dry& 60-10-30\\
\hline
TDC\cite{li2016new} & 1&1&1&14674&8919&${2048}\times{1024}$ &dry & 71-8-21\\
\hline 
EuroCity Persons\cite{8634919} & 12&31&4&40217/7118*&183004/35309*&${1920}\times{1024}$ & dry, wet & 60-10-30\\
\hline
\end{tabular}
\end{table*}

\begin{table*}[!t]
%% increase table row spacing, adjust to taste
\renewcommand{\arraystretch}{1.3}
% if using array.sty, it might be a good idea to tweak the value of
% \extrarowheight as needed to properly center the text within the cells
\caption{Comparison of pedestrian detection datasets. The 3rd, 4th, 5th are training set. The 6th, 7th, 8th are test set. Table information from Piotr et al. IEEE TPAMI2012 \cite{1}}
\label{table_example}
\centering
%% Some packages, such as MDW tools, offer better commands for making tables
%% than the plain LaTeX2e tabular which is used here.
\begin{tabular}{|c|c|c|c|c|c|c|c|c|}
\hline
Dataset	&  imaging setup & pedestrians & neg. images & pos. images & pedestrians  & neg. images & pos. images\\
\hline
Caltech\cite{1} & mobile & 192k & 61k & 67k & 155k & 56k & 65k \\
\hline
INRIA\cite{dalal2005histograms} & photo & 1208 & 1218  & 614 & 566 & 453 & 288\\
\hline
ETH\cite{ess2007depth} & mobile &2388& - &499&12k& - &1804\\
\hline
TUD-Brussels\cite{wojek2009multi} & mobile &1776&218&1092&1498& - &508\\
\hline 
Daimler-DB\cite{enzweiler2008monocular} & mobile &192k&61k&67k&155k&56k & 65k\\
\hline
\end{tabular}
\end{table*}

\section{Analysis of general image object detection methods}
Deep neural network based object detection pipelines have four steps in general, image pre-processing, feature extraction, classification and localization, post-processing. Firstly, raw images from the dataset can’t be fed into the network directly. Therefore, we need to resize them to any special sizes and make them clearer, such as enhancing brightness, color, contrast. Data augmentation is also available to meet some requirements, such as flipping, rotation, scaling, cropping, translation, adding Gaussian noise. In addition, GANs \cite{goodfellow2014generative} (generative adversarial networks) can generate new images to enrich the diversity of input according to people's needs. For more details about data augmentation, please refer to \cite{DBLP:journals/corr/abs-1906-11172} for more details. Secondly, feature extraction is a key step for further detection. The feature quality directly determines the upper bound of subsequent tasks like classification and localization. Thirdly, the detector head is responsible to propose and refine bounding box concluding classification scores and bounding box coordinates. Fig. 1 illustrates the basic procedure of the second and the third step. At last, the post-processing step deletes any weak detecting results. For example, NMS is a widely used method in which the highest scoring object deletes its nearby objects with inferior classification scores.

To obtain precise detection results, there exists several methods can be used alone or in combination with other methods. 

\subsection{Enhanced features}
Extracting effective features from input images is a vital prerequisite for further accurate classification and localization steps. To fully utilize the output feature maps of consecutive backbone layers, Lin et al. \cite{14fpn} aim to extract richer features by dividing them into different levels to detect objects of different sizes, as shown in Fig. 3 (d). Some works \cite{9mask_rcnn} \cite{17retinanet} \cite{tian2019fcos} \cite{kong2019foveabox} utilize FPN as their multi-level feature pyramid backbone. Furthermore, a series of improved FPN \cite{ghiasi2019fpn} \cite{zhao2018m2det} \cite{pang2019libra} enriching features for detection task. Kim et al. \cite{kim2018parallel} propose a parallel feature pyramid (FP) network (PFPNet), where the FP is constructed by widening the network width instead of increasing the network depth. The additional feature transformation operation is to generate a pool of feature maps with different sizes, which yields the feature maps with similar levels of semantic abstraction across the scales. Li et al. \cite{li2017fssd} concatenate features from different layers with different scales and then generates new feature pyramid to feed into multibox detectors predicting the final detection results. Chen et al. \cite{chen2017weaving} introduce WeaveNet which iteratively weaves context information from adjacent scales together to enable more sophisticated context reasoning. Zheng et al. \cite{zheng2018extend} extend better context information for the shallow layers of one-stage detector \cite{8ssd}.

Semantic relationships between different objects or regions of an image can help detect occluded and small objects. Bae et al. \cite{bae2019object} utilize the combined and high-level semantic features for object classification and localization which combine the multi-region features stage by stage. Zhang et al. \cite{zhang2018single} combine a semantic segmentation branch and a global activation module to enrich the semantics of object detection features within a typical deep detector. Scene contextual relations \cite{DBLP:journals/corr/abs-1711-05471} can provide some useful information for accurate visual recognition. Liu et al. \cite{liu2018structure} adopt scene contextual information to further improve accuracy. Modeling relations between objects can help object detection. Singh et al. \cite{singh2018sniper} process context regions around the ground-truth object on an appropriate scale. Hu et al. \cite{hu2018relation} propose a relation module that processes a set of objects simultaneously considering both appearance and geometry features through interaction. Mid-level semantic properties of objects can benefit object detection containing visual attributes \cite{8371732}.

Attention mechanism is an effective method for networks focusing on the most significant region part. Some typical works \cite{zhang2019object}\cite{yoo2015attentionnet}\cite{xu2015show}\cite{ba2014multiple}\cite{li2019attention}\cite{hara2017attentional}\cite{DBLP:journals/corr/abs-1904-02874} focus on attention mechanism so as to capture more useful features what detecting objects need. Kong et al. \cite{kong2018deep} design an architecture combining both global attention and local reconfigurations to gather task-oriented features across different spatial locations and scales.

Fully utilizing the effective region of one object can promote the accuracy. Original ConvNets only focus on features of fixed square size (according to the kernel), thus the receptive field does not properly cover the entire pixel of a target object to represent it well. The deformable ConvNets can produce deformable kernel and the offset from the initial convolution kernel (of fixed size) are learned from the networks. Deformable RoI Pooling can also adapt to part location for objects with different shapes. In \cite{dai2017} \cite{zhu2018deformable}, network weights and sampling locations jointly determine the effective support region.

Above all, richer and proper representations of an object can promote detection accuracy remarkably. Brain-inspired mechanism is a powerful way to further improve detection performance.

\subsection{Increasing localization accuracy}
Localization and classification are two missions of object detection. Under object detection evaluation metrics, the precision of localization is a vital measurable indicator, thus increasing localization accuracy can promote detection performance remarkably. Designing a novel loss function to measure the accuracy of predicted boxes is an effective way to increase localization accuracy. Considering intersection over union (IoU) is the most commonly used evaluation metric of object detection, estimating regression quality can judge the IoU between predicted bounding box and its corresponding assignment ground truth box. For two bounding boxes, IoU can be calculated as the intersection area divided by the union area.  $$IoU=\frac{bbox\cap{gt}}{bbox\cup{gt}}$$  A typical work \cite{yu2016unitbox} adopts IoU loss to measure the degree of accuracy the network predicting. This loss function is robust to varied shapes and scales of different objects and can converge well in a short time. Rezatofighi et al. \cite{rezatofighi2019generalized} incorporate generalized IoU as a loss function and a new metric into existing object detection pipeline which makes a consistent improvement than the original smooth L1 loss counterpart. Tychsen et al. \cite{tychsen2018improving} adopt a novel bounding box regression loss for localization branch. IoU loss in this research considers the intersection over union between predicted box and assigned ground truth box which is higher than a preset threshold but not concludes only the highest one. He et al. \cite{he2019bounding} propose a novel bounding box regression loss for learning bounding box localization and transformation variance together. He et al. \cite{he2018softer} introduce a novel bounding box regression loss which has a strong connection to localization accuracy. Pang et al. \cite{pang2019libra} propose a novel balanced L1 Loss to further improve localization accuracy. Cabriel et al. \cite{naturecommunications1} present Axially Localized Detection method to achieve a very high localization precision at the cellular level.

In general, researchers design new loss function of localization branch to make the retained predictions more accurate.

\subsection{Solving negatives-positives imbalance issue}
In the first stage, that networks produce proposals and filter out a large number of negative samples are mainly well designed steps of two-stage detectors. When feed into the detector the proposal bounding boxes belong to a sparse set. However, in a one-stage detector, the network has no steps to filter out bad samples, thus the dense sample sets are difficult to train. The proportion of positive and negative samples is extremely unbalanced as well. The typical solution is hard negative mining \cite{bucher2016hard}. The popularized hard mining methods OHEM \cite{shrivastava2016training} can help drive the focus towards hard samples. Liu et al. \cite{8ssd} adopt hard negative mining method which sorts all of the negative samples using the highest confidence loss for each pre-defined boxes and picking the top ones to make the ratio between the negative and positive samples at most 3:1. Considering hard samples is more effective to improve the detection performance when training an object detector. Pang et al. \cite{pang2019libra} propose a novel hard mining method called IoU-balanced sampling. Yu et al. \cite{yu2018loss} concentrate on real-time requirements.

Another effective way is adding some items in classification loss function. Lin et al. \cite{17retinanet} propose a loss function, called focal loss, which can down-weight the loss assigned to well-classified or easy examples, focusing on the hard training examples and avoiding the vast number of easy negative examples that overwhelm the detector during training. Chen et al. \cite{chen2019towards} consider designing a novel ranking task to replace the conventional classification task and a newly Average-Precision loss for this task, which can alleviate the extreme negative-positive class imbalance issue remarkably.

\subsection{Improving post-processing NMS methods}
Only one detected object can be successfully matched to a ground truth object which will be preserved as a result, while others matched to it are classified as duplicate. NMS (non-maximum suppression) is a heuristic method which selects only the object of the highest classification score, otherwise the object will be ignored. Hu et al. \cite{hu2018relation} use the intermediate results produced by relation module to better determine which object will be saved while it does not need NMS. NMS considers the classification score but the localization confidence is absent, which causes less accurate in deleting weak results. Jiang et al. \cite{jiang2018acquisition} propose IoU-Net learning to predict the IoU between each detected bounding box and the matched ground-truth. Because of its consideration of localization confidence, it improves the NMS method by preserving accurately localized bounding boxes. Tychsen et al. \cite{tychsen2018improving} present a novel fitness NMS method which considers both greater estimated IoU overlap and classification score of predicted bounding boxes. Liu et al. \cite{liu2019adaptive} propose adaptive-NMS which applies a dynamic suppression threshold to an instance decided by the target density. Bodla et al. \cite{bodla2017soft} adopt an improved NMS method without any extra training and is simple to implement. He et al. \cite{he2018softer} further improve soft-NMS method. Jan et al. \cite{hosang2016convnet} feed network score maps resulting from NMS at multiple IoU thresholds. Hosang et al. \cite{hosang2017learning} design a novel ConvNets which does NMS directly without a subsequent post-processing step. Yu et al. \cite{yu2018loss} utilize the final feature map to filter out easy samples so the network concentrates on hard samples.

\subsection{Combining one-stage and two-stage detectors to make good results}
In general, pre-existing object detectors are divided into two categories, the one is two-stage detector, the representative one, Faster R-CNN \cite{6faster_rcnn}. The other is one-stage detector, such as YOLO \cite{7yolo}, SSD \cite{8ssd}. Two-stage detectors have high localization and object recognition precision, while one-stage detectors achieve high inference and test speed. The two stages of two-stage detectors are divided by ROI (Region of Interest) pooling layer. In Faster R-CNN detector, the first stage, called RPN, a Region Proposal Network, proposes candidate object bounding boxes. The second stage, the network extracts features using RoIPool from each candidate box and performs classification and bounding-box regression.

To fully inherit the advantages of one-stage and two-stage detectors while overcoming their disadvantages, Zhang et al. \cite{zhang2018single} present a novel RefineDet which achieves better accuracy than two-stage detectors and maintains comparable efficiency of one-stage detectors.

\subsection{Complicated scene solutions}
Object detection always meets some challenges like small objects hard to detect and heavy occluded situation. Due to low resolution and noisy representation, detecting small objects is a very hard problem. Object detection pipelines \cite{8ssd} \cite{17retinanet} detect small objects through learning representations of objects at multiple scales. Some works \cite{jeong2017enhancement}\cite{xiang2018context}\cite{cao2018feature} improve detection accuracy on the basis of \cite{8ssd}. Li et al. \cite{li2017perceptual} utilize GAN model in which generator transfer perceived poor representations of the small objects to super-resolved ones that are similar enough to real large objects to fool a competing discriminator. This makes the representation of small objects similar to the large one thus improves accuracy without heavy computing cost. Some methods \cite{cai2018cascade}\cite{liu2018learning} improve detection accuracy of small objects by enhancing IoU thresholds to train multiple localization modules. Hu et al. \cite{hu2017finding} adopt feature fusion to better detect small faces which is produced by image pyramid. Xu et al. \cite{xu2018mdssd} fuse high level features with rich semantic information and low level features via Deconvolution Fusion Block to enhance representation of small objects.

Target occlusion is another difficult problem in the field of object detection. Wang et al. \cite{wang2017face} improve the recall of face detection problem in the occluded case without speed decay. Wang et al. \cite{wang2018repulsion} propose a novel bounding box regression loss specifically designed for crowd scenes, called repulsion loss. Zhang et al. \cite{zhang2018occlusion} present a newly designed occlusion-aware R-CNN (OR-CNN) to improve the detection accuracy in the crowd. Baqu et al. \cite{baque2017deep} combine Convolutional Neural Nets and Conditional Random Fields that model potential occlusions.

\begin{figure*}[!t]
\centering
\includegraphics[width=7in]{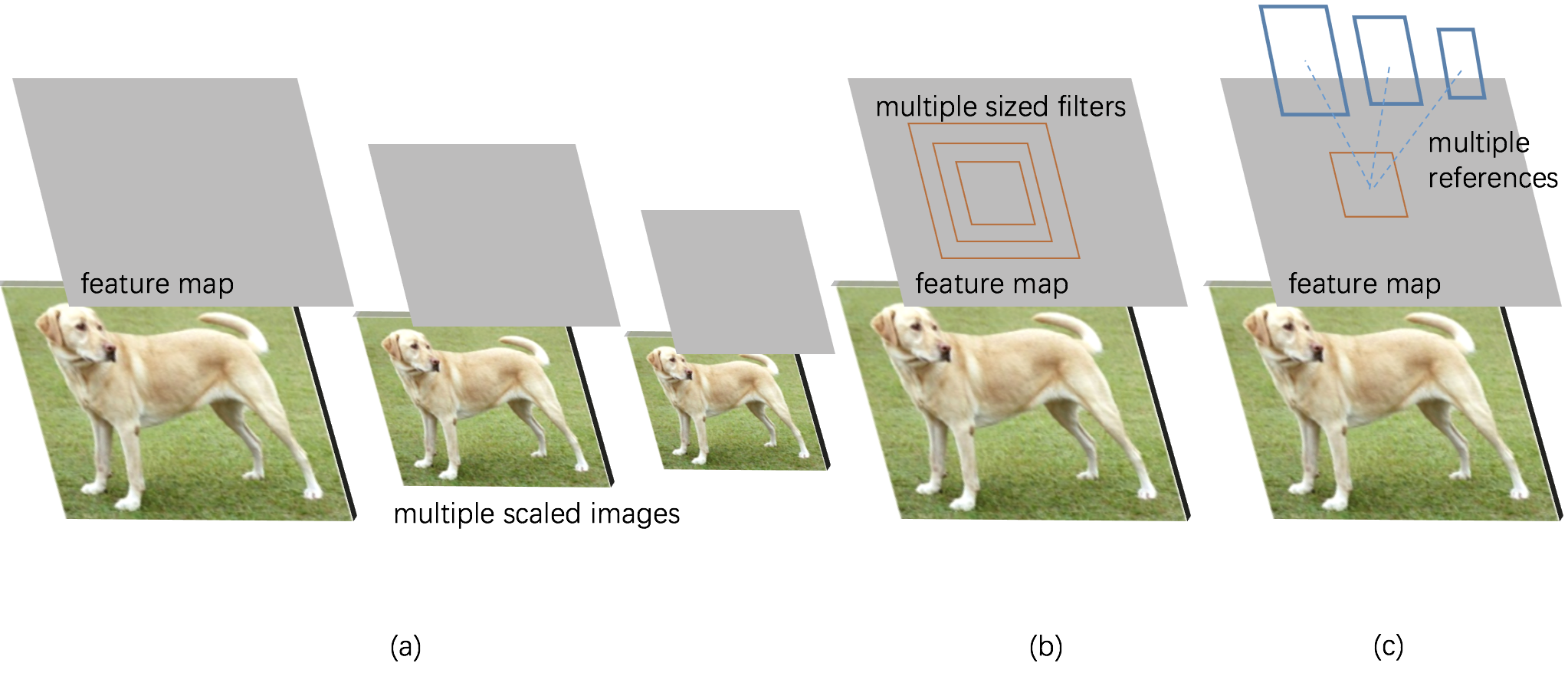}
% where an .eps filename suffix will be assumed under latex,
% and a .pdf suffix will be assumed for pdflatex; or what has been declared
% via \DeclareGraphicsExtensions.
\caption{To meet various scales of objects issue, there are three ways. (a) multiple scaled images detector trains each of them. (b) multiple sized filters separately act on the same sized image. (c) multiple pre-defined boxes are the reference of predicted boxes.}
\label{fig_sim}
\end{figure*}

As for the size of different objects in a dataset varies greatly, to address it, there are three commonly used methods. Firstly, input images are resized at multiple specified scales and feature maps are computed for each scale, called multi-scale training. Typical examples \cite{12fast_rcnn}\cite{singh2018analysis}\cite{he2015spatial}\cite{sermanet2013overfeat} use this method. Singh et al. \cite{singh2018sniper} adaptively sample regions from multiple scales of an image pyramid, conditioned on the image content. Secondly, researchers use convolutional filters of multiple scales on the feature maps. For instance, in \cite{felzenszwalb2009object}, models of different aspect ratios are trained separately using different filter sizes (such as ${5}\times{7}$ and ${7}\times{5}$ ). Thirdly, pre-defined anchors with multi-scales and multiple aspect ratios are reference boxes of the predicted bounding boxes. Faster R-CNN \cite{6faster_rcnn} and SSD \cite{8ssd} use reference box in two-stage and one-stage detectors for the first time, respectively. Fig. 7 is a schematic diagram of the above three cases.

\begin{figure}[!t]
\centering
\includegraphics[width=3.6in]{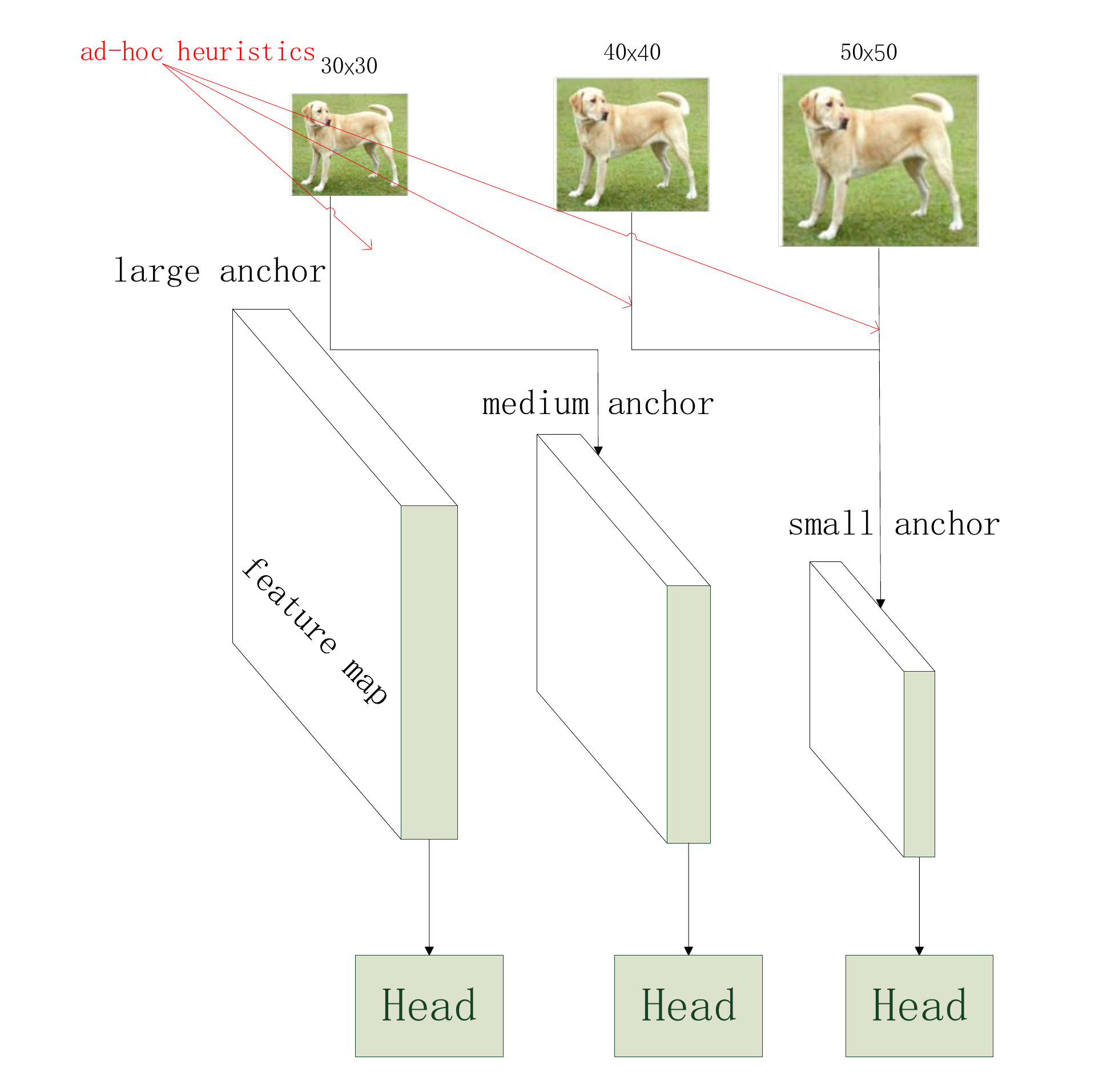}
% where an .eps filename suffix will be assumed under latex,
% and a .pdf suffix will be assumed for pdflatex; or what has been declared
% via \DeclareGraphicsExtensions.
\caption{An anchor-based architecture require heuristics to determine which size level anchors are responsible for what scale range of objects.}
\label{fig_sim}
\end{figure}

\subsection{anchor-free}
While there are constellation anchor-based object detectors being mainstream method which contain both one-stage and two-stage detectors making significant performance improvements, such as SSD, Faster R-CNN, YOLOv2, YOLOv3, they still suffer some drawbacks. 

(1) The pre-defined anchor boxes have a set of hand-crafted scales and aspect ratios which are sensitive to dataset and affect the detection performance by a large margin. 

(2) The scales and aspect ratios of pre-defined anchor boxes are kept fixed during training, thus the next step can’t get adaptively adjust boxes. Meanwhile, detectors have trouble handling objects of all sizes.

(3) For densely place anchor boxes to achieve high recall, especially on large-scale dataset, the computation cost and memory requirements bring huge overhead during processing procedure.

(4) Most of pre-defined anchors are negative samples, which causes great imbalance between positive and negative sample during training.

To address that, recently a series of anchor-free methods \cite{law2018cornernet} \cite{tian2019fcos} \cite{kong2019foveabox} \cite{law2019cornernet}\cite{duan2019centernet} \cite{wang2019region} \cite{zhou2019bottom} \cite{DBLP:journals/corr/abs-1904-07850} \cite{chen2019dubox} \cite{zhu2019feature} are proposed. CenterNet \cite{duan2019centernet} locates the center point, top-left and bottom-right point of an object. Tian et al. \cite{tian2019fcos} propose a localization method which is based on the four distance values between the predicted center point and four sides of a bounding box. It is still a novel direction for further research.

\subsection{Training from scratch}
Almost all of the state-of-the-art detectors utilize off-the-shelf classification backbone pre-trained on large scale classification dataset \cite{3Russakovsky2015} as their initial parameter set then fine-tune parameters to adapt to the new detection task. Another way to implement training procedure is that all parameters are assigned from scratch. Zhu et al. \cite{zhu2019scratchdet} train detector from scratch thus do not need pre-trained classification backbone because of stable and predictable gradient brought by batch normalization operation. Some works \cite{shen2017dsod} \cite{shen2017learning} \cite{li2018tiny} \cite{shen2018object} train object detectors from scratch by dense layer-wise connections.

\subsection{Designing new architecture}
Because of different propose of classification and localization task, there exists a gap between classification network and detection architecture. Localization needs fine-grained representations of objects while classification needs high semantic information. Li et al. \cite{li2018detnet} propose a newly designed object detection architecture to specially focus on detection task which maintains high spatial resolution in deeper layers and does not need to pre-train on large scale classification dataset.

The two-stage detectors are always slower than one-stage detectors. By studying the structure of two-stage network, researchers find two-stage detectors like Faster R-CNN and R-FCN have a heavy head which slows it down. Li et al. \cite{li2017light} present a light head two-stage detector to keep time efficiency.

\subsection{Speeding up detection}
For limited computing power and memory resource such as mobile devices, real-time devices, webcam, automatic driving encourage research into efficient detection architecture design. The most typical real-time detector is the \cite{7yolo} \cite{15yolov2} \cite{16Redmon2018YOLOv3AI} series and \cite{8ssd} \cite{18dssd_fu2017dssd} and their improved architecture \cite{chen2017weaving} \cite{zheng2018extend} \cite{cao2018feature} \cite{womg2018tiny}. Some methods \cite{wang2018pelee} \cite{yu2018loss} \cite{tychsen2017denet} \cite{tripathi2017lcdet} \cite{lee2017wide} \cite{li2017mimicking} are aim to reach real-time detection.

\subsection{Achieving Fast and Accurate Detections}
The best object detector needs both high efficiency and high accuracy which is the ultimate goal of this task. Lin et al. \cite{17retinanet} aim to surpass the accuracy of existing two-stage detectors while maintain fast speed. Zhou et al. \cite{zhou2017adaptive} combine an accurate (but slow) detector and a fast (but less accurate) detector adaptively determining whether an image is easy or hard to detect and choosing an appropriate detector to detect it. Liu et al. \cite{liu2018receptive} build a fast and accurate detector by strengthening lightweight network features using receptive fields block.
\section{Applications and branches}
\subsection{Typical application areas}
Object detection has been widely used in some fields to assist people to complete some tasks, such as security field, military field, transportation field, medical field and life field. We describe the typical and recent methods utilized in these fields in detail.
\subsubsection{Security field}
The most well known applications in the security field are face detection, pedestrian detection, fingerprint identification, fraud detection, anomaly detection etc.

$\bullet$ \textbf{Face detection} aims at detecting people faces in an image, as shown in Fig. 9. Because of extreme poses, illumination and resolution variations, face detection is still a difficult mission. Many works focus on precise detector designing. Ranjan et al. \cite{8170321} learn correlated tasks (face detection, facial landmarks localization, head pose estimation and gender recognition) simultaneously to boost the performance of individual tasks.  He et al. \cite{8370677} propose a novel Wasserstein convolutional neural network approach to learn invariant features between near-infrared (NIR) and visual (VIS) face images. Designing appropriate loss functions can enhance discriminative power of DCNNs based large-scale face recognition. The cosine-based softmax losses \cite{zhang2019adacos}\cite{liu2017rethinking}\cite{ranjan2017l2}\cite{wang2017normface} achieve great success in deep learning based face recognition. Deng et al. \cite{deng2018arcface} propose an Additive Angular Margin Loss (ArcFace) to get highly discriminative features for face recognition. Guo et al. \cite{guo2017fuzzy} give a fuzzy sparse auto-encoder framework for single image per person face recognition. Please refer to \cite{wang2018deep} for more details.

$\bullet$ \textbf{Pedestrian detection} focuses on detecting pedestrians in the natural scenes.  Braun et al. \cite{8634919} release an EuroCity Persons dataset containing pedestrians, cyclists and other riders in urban traffic scenes. Complexity-aware cascaded pedestrian detectors \cite{8686227}\cite{saberian2012learning}\cite{dollar2014fast} devote to real time pedestrian detection. Please refer to a survey \cite{brunetti2018computer} for more details.

$\bullet$ \textbf{Anomaly detection} plays a significant role in fraud detection, climate analysis, and healthcare monitoring.  Existing anomaly detection techniques \cite{liu2013change}\cite{senin2018grammarviz}\cite{jiang2015general}\cite{wu2008spatio} analyze the data on a point-wise basis. To point the expert analysts to the interesting regions (anomalies) of the data, Barz et al. \cite{8352745} propose a novel unsupervised method called “Maximally Divergent Intervals” (MDI), which searches for contiguous intervals of time and regions in space.

\begin{figure}[!t]
\centering
\includegraphics[width=3.6in]{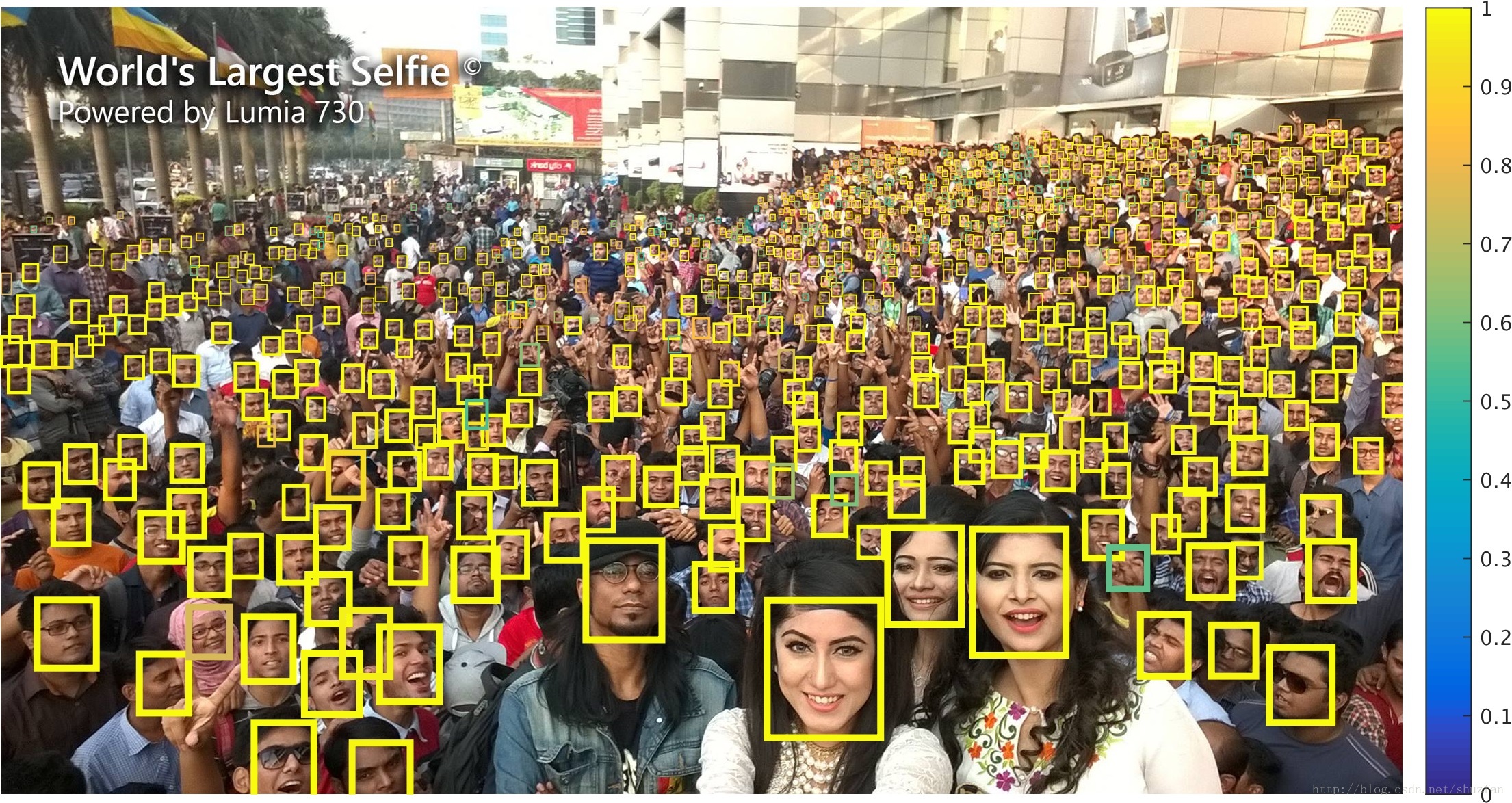}
% where an .eps filename suffix will be assumed under latex,
% and a .pdf suffix will be assumed for pdflatex; or what has been declared
% via \DeclareGraphicsExtensions.
\caption{A challenging densely tiny human faces detection results. Image from Hu et al. \cite{hu2017finding}.}
\label{fig_sim}
\end{figure}

\subsubsection{Military field}
In military field, remote sensing object detection, topographic survey, flyer detection, etc. are representative applications.

$\bullet$ \textbf{Remote sensing object detection} aims at detecting objects on remote sensing images or videos, which meets some challenges. Firstly, the extreme large input size but small targets makes the existing object detection procedure too slow for practical use and too hard to detect. Secondly, the massive and complex backgrounds cause serious false detection. To address these issues, researchers adopt the method of data fusion. Due to the lack of information and small deviation, which caused great inaccuracy, they focused on the detection of small targets. Remote sensing images have some characteristics far from natural images, thus strong pipelines such as Faster R-CNN, FCN, SSD, YOLO can’t transfer well to the new data domain. Designing remote sensing dataset adapted detectors remains a research hot spot in this domain.

Cheng et al. \cite{cheng2016learning} propose a CNN-based Remote Sensing Image (RSI) object detection model dealing with the rotation problem by designing a rotation-invariant layer. Zhang et al. \cite{zhang2019hierarchical} present a rotation and scaling robust structure to address lacking rotation and scaling invariance in RSI object detection. Li et al. \cite{li2019r3} raise a rotatable region proposal network and a rotatable detection network considering the orientation of vehicles. Deng et al. \cite{deng2017toward} put forward an accurate-vehicle-proposal-network (AVPN) for small object detection. Audebert et al. \cite{audebert2017segment} utilize accurate semantic segmentation results to obtain detection of vehicles.  Li et al. \cite{li2018hsf} address large range of resolutions of ships (ranging from dozens of pixels to thousands) issue in ship detection. Pang et al. \cite{pang2019r2} propose a real-time remote sensing method. Pei et al. \cite{pei2017sar} present a deep learning framework on synthetic aperture radar (SAR) automatic target recognition. Long et al. \cite{long2017accurate} concentrate on automatically and accurately locating objects. Shahzad et al. \cite{shahzad2018buildings} propose a novel framework containing automatic labeling and recurrent neural network for detection.

Typical methods \cite{zhang2016weakly}\cite{han2014object}\cite{li2018hough}\cite{mou2018vehicle}\cite{chen2014vehicle}\cite{ammour2017deep}\cite{wang2016new}\cite{ma2019novel}\cite{dong2019sig}\cite{chen2019deep}\cite{zhu2019multiscale} all utilize deep neural networks to achieve detection task on remote sensing datasets. NWPU VHR-10 \cite{cheng2014multi}, HRRSD \cite{zhang2019hierarchical}, DOTA \cite{xia2018dota}, DLR 3K Munich \cite{liu2015fast} and VEDAI \cite{razakarivony2016vehicle} are remote sensing object detection benchmarks. We recommend readers refer to \cite{cheng2016survey} for more details on remote sensing object detection.

\subsubsection{Transportation field}
As we known that, license plate recognition, automatic driving and traffic sign recognition etc.  greatly facilitate people's life.

$\bullet$ With the popularity of cars, \textbf{license plate recognition} is required in tracking crime, residential access, traffic violations tracking etc. Edge information, mathematical morphology, texture features, sliding concentric windows, connected component analysis etc. can bring license plate recognition system more robust and stable. Recently, deep learning-based methods \cite{shivakumara2018cnn}\cite{sarfraz2019approach}\cite{li2018toward}\cite{qian2018fast}\cite{laroca2018robust} provide a variety of solutions for license plate recognition. Please refer to \cite{nair2018survey} for more details.

$\bullet$ An autonomous vehicle (AV) needs an accurate perception of its surroundings to operate reliably. The perception system of an AV normally employs machine learning (e.g., deep learning) and transforms sensory data into semantic information which enables \textbf{autonomous driving}. Object detection is a fundamental function of this perception system. 3D object detection methods involve a third dimension that reveals more detailed object's size and location information, which are divided into three categories, monocular, point-cloud and fusion. First, monocular image based methods predict 2D bounding boxes on the image then extrapolate them to 3D, which lacks explicit depth information so limits the accuracy of localization. Second, point-cloud based methods project point clouds into a 2D image to process or generate a 3D representation of the point cloud directly in a voxel structure, where the former loses information and the latter is time consuming. Third, fusion based methods fuse both front view images and point clouds to generate a robust detection, which represent state-of-the-art detectors while computationally expensive.  Recently, Lu et al. \cite{lu2019l3} utilize a novel architecture contains 3D convolutions and RNNs to achieve centimeter-level localization accuracy in different real-world driving scenarios. Song et al. \cite{song2018apollocar3d} release a 3D car instance understanding benchmark for autonomous driving. Banerjee et al. \cite{banerjee2018online} utilize sensor fusion to obtain better features. Please refer to a recently published survey \cite{arnold2019survey} for more details.

$\bullet$ Both unmanned vehicles and autonomous driving systems need to solve the problem of \textbf{traffic sign recognition}. For the sake of safety and obeying the rules, real-time accurate traffic sign recognition assists in driving by acquiring the temporal and spatial information of the potential signs. Deep learning methods \cite{li2018real}\cite{moritani2018traffic}\cite{khalid2018automatic}\cite{arcos2018deep}\cite{li2018deepsign}\cite{wu2018traffic}\cite{zhou2018improved} solve this problem with high performance.

\subsubsection{Medical field}
In medical field, medical image detection, cancer detection, disease detection, skin disease detection and healthcare monitoring etc. have become a means of supplementary medical treatments. 

$\bullet$ \textbf{Computer Aided Diagnosis (CAD) systems} can help doctors classify different types of cancer. In detail, after an appropriate acquisition of the images, the fundamental steps carried out by a CAD framework can be identified as image segmentation, feature extraction, classification and object detection. Due to significant individual differences, data scarcity and privacy, there usually exists data distribution difference between source domain and target domain. A domain adaptation framework  \cite{li2019clu} is needed for medical image detection. 

$\bullet$ Li et al. \cite{li2019attention} incorporate the attention mechanism in CNN for \textbf{glaucoma detection} and establish a large-scale attention-based glaucoma dataset. Liu et al. \cite{naturecommunications2} design a bidirectional recurrent neural network (RNN) with long short-term memory (LSTM) to detect DNA modifications called DeepMod. Schubert et al. \cite{naturecommunications3} propose cellular morphology neural networks (CMNs) for automated neuron reconstruction and \textbf{automated detection of synapses}.
Codella et al. \cite{codella2018skin} organize a challenge of \textbf{skin lesion analysis} toward melanoma detection. Please refer to two representative surveys \cite{naji2018survey} \cite{altaf2019going} for more details.

\subsubsection{Life field}
In life field, intelligent home, commodity detection, event detection, pattern detection, image caption generation, rain/shadow detection, species identification etc. are the most representative applications.

$\bullet$ On densely packed scenes like \textbf{retail shelf displays}, Goldman et al. \cite{DBLP:journals/corr/abs-1904-00853} propose a novel precise object detector and release a new SKU-110K dataset to meet this challenge.

$\bullet$ \textbf{Event detection} aims to discover real-world events from the Internet such as festivals, talks, protests, natural disasters, elections. With the popularity of social media and its new characters, the data type of which are more diverse than before. Multi-domain event detection (MED) provides comprehensive descriptions of events. Yang et al. \cite{8618422} present an event detection framework to dispose multi-domain data. Wang et al. \cite{wang2012social} incorporate online social interaction features by constructing affinity graphs for event detection tasks. Schinas et al. \cite{schinas2015multimodal} design a multimodal graph-based system to detect events from 100 million photos/videos. Please refer to a survey \cite{hasan2018survey} for more details.

$\bullet$ \textbf{Pattern detection} always meet some challenges such as, scene occlusion, pose variation, varying illumination and sensor noise. To better address repeated pattern or periodic structure detection, researches design strong baselines in both 2D images  \cite{teboul2011shape} \cite{zhao2010rectilinear} and 3D point clouds \cite{friedman2013online} \cite{shen2011adaptive} \cite{schindler2008detecting}\cite{wu2010detecting}\cite{muller2007image}\cite{barinova2010geometric}\cite{kozinski2015mrf}\cite{cohen2014efficient}\cite{gandy2011tensor}\cite{candes2011robust}\cite{liu2012tensor}\cite{8417901}.

$\bullet$ \textbf{Image caption generation} means that computers automatically generate a caption for a given image. The most important part is to capture semantic information of images and express it to natural languages. Image captioning needs to connect computer vision and natural language processing technologies, which is a great challenge task. To address this issue, multimodal embedding, encoder–decoder frameworks, attention mechanism \cite{xu2015show} \cite{anderson2018bottom}, and reinforcement learning \cite{vinyals2015show} \cite{gu2018stack} are widely adopted in this field. Yao et al. \cite{yao2018exploring} introduce a new design to explore the connections between objects by constructing Graph Convolutional Networks and Long Short-Term Memory (dubbed as GCN-LSTM) architecture. This framework integrates both semantic and spatial object relationships. Apart from LSTM (long short term memory)-based methods, deep convolutional networks based method \cite{aneja2018convolutional} is verified effective and efficient. Please refer to a survey \cite{bai2018survey} for more details.

$\bullet$ Yang et al. \cite{8627954} present a novel rain model accompany with a deep learning architecture to address \textbf{rain detection} in a single image. Hu et al. \cite{8723605} analyze the spatial image context in a direction-aware manner and design a novel deep neural network to \textbf{detect shadow}. Accurate species identification is the basis for taxonomic research, a recently work \cite{waldchen2018machine} introduces a deep learning method for \textbf{species identification}. 

\subsection{Object detection branches}
Object detection has a wide range of application scenarios. The research of this domain contains a large variety of branches. We describe some representative branches in this part.
\subsubsection{Weakly supervised object detection}
Weakly supervised object detection (WSOD) aims at utilizing a few fully annotated images (supervision) to detect a large amount of non-fully annotated ones. Traditionally models are learnt from images labelled only with the object class and not the object bounding box. Annotating a bounding box for each object in large datasets is expensive, laborious and impractical. Weakly supervised learning relies on incomplete annotated training data to learn detection models. 

Weakly supervised deep detection network in \cite{bilen2016weakly} is a representative framework for weakly supervised object detection. Context information \cite{kantorov2016contextlocnet}, instance classifier refinement \cite{tang2017multiple} and  image segmentation \cite{diba2017weakly}\cite{li2016image} are adopted to tackle hardly optimized problems. Yang et al. \cite{yang2019activity} show that the action depicted in the image can provide strong cues about the location of the associated object. Wan et al. \cite{8640243} design a min-entropy latent model optimized with a recurrent learning algorithm for weakly supervised object detection. Tang et al. \cite{8493315} utilize an iterative procedure to generate proposal clusters and learn refined instance classifiers, which makes the network concentrate on the whole object rather than part of it. Cao et al. \cite{8370896} design a novel feedback convolutional neural network for weakly supervised object localization. Wan et al. \cite{wan2019c} present continuation multiple instance learning to alleviate the non-convexity problem in WSOD.

\subsubsection{Salient object detection}
Salient object detection utilizes deep neural network to predict saliency scores of image regions and obtain accurate saliency maps, as shown in Fig. 10. Salient object detection networks usually need to aggregate multi-level features of backbone network. For fast speed without accuracy dropping, Wu et al. \cite{wu2019cascaded} present that discarding the shallower layer features can achieve fast speed and the deeper layer features are sufficient to obtain precisely salient map. Liu et al. \cite{liu2019simple} expand the role of pooling in convolutional neural networks. Wang et al. \cite{8668551} utilize fixation prediction to detect salient objects. Wang et al. \cite{8382302} adopt recurrent fully convolutional networks and incorporate saliency prior knowledge for accurate salient object detection. Feng et al. \cite{feng2019attentive} design an attentive feedback module to better explore the structure of objects.

Video salient object detection datasets \cite{fan2019shifting}\cite{kim2015spatiotemporal}\cite{li2013video}\cite{li2017benchmark}\cite{liu2016saliency}\cite{ochs2013segmentation}\cite{wang2015consistent} provide benchmarks for video salient object detection, and existing good algorithms \cite{chen2017video} \cite{chen2018scom} \cite{kim2015spatiotemporal}  \cite{liu2016saliency} \cite{li2018unsupervised} \cite{liu2014superpixel} \cite{song2018pyramid} \cite{tang2018weakly} \cite{wang2017video} \cite{tu2016real} \cite{wang2015saliency}\cite{xi2016salient}\cite{zhang2015minimum}\cite{zhou2014time} devote to the development of this field.

\begin{figure*}[!t]
\centering
\includegraphics[width=7in]{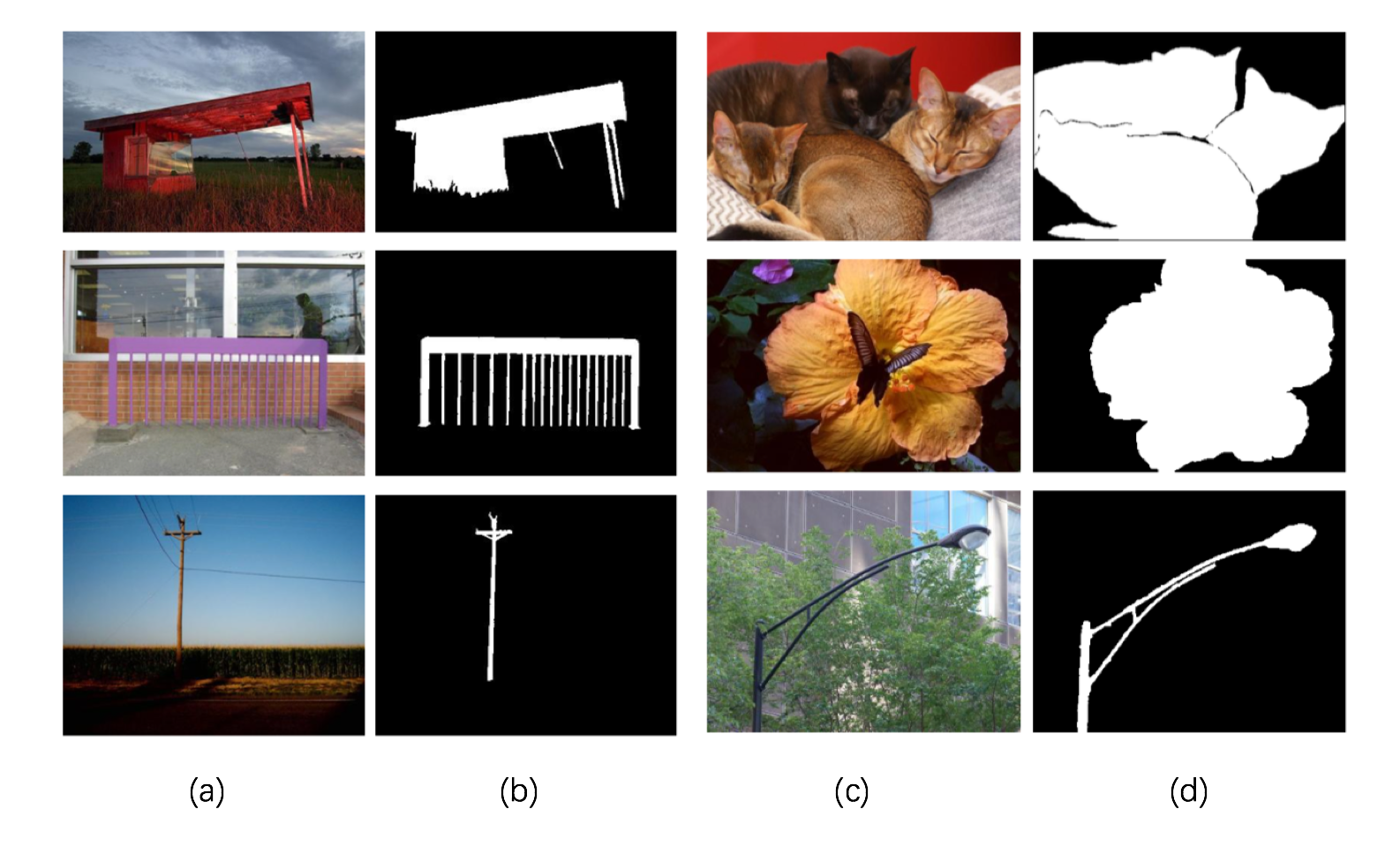}
% where an .eps filename suffix will be assumed under latex,
% and a .pdf suffix will be assumed for pdflatex; or what has been declared
% via \DeclareGraphicsExtensions.
\caption{Some examples from the salient object detection datasets. (a), (c) are images, (b), (d) ground truth. Image from Liu et al. \cite{liu2019simple} and Wu et al. \cite{wu2019cascaded}.}
\label{fig_sim}
\end{figure*}

\subsubsection{Highlight detection}
Highlight detection is to retrieve a moment in a short video clip that captures a user’s primary attention or interest, which can accelerate browsing many videos, enhance social video sharing and facilitate video recommendation. Typical highlight detectors  \cite{sun2014ranking} \cite{yao2016highlight} \cite{yang2015unsupervised} \cite{liu2015multi} \cite{panda2017weakly} \cite{potapov2014category} are domain-specific for they are tailored to a category of videos. All object detection tasks require a large amount of manual annotation data and highlight detection is no exception. Xiong et al. \cite{xiong2019less} propose a weakly supervised method on shorter user-generated videos to address this issue.

\subsubsection{Edge detection}
Edge detection aims at extracting object boundaries and perceptually salient edges from images, which is important to a series of higher level vision tasks like segmentation, object detection and recognition. Edge detection meets some challenges. First, the edges of various scales in an image need both object-level boundaries and useful local region details. Second, convolutional layers of different levels are specialized to predict different parts of the final detection, thus each layer in CNN should be trained by proper layer-specific supervision. To address these issues, He et al. \cite{he2019bi} propose a Bi-Directional Cascade Network to let one layer supervised by labeled edges while adopt dilated convolution to generate multi-scale features. Liu et al. \cite{8516362} present an accurate edge detector which utilizes richer convolutional features.

\subsubsection{Text detection}
Text detection aims to identify text regions of given images or videos which is also an important prerequisite for many computer vision tasks, such as classification, video analysis. There have been many successful commercial optical character recognition (OCR) systems for internet content and documentary texts recognition. The detection of text in natural scenes remains a challenge due to complex situations such as blurring, uneven lighting, perspective distortion, various orientation. Some typical works \cite{ren2016convolutional}\cite{liao2017textboxes}\cite{bazazian2017improving} focus on horizontal or nearly horizontal text detection. Recently, researchers find that arbitrary-oriented text detection \cite{zhang2016multi}\cite{yao2016scene}\cite{he2016accurate}\cite{lyu2018multi}\cite{ma2018arbitrary} is a direction that needs to pay attention to. In general, deep learning based scene text detection methods can be classified into two categories. The first category takes scene text as a type of general object, following the general object detection paradigm and locating scene text by text box regression. These methods have difficulties to deal with the large aspect ratios and arbitrary-orientation of scene text. The second one directly segments text regions, but mostly requires complicated post-processing step. Usually, some methods in this category mainly involve two steps, segmentation (generating text prediction maps) and geometric approaches (for inclined proposals), which is time-consuming. In addition, in order to obtain the desired orientation of text boxes, some methods require complex post-processing step, so it's not as efficient as those architectures that are directly based on detection networks.

Lyu et al. \cite{lyu2018multi} combine the ideas of the two categories above avoiding their shortcomings by locating corner points of text bounding boxes and dividing text regions in relative positions to detect scene text, which can handle long oriented text and only need a simple NMS post-processing step. Ma et al. \cite{ma2018arbitrary} develop a novel rotation-based approach and an end-to-end text detection system in which Rotation Region Proposal Networks (RRPN) generate inclined proposals with text orientation angle information.

\subsubsection{Multi-domain object detection}
Domain-specific detectors always achieve high detection performance on the specified dataset. So as to get a universal detector which is capable of working on various image domains, recently many works focus on training a multi-domain detector while do not require prior knowledge of the newly domain of interest. Wang et al. \cite{wang2019towards} propose a universal detector which utilizes a new domain-attention mechanism working on a variety of image domains (human faces, traffic signs and medical CT images) without prior knowledge of the domain of interest. Wang et al. \cite{wang2019towards} release a newly established universal object detection benchmark consisting of 11 diverse datasets to better meet the challenges of generalization in different domains. 

To learn a universal representation of vision, Bilen et al. \cite{bilen2017universal} add domain-specific BN (batch normalization) layers to a multi-domain shared network.  Rebuffi et al. \cite{rebuffi2017learning} propose adapter residual modules which achieve a high degree of parameter sharing while maintaining or even improving the accuracy of domain-specific representations. Rebuffi et al. \cite{rebuffi2017learning} introduce the Visual Decathlon Challenge, a benchmark contains ten very different visual domains. Inspired by transfer learning, Rebuffi et al. \cite{rebuffi2018efficient} empirically study efficient parameterizations and outperform traditional fine-tuning techniques.

Another requirement for multi-domain object detection is to reduce annotation costs. Object detection datasets need heavily annotation works which is time consuming and mechanical. Transferring pre-trained models from label-rich domains to label-poor datasets can solve label-poor detection works. One way is to use unsupervised domain adaptation methods to tackle dataset bias problems. In recent years, researchers have adopted adversarial learning to align the source and target distribution of samples. Chen et al. \cite{chen2018domain} utilize Faster R-CNN with a domain classifier trained to distinguish source and target samples, like adversarial learning, where the feature extractor learns to deceive the domain classifier. Saito et al. \cite{saito2018strong} propose a weak alignment model to focus on similarity between different images from domains with large discrepancy rather than aligning images that are globally dissimilar. Only in the source domain manual annotations are available, which can be addressed by using Unsupervised Domain Adaptation methods. Haupmann et al. \cite{haupmann2019contrastive} propose an Unsupervised Domain Adaptation method which models both intra-class and inter-class domain discrepancy.

\subsubsection{Object detection in videos}
Object detection in videos aims at detecting objects in videos, which brings additional challenges due to degraded image qualities such as motion blur and video defocus, leading to unstable classifications for the same object across video. Video detectors \cite{han2016seq}\cite{feichtenhofer2017detect}\cite{kang2016object}\cite{kang2017t}\cite{kang2017object}\cite{zhu2017deep}\cite{zhu2017flow}\cite{wang2015visual}\cite{bertasius2018object}\cite{xiao2018video} exploit temporal contexts to meet this challenge. Some static detectors \cite{han2016seq}\cite{feichtenhofer2017detect}\cite{kang2016object}\cite{kang2017t} first detect objects in each frame then check them by linking detections of the same object in neighbor frames. Due to object motion, the same object in neighbor frames may not have a large overlap. On the other hand, the predicted object movements are not accurate enough to link neighbor frames. Tang et al. \cite{8686124} propose an architecture which links objects in the same frame instead of neighboring frames to address it.

\subsubsection{Point clouds 3D object detection}
Compared to image based detection, LiDAR point cloud provides reliable depth information that can be used to accurately locate objects and characterize their shapes. In autonomous navigation, autonomous driving, housekeeping robots and augmented/virtual reality applications, LiDAR point cloud based 3D object detection plays an important role. Point cloud based 3D object detection meets some challenges, the sparsity of LiDAR point clouds, highly variable point density, non-uniform sampling of the 3D space, effective range of the sensors, occlusion, and the relative pose variation. Engelcke et al. \cite{engelcke2017vote3deep} first propose sparse convolutional layers and L1 regularization for efficient large-scale processing of 3D data. Qi et al. \cite{qi2017pointnet} raise an end-to-end deep neural network called PointNet, which learns point-wise features directly from point clouds. Qi et al. \cite{qi2017pointnet++} improve PointNet which learns local structures at different scales. Zhou et al. \cite{zhou2018voxelnet} close the gap between RPN and point set feature learning for 3D detection task. Zhou et al. \cite{zhou2018voxelnet} present a generic end-to-end 3D detection framework called VoxelNet, which learns a discriminative feature representation from point clouds and predicts accurate 3D bounding boxes simultaneously.   

In autonomous driving application, Chen et al. \cite{chen2016monocular} perform 3D object detection from a single monocular image. Chen et al. \cite{chen2017multi} take both LiDAR point cloud and RGB images as input then predict oriented 3D bounding boxes for high-accuracy 3D object detection. Example 3D detection result is shown in Fig. 11.

\begin{figure*}[!t]
\centering
\includegraphics[width=7in]{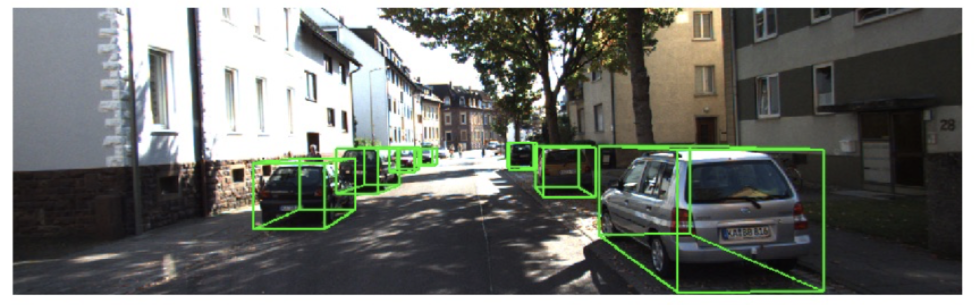}
% where an .eps filename suffix will be assumed under latex,
% and a .pdf suffix will be assumed for pdflatex; or what has been declared
% via \DeclareGraphicsExtensions.
\caption{Example 3D detection result from the KITTI validation set projected onto an image. Image from Vishwanath A. Sindagi et al. \cite{sindagi2019mvx}.}
\label{fig_sim}
\end{figure*}

\subsubsection{2D, 3D pose detection}
Human pose detection aims at estimating the 2D or 3D pose location of the body joints and defining pose classes then returning the average pose of the top scoring class, as shown in Fig. 12. Typical 2D human pose estimation methods \cite{cao2017realtime}\cite{bulat2016human}\cite{newell2016stacked}\cite{chen2014articulated}\cite{toshev2014deeppose}\cite{fan2015combining}\cite{ouyang2014multi} utilize deep CNN architectures. Rogez et al. \cite{8611390} propose an end-to-end architecture for joint 2D and 3D human pose estimation in natural images which predicts 2D and 3D poses of multiple people simultaneously. Benefit by full-body 3D pose, it can recover body part locations in cases of occlusion between different targets. Human pose estimation approaches can be divided into two categories, one-stage and multi-stage methods. The best performing methods \cite{chen2018cascaded}\cite{9mask_rcnn}\cite{papandreou2017towards}\cite{xiao2018simple} typically base on one-stage backbone networks. The most representative multi-stage methods are convolutional pose machine \cite{wei2016convolutional}, Hourglass network \cite{newell2016stacked}, and MSPN \cite{li2019rethinking}.

\begin{figure*}[!t]
\centering
\includegraphics[width=7in]{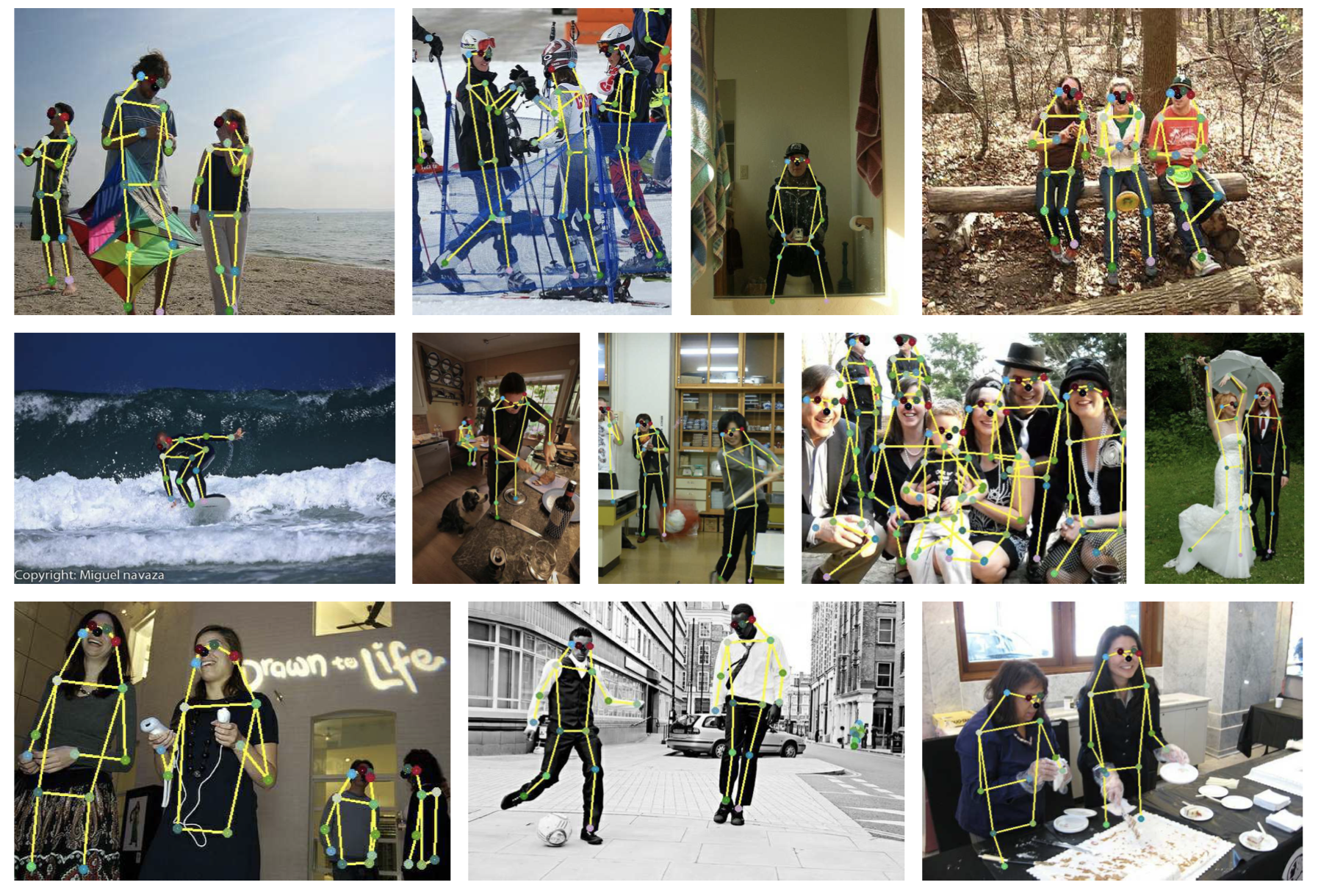}
% where an .eps filename suffix will be assumed under latex,
% and a .pdf suffix will be assumed for pdflatex; or what has been declared
% via \DeclareGraphicsExtensions.
\caption{Some examples of multi-person pose estimation. Image from Chen et al. \cite{chen2018cascaded}.}
\label{fig_sim}
\end{figure*}

\subsubsection{Fine-Grained Visual Recognition}
Fine-grained recognition aims to identify an exact category of objects in each basic-level category, such as identifying the species of a bird, or the model of an aircraft. This task is quite challenging because the visual differences between the categories are small and can be easily overwhelmed by those caused by factors such as pose, viewpoint, and location of the object in the image. To generalize across viewpoints, Krause et al. \cite{Krause_2013_ICCV_Workshops} utilize 3D object representations on the level of both local feature appearance and location. Lin et al. \cite{lin2015bilinear} introduce bilinear models that consists of two feature extractors (two CNN streams). The outputs of these two feature extractors are multiplied using outer product at each location of the image and then pooled to obtain an image descriptor. He et al. \cite{he2017fine} introduce a fine-grained discriminative localization method via saliency-guided Faster R-CNN. After that, He et al. \cite{he2018fast} propose a weakly supervised discriminative localization approach (WSDL) for fast fine-grained image classification. Classical datasets \cite{khosla2011novel} \cite{maji2013fine} provide useful information on some interesting categories. Please refer to a survey \cite{zhao2017survey} for more details.

\section{Conclusions and trends}
\subsection{Conclusions}
With the continuous upgrading of powerful computing equipment, object detection technology based on deep learning has been developed rapidly. In order to deploy on more accurate applications, the need for high precision real-time systems is becoming more and more urgent. Since achieving high accuracy and efficiency detectors is the ultimate goal of this task, researchers have developed a series of directions such as, constructing new architecture, extracting rich features, exploiting good representations, improving processing speed, training from scratch, anchor-free methods, solving sophisticated scene issues (small objects, occluded objects), combining one-stage and two-stage detectors to make good results, improving post-processing NMS method, solving negatives-positives imbalance issue, increasing localization accuracy, enhancing classification confidence. With the increasingly powerful object detectors in security field, military field, transportation field, medical field, and life field, the application of object detection is gradually extensive. In addition, a variety of branches in detection domain arise. Although the achievement of this domain has been effective recently, there is still much room for further development.
\subsection{Trends}
\subsubsection{Combining one-stage and two-stage detectors}
On the one hand, the two-stage detectors have a densely tailing process to obtain as many as reference boxes, which is time consuming and inefficient. To address this issue, researchers are required to eliminate so much redundancy while maintaining high accuracy. On the other hand, the one-stage detectors achieve fast processing speed which have been used successfully in real-time applications. Although fast, the lower accuracy is still a bottleneck for high precision requirements. How to combine the advantages of both one-stage and two-stage detectors remains a big challenge. 

\subsubsection{Video object detection}
In video object detection, motion blur, video defocus, motion target ambiguity, intense target movements, small targets, occlusion and truncation etc. make it difficult for this task to achieve good performance in real life scene and remote sensing scene. Delving into moving goals and more complex source data such as video is one of the key points for future research.

\subsubsection{Efficient post-processing methods}
In the three (for one-stage detectors) or four (for two-stage detectors) stage detection procedure, post-processing is an initial step for the final results. On most of the detection metrics, only the highest prediction result of one object can be send to the metric program to calculate accuracy score.  The post-processing methods like NMS and its improvements may eliminate well located but high classification confidence objects, which is detrimental to the accuracy of the measurement. Exploiting more efficient and accurate post-processing method is another direction for object detection domain.

\subsubsection{Weakly supervised object detection methods}
Utilizing high proportion labelled images only with object class but not with object bounding box to replace a large amount of fully annotated images to train the network is of high efficiency and easy to get. Weakly supervised object detection (WSOD) aims at utilizing a few fully annotated images (supervision) to detect a large amount of non-fully annotated ones. Therefore developing WSOD methods is a significant problem for further study.
\subsubsection{Multi-domain object detection}
Domain-specific detectors always achieve high detection performance on the specified dataset. So as to get a universal detector which is capable of working on various image domains, multi-domain detectors can solve this problem without prior knowledge of new domain. Domain transfer is a challenging mission for further study.

\subsubsection{3D object detection}
With the advent of 3D sensors and diverse applications of 3D understanding, 3D object detection gradually becomes a hot research direction. Compared to 2D image based detection, LiDAR point cloud provides reliable depth information that can be used to accurately locate objects and characterize their shapes. LiDAR enables accurate localization of objects in the 3D space. Object detection techniques based on LiDAR data often outperform the 2D counterparts as well.

\subsubsection{Salient object detection}
Salient object detection (SOD) aims at highlighting salient object regions in images. Video object detection is to classify and locate objects of interest in a continuous scene. SOD is driven by and applied to a widely spectrum of object-level applications in various areas. Given salient object regions of interest in each frame can assist accurate object detection in videos. Therefore, for high-level recognition task and challenging detection task, highlighting target detection is a crucial preliminary process.

\subsubsection{Unsupervised object detection} 
Supervised methods are time consuming and inefficient in training process, which need well annotated dataset used for supervision information. Annotating a bounding box for each object in large datasets is expensive, laborious and impractical. Developing automatic annotation technology to release human annotation work is a promising trend for unsupervised object detection. Unsupervised object detection is a future research direction for intelligent detection mission.

\subsubsection{Multi-task learning}
Aggregating multi-level features of backbone network is a significant way to improve detection performance. Furthermore, performing multiple computer vision tasks simultaneously such as object detection, semantic segmentation, instance segmentation, edge detection, highlight detection can enhance performance of separate task by a large margin because of richer information. Adopting multi-task learning is a good way to aggregate multiple tasks in a network, and it presents great challenges to researchers to maintain processing speed and improve accuracy as well.

\subsubsection{Multi-source information assistance} 
Due to the popularity of social media and the development of big data technology, multi-source information becomes easy to access. Many social media information can provide both pictures and descriptions of them in textual form, which can help detection task. Fusing multi-source information is an emerging research direction with the development of various technologies.

\subsubsection{Constructing terminal object detection system}
From the cloud to the terminal, the terminalization of artificial intelligence can help people deal with mass information and solve problems better and faster. With the emergence of lightweight networks, terminal detectors are developed into more efficient and reliable devices with broad application scenarios. The chip detection network based on FPGA will make real-time application possible.

\subsubsection{Medical imaging and diagnosis}
FDA (U.S. Food and Drug Administration) is promoting “AI-based Medical Devices”. In April 2018, FDA first approved an artificial intelligence software called IDx-DR, a diabetic retinopathy detector with an accuracy of more than 87.4\%. For customers, the combination of image recognition systems and mobile devices can make cell phone a powerful family diagnostic tool. This direction is full of challenges and expectations.

\subsubsection{Advanced medical biometrics}
Utilizing deep neural network, researchers began to study and measure atypical risk factors that had been difficult to quantify previously. Using neural networks to analyze retinal images and speech patterns may help identify the risk of heart disease. In the near future, medical biometrics will be used for passive monitoring.

\subsubsection{Remote sensing airborne and real-time detection}
Both military and agricultural fields require accurate analysis of remote sensing images. Automated detection software and integrated hardware will bring unprecedented development to these fields. Loading deep learning based object detection system to SoC (System on Chip) realizes real-time high-altitude detection.

\subsubsection{GAN based detector}
Deep learning based systems always require large amounts of data for training, whereas Generative Adversarial Network is a powerful structure to generate fake images. How much you need, how much it can produce. Mixing the real world scene and simulated data generated by GAN trains object detector to make the detector grow more robust and obtain stronger generalization ability.

The research of object detection still needs further study. We hope that deep learning methods will make breakthroughs in the near future.

\section*{Acknowledgment}

Thanks to the scholars involved in this paper. This paper quotes the research literature of several scholars. Without the help and inspiration of the research results of all scholars, it would be difficult for me to complete the writing of this paper.

We would like to express our gratitude to all those who helped us during the writing of this thesis.

\renewcommand{\refname}{Reference}

\bibliographystyle{ieeetr}

\bibliography{bare_jrnl.bib}

\begin{thebibliography}{100}

\bibitem{1}
P.~Dollar, C.~Wojek, B.~Schiele, and P.~Perona, ``Pedestrian detection: An
  evaluation of the state of the art,'' {\em IEEE Transactions on Pattern
  Analysis and Machine Intelligence}, vol.~34, pp.~743--761, April 2012.

\bibitem{2}
A.~Geiger, P.~Lenz, and R.~Urtasun, ``Are we ready for autonomous driving? the
  kitti vision benchmark suite,'' in {\em 2012 IEEE Conference on Computer
  Vision and Pattern Recognition}, pp.~3354--3361, June 2012.

\bibitem{3Russakovsky2015}
O.~Russakovsky, J.~Deng, H.~Su, J.~Krause, S.~Satheesh, S.~Ma, Z.~Huang,
  A.~Karpathy, A.~Khosla, M.~Bernstein, A.~C. Berg, and L.~Fei-Fei, ``Imagenet
  large scale visual recognition challenge,'' {\em International Journal of
  Computer Vision}, vol.~115, pp.~211--252, Dec 2015.

\bibitem{4Everingham2010}
M.~Everingham, L.~Van~Gool, C.~K.~I. Williams, J.~Winn, and A.~Zisserman, ``The
  pascal visual object classes (voc) challenge,'' {\em International Journal of
  Computer Vision}, vol.~88, pp.~303--338, Jun 2010.

\bibitem{5_10.1007/978-3-319-10602-1_48}
T.-Y. Lin, M.~Maire, S.~Belongie, J.~Hays, P.~Perona, D.~Ramanan,
  P.~Doll{\'a}r, and C.~L. Zitnick, ``Microsoft coco: Common objects in
  context,'' in {\em Computer Vision -- ECCV 2014} (D.~Fleet, T.~Pajdla,
  B.~Schiele, and T.~Tuytelaars, eds.), (Cham), pp.~740--755, Springer
  International Publishing, 2014.

\bibitem{OpenImages}
A.~Kuznetsova, H.~Rom, N.~Alldrin, J.~Uijlings, I.~Krasin, J.~Pont-Tuset,
  S.~Kamali, S.~Popov, M.~Malloci, T.~Duerig, and V.~Ferrari, ``The open images
  dataset v4: Unified image classification, object detection, and visual
  relationship detection at scale,'' {\em arXiv:1811.00982}, 2018.

\bibitem{11visdrone_zhu2018vision}
P.~Zhu, L.~Wen, X.~Bian, H.~Ling, and Q.~Hu, ``Vision meets drones: A
  challenge,'' {\em arXiv preprint arXiv:1804.07437}, 2018.

\bibitem{6faster_rcnn}
S.~Ren, K.~He, R.~Girshick, and J.~Sun, ``Faster r-cnn: Towards real-time
  object detection with region proposal networks,'' {\em IEEE Transactions on
  Pattern Analysis and Machine Intelligence}, vol.~39, pp.~1137--1149, June
  2017.

\bibitem{7yolo}
J.~Redmon, S.~Divvala, R.~Girshick, and A.~Farhadi, ``You only look once:
  Unified, real-time object detection,'' in {\em 2016 IEEE Conference on
  Computer Vision and Pattern Recognition (CVPR)}, pp.~779--788, June 2016.

\bibitem{8ssd}
W.~Liu, D.~Anguelov, D.~Erhan, C.~Szegedy, S.~Reed, C.-Y. Fu, and A.~C. Berg,
  ``Ssd: Single shot multibox detector,'' in {\em Computer Vision -- ECCV 2016}
  (B.~Leibe, J.~Matas, N.~Sebe, and M.~Welling, eds.), (Cham), pp.~21--37,
  Springer International Publishing, 2016.

\bibitem{9mask_rcnn}
K.~He, G.~Gkioxari, P.~Dollár, and R.~Girshick, ``Mask r-cnn,'' in {\em 2017
  IEEE International Conference on Computer Vision (ICCV)}, pp.~2980--2988, Oct
  2017.

\bibitem{khan2019survey}
A.~Khan, A.~Sohail, U.~Zahoora, and A.~S. Qureshi, ``A survey of the recent
  architectures of deep convolutional neural networks,'' {\em arXiv preprint
  arXiv:1901.06032}, 2019.

\bibitem{DBLP:journals/corr/abs-1905-05055}
Z.~Zou, Z.~Shi, Y.~Guo, and J.~Ye, ``Object detection in 20 years: {A}
  survey,'' {\em CoRR}, vol.~abs/1905.05055, 2019.

\bibitem{liu2018deep}
L.~Liu, W.~Ouyang, X.~Wang, P.~Fieguth, J.~Chen, X.~Liu, and
  M.~Pietik{\"a}inen, ``Deep learning for generic object detection: A survey,''
  {\em arXiv preprint arXiv:1809.02165}, 2018.

\bibitem{14fpn}
T.~Lin, P.~Dollár, R.~Girshick, K.~He, B.~Hariharan, and S.~Belongie,
  ``Feature pyramid networks for object detection,'' in {\em 2017 IEEE
  Conference on Computer Vision and Pattern Recognition (CVPR)}, pp.~936--944,
  July 2017.

\bibitem{li2018detnet}
Z.~Li, C.~Peng, G.~Yu, X.~Zhang, Y.~Deng, and J.~Sun, ``Detnet: A backbone
  network for object detection,'' {\em arXiv preprint arXiv:1804.06215}, 2018.

\bibitem{xie2017aggregated}
S.~Xie, R.~Girshick, P.~Doll{\'a}r, Z.~Tu, and K.~He, ``Aggregated residual
  transformations for deep neural networks,'' in {\em Proceedings of the IEEE
  conference on computer vision and pattern recognition}, pp.~1492--1500, 2017.

\bibitem{ghiasi2019fpn}
G.~Ghiasi, T.-Y. Lin, R.~Pang, and Q.~V. Le, ``Nas-fpn: Learning scalable
  feature pyramid architecture for object detection,'' {\em arXiv preprint
  arXiv:1904.07392}, 2019.

\bibitem{howard2017mobilenets}
A.~G. Howard, M.~Zhu, B.~Chen, D.~Kalenichenko, W.~Wang, T.~Weyand,
  M.~Andreetto, and H.~Adam, ``Mobilenets: Efficient convolutional neural
  networks for mobile vision applications,'' {\em arXiv preprint
  arXiv:1704.04861}, 2017.

\bibitem{zhang2018shufflenet}
X.~Zhang, X.~Zhou, M.~Lin, and J.~Sun, ``Shufflenet: An extremely efficient
  convolutional neural network for mobile devices,'' in {\em Proceedings of the
  IEEE Conference on Computer Vision and Pattern Recognition}, pp.~6848--6856,
  2018.

\bibitem{iandola2016squeezenet}
F.~N. Iandola, S.~Han, M.~W. Moskewicz, K.~Ashraf, W.~J. Dally, and K.~Keutzer,
  ``Squeezenet: Alexnet-level accuracy with 50x fewer parameters and< 0.5 mb
  model size,'' {\em arXiv preprint arXiv:1602.07360}, 2016.

\bibitem{chollet2017xception}
F.~Chollet, ``Xception: Deep learning with depthwise separable convolutions,''
  in {\em Proceedings of the IEEE conference on computer vision and pattern
  recognition}, pp.~1251--1258, 2017.

\bibitem{sandler2018mobilenetv2}
M.~Sandler, A.~Howard, M.~Zhu, A.~Zhmoginov, and L.-C. Chen, ``Mobilenetv2:
  Inverted residuals and linear bottlenecks,'' in {\em Proceedings of the IEEE
  Conference on Computer Vision and Pattern Recognition}, pp.~4510--4520, 2018.

\bibitem{wang2018pelee}
R.~J. Wang, X.~Li, and C.~X. Ling, ``Pelee: A real-time object detection system
  on mobile devices,'' in {\em Advances in Neural Information Processing
  Systems}, pp.~1963--1972, 2018.

\bibitem{13resnet}
K.~He, X.~Zhang, S.~Ren, and J.~Sun, ``Deep residual learning for image
  recognition,'' in {\em 2016 IEEE Conference on Computer Vision and Pattern
  Recognition (CVPR)}, pp.~770--778, June 2016.

\bibitem{19vgg_simonyan2014very}
K.~Simonyan and A.~Zisserman, ``Very deep convolutional networks for
  large-scale image recognition,'' {\em arXiv preprint arXiv:1409.1556}, 2014.

\bibitem{rawat2017deep}
W.~Rawat and Z.~Wang, ``Deep convolutional neural networks for image
  classification: A comprehensive review,'' {\em Neural computation}, vol.~29,
  no.~9, pp.~2352--2449, 2017.

\bibitem{10rcnn}
R.~Girshick, J.~Donahue, T.~Darrell, and J.~Malik, ``Rich feature hierarchies
  for accurate object detection and semantic segmentation,'' in {\em 2014 IEEE
  Conference on Computer Vision and Pattern Recognition}, pp.~580--587, June
  2014.

\bibitem{12fast_rcnn}
R.~Girshick, ``Fast r-cnn,'' in {\em 2015 IEEE International Conference on
  Computer Vision (ICCV)}, pp.~1440--1448, Dec 2015.

\bibitem{15yolov2}
J.~Redmon and A.~Farhadi, ``Yolo9000: Better, faster, stronger,'' in {\em 2017
  IEEE Conference on Computer Vision and Pattern Recognition (CVPR)},
  pp.~6517--6525, July 2017.

\bibitem{23batchnormalization_ioffe2015batch}
S.~Ioffe and C.~Szegedy, ``Batch normalization: Accelerating deep network
  training by reducing internal covariate shift,'' {\em arXiv preprint
  arXiv:1502.03167}, 2015.

\bibitem{16Redmon2018YOLOv3AI}
J.~Redmon and A.~Farhadi, ``Yolov3: An incremental improvement,'' {\em CoRR},
  vol.~abs/1804.02767, 2018.

\bibitem{17retinanet}
T.~Lin, P.~Goyal, R.~Girshick, K.~He, and P.~Dollár, ``Focal loss for dense
  object detection,'' in {\em 2017 IEEE International Conference on Computer
  Vision (ICCV)}, pp.~2999--3007, Oct 2017.

\bibitem{18dssd_fu2017dssd}
C.-Y. Fu, W.~Liu, A.~Ranga, A.~Tyagi, and A.~C. Berg, ``Dssd: Deconvolutional
  single shot detector,'' {\em arXiv preprint arXiv:1701.06659}, 2017.

\bibitem{zhao2018m2det}
Q.~Zhao, T.~Sheng, Y.~Wang, Z.~Tang, Y.~Chen, L.~Cai, and H.~Ling, ``M2det: A
  single-shot object detector based on multi-level feature pyramid network,''
  {\em arXiv preprint arXiv:1811.04533}, 2018.

\bibitem{zhang2018single}
S.~Zhang, L.~Wen, X.~Bian, Z.~Lei, and S.~Z. Li, ``Single-shot refinement
  neural network for object detection,'' in {\em Proceedings of the IEEE
  Conference on Computer Vision and Pattern Recognition}, pp.~4203--4212, 2018.

\bibitem{hu2018relation}
H.~Hu, J.~Gu, Z.~Zhang, J.~Dai, and Y.~Wei, ``Relation networks for object
  detection,'' in {\em Proceedings of the IEEE Conference on Computer Vision
  and Pattern Recognition}, pp.~3588--3597, 2018.

\bibitem{dai2017}
J.~Dai, H.~Qi, Y.~Xiong, Y.~Li, G.~Zhang, H.~Hu, and Y.~Wei, ``Deformable
  convolutional networks,'' in {\em Proceedings of the IEEE international
  conference on computer vision}, pp.~764--773, 2017.

\bibitem{zhu2018deformable}
X.~Zhu, H.~Hu, S.~Lin, and J.~Dai, ``Deformable convnets v2: More deformable,
  better results,'' {\em arXiv preprint arXiv:1811.11168}, 2018.

\bibitem{shrivastava2016training}
A.~Shrivastava, A.~Gupta, and R.~Girshick, ``Training region-based object
  detectors with online hard example mining,'' in {\em Proceedings of the IEEE
  Conference on Computer Vision and Pattern Recognition}, pp.~761--769, 2016.

\bibitem{bell2016inside}
S.~Bell, C.~Lawrence~Zitnick, K.~Bala, and R.~Girshick, ``Inside-outside net:
  Detecting objects in context with skip pooling and recurrent neural
  networks,'' in {\em Proceedings of the IEEE conference on computer vision and
  pattern recognition}, pp.~2874--2883, 2016.

\bibitem{22rfcn_dai2016r}
J.~Dai, Y.~Li, K.~He, and J.~Sun, ``R-fcn: Object detection via region-based
  fully convolutional networks,'' in {\em Advances in neural information
  processing systems}, pp.~379--387, 2016.

\bibitem{zhu2017couplenet}
Y.~Zhu, C.~Zhao, J.~Wang, X.~Zhao, Y.~Wu, and H.~Lu, ``Couplenet: Coupling
  global structure with local parts for object detection,'' in {\em Proceedings
  of the IEEE International Conference on Computer Vision}, pp.~4126--4134,
  2017.

\bibitem{huang2017speed}
J.~Huang, V.~Rathod, C.~Sun, M.~Zhu, A.~Korattikara, A.~Fathi, I.~Fischer,
  Z.~Wojna, Y.~Song, S.~Guadarrama, {\em et~al.}, ``Speed/accuracy trade-offs
  for modern convolutional object detectors,'' in {\em Proceedings of the IEEE
  conference on computer vision and pattern recognition}, pp.~7310--7311, 2017.

\bibitem{shrivastava2016beyond}
A.~Shrivastava, R.~Sukthankar, J.~Malik, and A.~Gupta, ``Beyond skip
  connections: Top-down modulation for object detection,'' {\em arXiv preprint
  arXiv:1612.06851}, 2016.

\bibitem{bodla2017soft}
N.~Bodla, B.~Singh, R.~Chellappa, and L.~S. Davis, ``Soft-nms--improving object
  detection with one line of code,'' in {\em Proceedings of the IEEE
  International Conference on Computer Vision}, pp.~5561--5569, 2017.

\bibitem{cai2018cascade}
Z.~Cai and N.~Vasconcelos, ``Cascade r-cnn: Delving into high quality object
  detection,'' in {\em Proceedings of the IEEE Conference on Computer Vision
  and Pattern Recognition}, pp.~6154--6162, 2018.

\bibitem{singh2018analysis}
B.~Singh and L.~S. Davis, ``An analysis of scale invariance in object detection
  snip,'' in {\em Proceedings of the IEEE conference on computer vision and
  pattern recognition}, pp.~3578--3587, 2018.

\bibitem{tychsen2018improving}
L.~Tychsen-Smith and L.~Petersson, ``Improving object localization with fitness
  nms and bounded iou loss,'' in {\em Proceedings of the IEEE Conference on
  Computer Vision and Pattern Recognition}, pp.~6877--6885, 2018.

\bibitem{kong2017ron}
T.~Kong, F.~Sun, A.~Yao, H.~Liu, M.~Lu, and Y.~Chen, ``Ron: Reverse connection
  with objectness prior networks for object detection,'' in {\em Proceedings of
  the IEEE Conference on Computer Vision and Pattern Recognition},
  pp.~5936--5944, 2017.

\bibitem{law2018cornernet}
H.~Law and J.~Deng, ``Cornernet: Detecting objects as paired keypoints,'' in
  {\em Proceedings of the European Conference on Computer Vision (ECCV)},
  pp.~734--750, 2018.

\bibitem{8634919}
M.~{Braun}, S.~{Krebs}, F.~{Flohr}, and D.~{Gavrila}, ``Eurocity persons: A
  novel benchmark for person detection in traffic scenes,'' {\em IEEE
  Transactions on Pattern Analysis and Machine Intelligence}, pp.~1--1, 2019.

\bibitem{zhang2017citypersons}
S.~Zhang, R.~Benenson, and B.~Schiele, ``Citypersons: A diverse dataset for
  pedestrian detection,'' in {\em Proceedings of the IEEE Conference on
  Computer Vision and Pattern Recognition}, pp.~3213--3221, 2017.

\bibitem{li2016new}
X.~Li, F.~Flohr, Y.~Yang, H.~Xiong, M.~Braun, S.~Pan, K.~Li, and D.~M. Gavrila,
  ``A new benchmark for vision-based cyclist detection,'' in {\em 2016 IEEE
  Intelligent Vehicles Symposium (IV)}, pp.~1028--1033, IEEE, 2016.

\bibitem{dalal2005histograms}
N.~Dalal and B.~Triggs, ``Histograms of oriented gradients for human
  detection,'' in {\em international Conference on computer vision \& Pattern
  Recognition (CVPR'05)}, vol.~1, pp.~886--893, IEEE Computer Society, 2005.

\bibitem{ess2007depth}
A.~Ess, B.~Leibe, and L.~Van~Gool, ``Depth and appearance for mobile scene
  analysis,'' in {\em 2007 IEEE 11th International Conference on Computer
  Vision}, pp.~1--8, IEEE, 2007.

\bibitem{wojek2009multi}
C.~Wojek, S.~Walk, and B.~Schiele, ``Multi-cue onboard pedestrian detection,''
  in {\em 2009 IEEE Conference on Computer Vision and Pattern Recognition},
  pp.~794--801, IEEE, 2009.

\bibitem{enzweiler2008monocular}
M.~Enzweiler and D.~M. Gavrila, ``Monocular pedestrian detection: Survey and
  experiments,'' {\em IEEE transactions on pattern analysis and machine
  intelligence}, vol.~31, no.~12, pp.~2179--2195, 2008.

\bibitem{goodfellow2014generative}
I.~Goodfellow, J.~Pouget-Abadie, M.~Mirza, B.~Xu, D.~Warde-Farley, S.~Ozair,
  A.~Courville, and Y.~Bengio, ``Generative adversarial nets,'' in {\em
  Advances in neural information processing systems}, pp.~2672--2680, 2014.

\bibitem{DBLP:journals/corr/abs-1906-11172}
B.~Zoph, E.~D. Cubuk, G.~Ghiasi, T.~Lin, J.~Shlens, and Q.~V. Le, ``Learning
  data augmentation strategies for object detection,'' {\em CoRR},
  vol.~abs/1906.11172, 2019.

\bibitem{tian2019fcos}
Z.~Tian, C.~Shen, H.~Chen, and T.~He, ``Fcos: Fully convolutional one-stage
  object detection,'' {\em arXiv preprint arXiv:1904.01355}, 2019.

\bibitem{kong2019foveabox}
T.~Kong, F.~Sun, H.~Liu, Y.~Jiang, and J.~Shi, ``Foveabox: Beyond anchor-based
  object detector,'' {\em arXiv preprint arXiv:1904.03797}, 2019.

\bibitem{pang2019libra}
J.~Pang, K.~Chen, J.~Shi, H.~Feng, W.~Ouyang, and D.~Lin, ``Libra r-cnn:
  Towards balanced learning for object detection,'' {\em arXiv preprint
  arXiv:1904.02701}, 2019.

\bibitem{kim2018parallel}
S.-W. Kim, H.-K. Kook, J.-Y. Sun, M.-C. Kang, and S.-J. Ko, ``Parallel feature
  pyramid network for object detection,'' in {\em Proceedings of the European
  Conference on Computer Vision (ECCV)}, pp.~234--250, 2018.

\bibitem{li2017fssd}
Z.~Li and F.~Zhou, ``Fssd: feature fusion single shot multibox detector,'' {\em
  arXiv preprint arXiv:1712.00960}, 2017.

\bibitem{chen2017weaving}
Y.~Chen, J.~Li, B.~Zhou, J.~Feng, and S.~Yan, ``Weaving multi-scale context for
  single shot detector,'' {\em arXiv preprint arXiv:1712.03149}, 2017.

\bibitem{zheng2018extend}
L.~Zheng, C.~Fu, and Y.~Zhao, ``Extend the shallow part of single shot multibox
  detector via convolutional neural network,'' in {\em Tenth International
  Conference on Digital Image Processing (ICDIP 2018)}, vol.~10806, p.~1080613,
  International Society for Optics and Photonics, 2018.

\bibitem{bae2019object}
S.-H. Bae, ``Object detection based on region decomposition and assembly,''
  {\em arXiv preprint arXiv:1901.08225}, 2019.

\bibitem{DBLP:journals/corr/abs-1711-05471}
E.~Barnea and O.~Ben{-}Shahar, ``On the utility of context (or the lack
  thereof) for object detection,'' {\em CoRR}, vol.~abs/1711.05471, 2017.

\bibitem{liu2018structure}
Y.~Liu, R.~Wang, S.~Shan, and X.~Chen, ``Structure inference net: Object
  detection using scene-level context and instance-level relationships,'' in
  {\em Proceedings of the IEEE Conference on Computer Vision and Pattern
  Recognition}, pp.~6985--6994, 2018.

\bibitem{singh2018sniper}
B.~Singh, M.~Najibi, and L.~S. Davis, ``Sniper: Efficient multi-scale
  training,'' in {\em Advances in Neural Information Processing Systems},
  pp.~9310--9320, 2018.

\bibitem{8371732}
K.~{Liang}, H.~{Chang}, B.~{Ma}, S.~{Shan}, and X.~{Chen}, ``Unifying visual
  attribute learning with object recognition in a multiplicative framework,''
  {\em IEEE Transactions on Pattern Analysis and Machine Intelligence},
  vol.~41, pp.~1747--1760, July 2019.

\bibitem{zhang2019object}
C.~Zhang and J.~Kim, ``Object detection with location-aware deformable
  convolution and backward attention filtering,'' in {\em Proceedings of the
  IEEE Conference on Computer Vision and Pattern Recognition}, pp.~9452--9461,
  2019.

\bibitem{yoo2015attentionnet}
D.~Yoo, S.~Park, J.-Y. Lee, A.~S. Paek, and I.~So~Kweon, ``Attentionnet:
  Aggregating weak directions for accurate object detection,'' in {\em
  Proceedings of the IEEE International Conference on Computer Vision},
  pp.~2659--2667, 2015.

\bibitem{xu2015show}
K.~Xu, J.~Ba, R.~Kiros, K.~Cho, A.~Courville, R.~Salakhutdinov, R.~Zemel, and
  Y.~Bengio, ``Show, attend and tell: Neural image caption generation with
  visual attention,'' {\em arXiv preprint arXiv:1502.03044}, 2015.

\bibitem{ba2014multiple}
J.~Ba, V.~Mnih, and K.~Kavukcuoglu, ``Multiple object recognition with visual
  attention,'' {\em arXiv preprint arXiv:1412.7755}, 2014.

\bibitem{li2019attention}
L.~Li, M.~Xu, X.~Wang, L.~Jiang, and H.~Liu, ``Attention based glaucoma
  detection: A large-scale database and cnn model,'' {\em arXiv preprint
  arXiv:1903.10831}, 2019.

\bibitem{hara2017attentional}
K.~Hara, M.-Y. Liu, O.~Tuzel, and A.-m. Farahmand, ``Attentional network for
  visual object detection,'' {\em arXiv preprint arXiv:1702.01478}, 2017.

\bibitem{DBLP:journals/corr/abs-1904-02874}
S.~Chaudhari, G.~Polatkan, R.~Ramanath, and V.~Mithal, ``An attentive survey of
  attention models,'' {\em CoRR}, vol.~abs/1904.02874, 2019.

\bibitem{kong2018deep}
T.~Kong, F.~Sun, C.~Tan, H.~Liu, and W.~Huang, ``Deep feature pyramid
  reconfiguration for object detection,'' in {\em Proceedings of the European
  Conference on Computer Vision (ECCV)}, pp.~169--185, 2018.

\bibitem{yu2016unitbox}
J.~Yu, Y.~Jiang, Z.~Wang, Z.~Cao, and T.~Huang, ``Unitbox: An advanced object
  detection network,'' in {\em Proceedings of the 24th ACM international
  conference on Multimedia}, pp.~516--520, ACM, 2016.

\bibitem{rezatofighi2019generalized}
H.~Rezatofighi, N.~Tsoi, J.~Gwak, A.~Sadeghian, I.~Reid, and S.~Savarese,
  ``Generalized intersection over union: A metric and a loss for bounding box
  regression,'' {\em arXiv preprint arXiv:1902.09630}, 2019.

\bibitem{he2019bounding}
Y.~He, C.~Zhu, J.~Wang, M.~Savvides, and X.~Zhang, ``Bounding box regression
  with uncertainty for accurate object detection,'' in {\em Proceedings of the
  IEEE Conference on Computer Vision and Pattern Recognition}, pp.~2888--2897,
  2019.

\bibitem{he2018softer}
Y.~He, X.~Zhang, M.~Savvides, and K.~Kitani, ``Softer-nms: Rethinking bounding
  box regression for accurate object detection,'' {\em arXiv preprint
  arXiv:1809.08545}, 2018.

\bibitem{naturecommunications1}
C.~Cabriel, N.~Bourg, P.~Jouchet, G.~Dupuis, C.~Leterrier, A.~Baron, M.-A.
  Badet-Denisot, B.~Vauzeilles, E.~Fort, and S.~L{\'e}v{\^e}que-Fort,
  ``Combining 3d single molecule localization strategies for reproducible
  bioimaging,'' {\em Nature Communications}, vol.~10, no.~1, p.~1980, 2019.

\bibitem{bucher2016hard}
M.~Bucher, S.~Herbin, and F.~Jurie, ``Hard negative mining for metric learning
  based zero-shot classification,'' in {\em European Conference on Computer
  Vision}, pp.~524--531, Springer, 2016.

\bibitem{yu2018loss}
H.~Yu, Z.~Zhang, Z.~Qin, H.~Wu, D.~Li, J.~Zhao, and X.~Lu, ``Loss rank mining:
  A general hard example mining method for real-time detectors,'' in {\em 2018
  International Joint Conference on Neural Networks (IJCNN)}, pp.~1--8, IEEE,
  2018.

\bibitem{chen2019towards}
K.~Chen, J.~Li, W.~Lin, J.~See, J.~Wang, L.~Duan, Z.~Chen, C.~He, and J.~Zou,
  ``Towards accurate one-stage object detection with ap-loss,'' {\em arXiv
  preprint arXiv:1904.06373}, 2019.

\bibitem{jiang2018acquisition}
B.~Jiang, R.~Luo, J.~Mao, T.~Xiao, and Y.~Jiang, ``Acquisition of localization
  confidence for accurate object detection,'' in {\em Proceedings of the
  European Conference on Computer Vision (ECCV)}, pp.~784--799, 2018.

\bibitem{liu2019adaptive}
S.~Liu, D.~Huang, and Y.~Wang, ``Adaptive nms: Refining pedestrian detection in
  a crowd,'' {\em arXiv preprint arXiv:1904.03629}, 2019.

\bibitem{hosang2016convnet}
J.~Hosang, R.~Benenson, and B.~Schiele, ``A convnet for non-maximum
  suppression,'' in {\em German Conference on Pattern Recognition},
  pp.~192--204, Springer, 2016.

\bibitem{hosang2017learning}
J.~Hosang, R.~Benenson, and B.~Schiele, ``Learning non-maximum suppression,''
  in {\em Proceedings of the IEEE Conference on Computer Vision and Pattern
  Recognition}, pp.~4507--4515, 2017.

\bibitem{jeong2017enhancement}
J.~Jeong, H.~Park, and N.~Kwak, ``Enhancement of ssd by concatenating feature
  maps for object detection,'' {\em arXiv preprint arXiv:1705.09587}, 2017.

\bibitem{xiang2018context}
W.~Xiang, D.-Q. Zhang, H.~Yu, and V.~Athitsos, ``Context-aware single-shot
  detector,'' in {\em 2018 IEEE Winter Conference on Applications of Computer
  Vision (WACV)}, pp.~1784--1793, IEEE, 2018.

\bibitem{cao2018feature}
G.~Cao, X.~Xie, W.~Yang, Q.~Liao, G.~Shi, and J.~Wu, ``Feature-fused ssd: fast
  detection for small objects,'' in {\em Ninth International Conference on
  Graphic and Image Processing (ICGIP 2017)}, vol.~10615, p.~106151E,
  International Society for Optics and Photonics, 2018.

\bibitem{li2017perceptual}
J.~Li, X.~Liang, Y.~Wei, T.~Xu, J.~Feng, and S.~Yan, ``Perceptual generative
  adversarial networks for small object detection,'' in {\em Proceedings of the
  IEEE Conference on Computer Vision and Pattern Recognition}, pp.~1222--1230,
  2017.

\bibitem{liu2018learning}
W.~Liu, S.~Liao, W.~Hu, X.~Liang, and X.~Chen, ``Learning efficient
  single-stage pedestrian detectors by asymptotic localization fitting,'' in
  {\em Proceedings of the European Conference on Computer Vision (ECCV)},
  pp.~618--634, 2018.

\bibitem{hu2017finding}
P.~Hu and D.~Ramanan, ``Finding tiny faces,'' in {\em Proceedings of the IEEE
  conference on computer vision and pattern recognition}, pp.~951--959, 2017.

\bibitem{xu2018mdssd}
M.~Xu, L.~Cui, P.~Lv, X.~Jiang, J.~Niu, B.~Zhou, and M.~Wang, ``Mdssd:
  Multi-scale deconvolutional single shot detector for small objects,'' {\em
  arXiv preprint arXiv:1805.07009}, 2018.

\bibitem{wang2017face}
J.~Wang, Y.~Yuan, and G.~Yu, ``Face attention network: an effective face
  detector for the occluded faces,'' {\em arXiv preprint arXiv:1711.07246},
  2017.

\bibitem{wang2018repulsion}
X.~Wang, T.~Xiao, Y.~Jiang, S.~Shao, J.~Sun, and C.~Shen, ``Repulsion loss:
  Detecting pedestrians in a crowd,'' in {\em Proceedings of the IEEE
  Conference on Computer Vision and Pattern Recognition}, pp.~7774--7783, 2018.

\bibitem{zhang2018occlusion}
S.~Zhang, L.~Wen, X.~Bian, Z.~Lei, and S.~Z. Li, ``Occlusion-aware r-cnn:
  detecting pedestrians in a crowd,'' in {\em Proceedings of the European
  Conference on Computer Vision (ECCV)}, pp.~637--653, 2018.

\bibitem{baque2017deep}
P.~Baqu{\'e}, F.~Fleuret, and P.~Fua, ``Deep occlusion reasoning for
  multi-camera multi-target detection,'' in {\em Proceedings of the IEEE
  International Conference on Computer Vision}, pp.~271--279, 2017.

\bibitem{he2015spatial}
K.~He, X.~Zhang, S.~Ren, and J.~Sun, ``Spatial pyramid pooling in deep
  convolutional networks for visual recognition,'' {\em IEEE transactions on
  pattern analysis and machine intelligence}, vol.~37, no.~9, pp.~1904--1916,
  2015.

\bibitem{sermanet2013overfeat}
P.~Sermanet, D.~Eigen, X.~Zhang, M.~Mathieu, R.~Fergus, and Y.~LeCun,
  ``Overfeat: Integrated recognition, localization and detection using
  convolutional networks,'' {\em arXiv preprint arXiv:1312.6229}, 2013.

\bibitem{felzenszwalb2009object}
P.~F. Felzenszwalb, R.~B. Girshick, D.~McAllester, and D.~Ramanan, ``Object
  detection with discriminatively trained part-based models,'' {\em IEEE
  transactions on pattern analysis and machine intelligence}, vol.~32, no.~9,
  pp.~1627--1645, 2009.

\bibitem{law2019cornernet}
H.~Law, Y.~Teng, O.~Russakovsky, and J.~Deng, ``Cornernet-lite: Efficient
  keypoint based object detection,'' {\em arXiv preprint arXiv:1904.08900},
  2019.

\bibitem{duan2019centernet}
K.~Duan, S.~Bai, L.~Xie, H.~Qi, Q.~Huang, and Q.~Tian, ``Centernet: Keypoint
  triplets for object detection,'' {\em arXiv preprint arXiv:1904.08189}, 2019.

\bibitem{wang2019region}
J.~Wang, K.~Chen, S.~Yang, C.~C. Loy, and D.~Lin, ``Region proposal by guided
  anchoring,'' {\em arXiv preprint arXiv:1901.03278}, 2019.

\bibitem{zhou2019bottom}
X.~Zhou, J.~Zhuo, and P.~Kr{\"a}henb{\"u}hl, ``Bottom-up object detection by
  grouping extreme and center points,'' {\em arXiv preprint arXiv:1901.08043},
  2019.

\bibitem{DBLP:journals/corr/abs-1904-07850}
X.~Zhou, D.~Wang, and P.~Kr{\"{a}}henb{\"{u}}hl, ``Objects as points,'' {\em
  CoRR}, vol.~abs/1904.07850, 2019.

\bibitem{chen2019dubox}
S.~Chen, J.~Li, C.~Yao, W.~Hou, S.~Qin, W.~Jin, and X.~Tang, ``Dubox: No-prior
  box objection detection via residual dual scale detectors,'' {\em arXiv
  preprint arXiv:1904.06883}, 2019.

\bibitem{zhu2019feature}
C.~Zhu, Y.~He, and M.~Savvides, ``Feature selective anchor-free module for
  single-shot object detection,'' {\em arXiv preprint arXiv:1903.00621}, 2019.

\bibitem{zhu2019scratchdet}
R.~Zhu, S.~Zhang, X.~Wang, L.~Wen, H.~Shi, L.~Bo, and T.~Mei, ``Scratchdet:
  Training single-shot object detectors from scratch,'' in {\em Proceedings of
  the IEEE conference on computer vision and pattern recognition}, 2019.

\bibitem{shen2017dsod}
Z.~Shen, Z.~Liu, J.~Li, Y.-G. Jiang, Y.~Chen, and X.~Xue, ``Dsod: Learning
  deeply supervised object detectors from scratch,'' in {\em Proceedings of the
  IEEE International Conference on Computer Vision}, pp.~1919--1927, 2017.

\bibitem{shen2017learning}
Z.~Shen, H.~Shi, R.~Feris, L.~Cao, S.~Yan, D.~Liu, X.~Wang, X.~Xue, and T.~S.
  Huang, ``Learning object detectors from scratch with gated recurrent feature
  pyramids,'' {\em arXiv preprint arXiv:1712.00886}, 2017.

\bibitem{li2018tiny}
Y.~Li, J.~Li, W.~Lin, and J.~Li, ``Tiny-dsod: Lightweight object detection for
  resource-restricted usages,'' {\em arXiv preprint arXiv:1807.11013}, 2018.

\bibitem{shen2018object}
Z.~Shen, Z.~Liu, J.~Li, Y.-G. Jiang, Y.~Chen, and X.~Xue, ``Object detection
  from scratch with deep supervision,'' {\em arXiv preprint arXiv:1809.09294},
  2018.

\bibitem{li2017light}
Z.~Li, C.~Peng, G.~Yu, X.~Zhang, Y.~Deng, and J.~Sun, ``Light-head r-cnn: In
  defense of two-stage object detector,'' {\em arXiv preprint
  arXiv:1711.07264}, 2017.

\bibitem{womg2018tiny}
A.~Womg, M.~J. Shafiee, F.~Li, and B.~Chwyl, ``Tiny ssd: A tiny single-shot
  detection deep convolutional neural network for real-time embedded object
  detection,'' in {\em 2018 15th Conference on Computer and Robot Vision
  (CRV)}, pp.~95--101, IEEE, 2018.

\bibitem{tychsen2017denet}
L.~Tychsen-Smith and L.~Petersson, ``Denet: Scalable real-time object detection
  with directed sparse sampling,'' in {\em Proceedings of the IEEE
  International Conference on Computer Vision}, pp.~428--436, 2017.

\bibitem{tripathi2017lcdet}
S.~Tripathi, G.~Dane, B.~Kang, V.~Bhaskaran, and T.~Nguyen, ``Lcdet:
  Low-complexity fully-convolutional neural networks for object detection in
  embedded systems,'' in {\em Proceedings of the IEEE Conference on Computer
  Vision and Pattern Recognition Workshops}, pp.~94--103, 2017.

\bibitem{lee2017wide}
Y.~Lee, H.~Kim, E.~Park, X.~Cui, and H.~Kim, ``Wide-residual-inception networks
  for real-time object detection,'' in {\em 2017 IEEE Intelligent Vehicles
  Symposium (IV)}, pp.~758--764, IEEE, 2017.

\bibitem{li2017mimicking}
Q.~Li, S.~Jin, and J.~Yan, ``Mimicking very efficient network for object
  detection,'' in {\em Proceedings of the IEEE Conference on Computer Vision
  and Pattern Recognition}, pp.~6356--6364, 2017.

\bibitem{zhou2017adaptive}
H.-Y. Zhou, B.-B. Gao, and J.~Wu, ``Adaptive feeding: Achieving fast and
  accurate detections by adaptively combining object detectors,'' in {\em
  Proceedings of the IEEE International Conference on Computer Vision},
  pp.~3505--3513, 2017.

\bibitem{liu2018receptive}
S.~Liu, D.~Huang, {\em et~al.}, ``Receptive field block net for accurate and
  fast object detection,'' in {\em Proceedings of the European Conference on
  Computer Vision (ECCV)}, pp.~385--400, 2018.

\bibitem{8170321}
R.~{Ranjan}, V.~M. {Patel}, and R.~{Chellappa}, ``Hyperface: A deep multi-task
  learning framework for face detection, landmark localization, pose
  estimation, and gender recognition,'' {\em IEEE Transactions on Pattern
  Analysis and Machine Intelligence}, vol.~41, pp.~121--135, Jan 2019.

\bibitem{8370677}
R.~{He}, X.~{Wu}, Z.~{Sun}, and T.~{Tan}, ``Wasserstein cnn: Learning invariant
  features for nir-vis face recognition,'' {\em IEEE Transactions on Pattern
  Analysis and Machine Intelligence}, vol.~41, pp.~1761--1773, July 2019.

\bibitem{zhang2019adacos}
X.~Zhang, R.~Zhao, Y.~Qiao, X.~Wang, and H.~Li, ``Adacos: Adaptively scaling
  cosine logits for effectively learning deep face representations,'' {\em
  arXiv preprint arXiv:1905.00292}, 2019.

\bibitem{liu2017rethinking}
Y.~Liu, H.~Li, and X.~Wang, ``Rethinking feature discrimination and
  polymerization for large-scale recognition,'' {\em arXiv preprint
  arXiv:1710.00870}, 2017.

\bibitem{ranjan2017l2}
R.~Ranjan, C.~D. Castillo, and R.~Chellappa, ``L2-constrained softmax loss for
  discriminative face verification,'' {\em arXiv preprint arXiv:1703.09507},
  2017.

\bibitem{wang2017normface}
F.~Wang, X.~Xiang, J.~Cheng, and A.~L. Yuille, ``Normface: l 2 hypersphere
  embedding for face verification,'' in {\em Proceedings of the 25th ACM
  international conference on Multimedia}, pp.~1041--1049, ACM, 2017.

\bibitem{deng2018arcface}
J.~Deng, J.~Guo, N.~Xue, and S.~Zafeiriou, ``Arcface: Additive angular margin
  loss for deep face recognition,'' {\em arXiv preprint arXiv:1801.07698},
  2018.

\bibitem{guo2017fuzzy}
Y.~Guo, L.~Jiao, S.~Wang, S.~Wang, and F.~Liu, ``Fuzzy sparse autoencoder
  framework for single image per person face recognition,'' {\em IEEE
  transactions on cybernetics}, vol.~48, no.~8, pp.~2402--2415, 2017.

\bibitem{wang2018deep}
M.~Wang and W.~Deng, ``Deep face recognition: A survey,'' {\em arXiv preprint
  arXiv:1804.06655}, 2018.

\bibitem{8686227}
Z.~{Cai}, M.~J. {Saberian}, and N.~{Vasconcelos}, ``Learning complexity-aware
  cascades for pedestrian detection,'' {\em IEEE Transactions on Pattern
  Analysis and Machine Intelligence}, pp.~1--1, 2019.

\bibitem{saberian2012learning}
M.~J. Saberian and N.~Vasconcelos, ``Learning optimal embedded cascades,'' {\em
  IEEE transactions on pattern analysis and machine intelligence}, vol.~34,
  no.~10, pp.~2005--2018, 2012.

\bibitem{dollar2014fast}
P.~Doll{\'a}r, R.~Appel, S.~Belongie, and P.~Perona, ``Fast feature pyramids
  for object detection,'' {\em IEEE transactions on pattern analysis and
  machine intelligence}, vol.~36, no.~8, pp.~1532--1545, 2014.

\bibitem{brunetti2018computer}
A.~Brunetti, D.~Buongiorno, G.~F. Trotta, and V.~Bevilacqua, ``Computer vision
  and deep learning techniques for pedestrian detection and tracking: A
  survey,'' {\em Neurocomputing}, vol.~300, pp.~17--33, 2018.

\bibitem{liu2013change}
S.~Liu, M.~Yamada, N.~Collier, and M.~Sugiyama, ``Change-point detection in
  time-series data by relative density-ratio estimation,'' {\em Neural
  Networks}, vol.~43, pp.~72--83, 2013.

\bibitem{senin2018grammarviz}
P.~Senin, J.~Lin, X.~Wang, T.~Oates, S.~Gandhi, A.~P. Boedihardjo, C.~Chen, and
  S.~Frankenstein, ``Grammarviz 3.0: Interactive discovery of variable-length
  time series patterns,'' {\em ACM Transactions on Knowledge Discovery from
  Data (TKDD)}, vol.~12, no.~1, p.~10, 2018.

\bibitem{jiang2015general}
M.~Jiang, A.~Beutel, P.~Cui, B.~Hooi, S.~Yang, and C.~Faloutsos, ``A general
  suspiciousness metric for dense blocks in multimodal data,'' in {\em 2015
  IEEE International Conference on Data Mining}, pp.~781--786, IEEE, 2015.

\bibitem{wu2008spatio}
E.~Wu, W.~Liu, and S.~Chawla, ``Spatio-temporal outlier detection in
  precipitation data,'' in {\em International Workshop on Knowledge Discovery
  from Sensor Data}, pp.~115--133, Springer, 2008.

\bibitem{8352745}
B.~{Barz}, E.~{Rodner}, Y.~G. {Garcia}, and J.~{Denzler}, ``Detecting regions
  of maximal divergence for spatio-temporal anomaly detection,'' {\em IEEE
  Transactions on Pattern Analysis and Machine Intelligence}, vol.~41,
  pp.~1088--1101, May 2019.

\bibitem{cheng2016learning}
G.~Cheng, P.~Zhou, and J.~Han, ``Learning rotation-invariant convolutional
  neural networks for object detection in vhr optical remote sensing images,''
  {\em IEEE Transactions on Geoscience and Remote Sensing}, vol.~54, no.~12,
  pp.~7405--7415, 2016.

\bibitem{zhang2019hierarchical}
Y.~Zhang, Y.~Yuan, Y.~Feng, and X.~Lu, ``Hierarchical and robust convolutional
  neural network for very high-resolution remote sensing object detection,''
  {\em IEEE Transactions on Geoscience and Remote Sensing}, 2019.

\bibitem{li2019r3}
Q.~Li, L.~Mou, Q.~Xu, Y.~Zhang, and X.~X. Zhu, ``R$^3$-net: A deep network for
  multioriented vehicle detection in aerial images and videos,'' {\em IEEE
  Transactions on Geoscience and Remote Sensing}, 2019.

\bibitem{deng2017toward}
Z.~Deng, H.~Sun, S.~Zhou, J.~Zhao, and H.~Zou, ``Toward fast and accurate
  vehicle detection in aerial images using coupled region-based convolutional
  neural networks,'' {\em IEEE Journal of Selected Topics in Applied Earth
  Observations and Remote Sensing}, vol.~10, no.~8, pp.~3652--3664, 2017.

\bibitem{audebert2017segment}
N.~Audebert, B.~Le~Saux, and S.~Lef{\`e}vre, ``Segment-before-detect: Vehicle
  detection and classification through semantic segmentation of aerial
  images,'' {\em Remote Sensing}, vol.~9, no.~4, p.~368, 2017.

\bibitem{li2018hsf}
Q.~Li, L.~Mou, Q.~Liu, Y.~Wang, and X.~X. Zhu, ``Hsf-net: Multiscale deep
  feature embedding for ship detection in optical remote sensing imagery,''
  {\em IEEE Transactions on Geoscience and Remote Sensing}, no.~99, pp.~1--15,
  2018.

\bibitem{pang2019r2}
J.~Pang, C.~Li, J.~Shi, Z.~Xu, and H.~Feng, ``R$^2$-cnn: Fast tiny object
  detection in large-scale remote sensing images,'' {\em IEEE Transactions on
  Geoscience and Remote Sensing}, 2019.

\bibitem{pei2017sar}
J.~Pei, Y.~Huang, W.~Huo, Y.~Zhang, J.~Yang, and T.-S. Yeo, ``Sar automatic
  target recognition based on multiview deep learning framework,'' {\em IEEE
  Transactions on Geoscience and Remote Sensing}, vol.~56, no.~4,
  pp.~2196--2210, 2017.

\bibitem{long2017accurate}
Y.~Long, Y.~Gong, Z.~Xiao, and Q.~Liu, ``Accurate object localization in remote
  sensing images based on convolutional neural networks,'' {\em IEEE
  Transactions on Geoscience and Remote Sensing}, vol.~55, no.~5,
  pp.~2486--2498, 2017.

\bibitem{shahzad2018buildings}
M.~Shahzad, M.~Maurer, F.~Fraundorfer, Y.~Wang, and X.~X. Zhu, ``Buildings
  detection in vhr sar images using fully convolution neural networks,'' {\em
  IEEE transactions on geoscience and remote sensing}, vol.~57, no.~2,
  pp.~1100--1116, 2018.

\bibitem{zhang2016weakly}
F.~Zhang, B.~Du, L.~Zhang, and M.~Xu, ``Weakly supervised learning based on
  coupled convolutional neural networks for aircraft detection,'' {\em IEEE
  Transactions on Geoscience and Remote Sensing}, vol.~54, no.~9,
  pp.~5553--5563, 2016.

\bibitem{han2014object}
J.~Han, D.~Zhang, G.~Cheng, L.~Guo, and J.~Ren, ``Object detection in optical
  remote sensing images based on weakly supervised learning and high-level
  feature learning,'' {\em IEEE Transactions on Geoscience and Remote Sensing},
  vol.~53, no.~6, pp.~3325--3337, 2014.

\bibitem{li2018hough}
Q.~Li, Y.~Wang, Q.~Liu, and W.~Wang, ``Hough transform guided deep feature
  extraction for dense building detection in remote sensing images,'' in {\em
  2018 IEEE International Conference on Acoustics, Speech and Signal Processing
  (ICASSP)}, pp.~1872--1876, IEEE, 2018.

\bibitem{mou2018vehicle}
L.~Mou and X.~X. Zhu, ``Vehicle instance segmentation from aerial image and
  video using a multitask learning residual fully convolutional network,'' {\em
  IEEE Transactions on Geoscience and Remote Sensing}, no.~99, pp.~1--13, 2018.

\bibitem{chen2014vehicle}
X.~Chen, S.~Xiang, C.-L. Liu, and C.-H. Pan, ``Vehicle detection in satellite
  images by hybrid deep convolutional neural networks,'' {\em IEEE Geoscience
  and remote sensing letters}, vol.~11, no.~10, pp.~1797--1801, 2014.

\bibitem{ammour2017deep}
N.~Ammour, H.~Alhichri, Y.~Bazi, B.~Benjdira, N.~Alajlan, and M.~Zuair, ``Deep
  learning approach for car detection in uav imagery,'' {\em Remote Sensing},
  vol.~9, no.~4, p.~312, 2017.

\bibitem{wang2016new}
S.~Wang, M.~Wang, S.~Yang, and L.~Jiao, ``New hierarchical saliency filtering
  for fast ship detection in high-resolution sar images,'' {\em IEEE
  Transactions on Geoscience and Remote Sensing}, vol.~55, no.~1, pp.~351--362,
  2016.

\bibitem{ma2019novel}
W.~Ma, Q.~Guo, Y.~Wu, W.~Zhao, X.~Zhang, and L.~Jiao, ``A novel multi-model
  decision fusion network for object detection in remote sensing images,'' {\em
  Remote Sensing}, vol.~11, no.~7, p.~737, 2019.

\bibitem{dong2019sig}
R.~Dong, D.~Xu, J.~Zhao, L.~Jiao, and J.~An, ``Sig-nms-based faster r-cnn
  combining transfer learning for small target detection in vhr optical remote
  sensing imagery,'' {\em IEEE Transactions on Geoscience and Remote Sensing},
  2019.

\bibitem{chen2019deep}
C.~Chen, C.~He, C.~Hu, H.~Pei, and L.~Jiao, ``A deep neural network based on an
  attention mechanism for sar ship detection in multiscale and complex
  scenarios,'' {\em IEEE Access}, 2019.

\bibitem{zhu2019multiscale}
H.~Zhu, P.~Zhang, L.~Wang, X.~Zhang, and L.~Jiao, ``A multiscale object
  detection approach for remote sensing images based on mse-densenet and the
  dynamic anchor assignment,'' {\em Remote Sensing Letters}, vol.~10, no.~10,
  pp.~959--967, 2019.

\bibitem{cheng2014multi}
G.~Cheng, J.~Han, P.~Zhou, and L.~Guo, ``Multi-class geospatial object
  detection and geographic image classification based on collection of part
  detectors,'' {\em ISPRS Journal of Photogrammetry and Remote Sensing},
  vol.~98, pp.~119--132, 2014.

\bibitem{xia2018dota}
G.-S. Xia, X.~Bai, J.~Ding, Z.~Zhu, S.~Belongie, J.~Luo, M.~Datcu, M.~Pelillo,
  and L.~Zhang, ``Dota: A large-scale dataset for object detection in aerial
  images,'' in {\em Proceedings of the IEEE Conference on Computer Vision and
  Pattern Recognition}, pp.~3974--3983, 2018.

\bibitem{liu2015fast}
K.~Liu and G.~Mattyus, ``Fast multiclass vehicle detection on aerial images,''
  {\em IEEE Geoscience and Remote Sensing Letters}, vol.~12, no.~9,
  pp.~1938--1942, 2015.

\bibitem{razakarivony2016vehicle}
S.~Razakarivony and F.~Jurie, ``Vehicle detection in aerial imagery: A small
  target detection benchmark,'' {\em Journal of Visual Communication and Image
  Representation}, vol.~34, pp.~187--203, 2016.

\bibitem{cheng2016survey}
G.~Cheng and J.~Han, ``A survey on object detection in optical remote sensing
  images,'' {\em ISPRS Journal of Photogrammetry and Remote Sensing}, vol.~117,
  pp.~11--28, 2016.

\bibitem{shivakumara2018cnn}
P.~Shivakumara, D.~Tang, M.~Asadzadehkaljahi, T.~Lu, U.~Pal, and M.~H. Anisi,
  ``Cnn-rnn based method for license plate recognition,'' {\em CAAI
  Transactions on Intelligence Technology}, vol.~3, no.~3, pp.~169--175, 2018.

\bibitem{sarfraz2019approach}
M.~Sarfraz and M.~J. Ahmed, ``An approach to license plate recognition system
  using neural network,'' in {\em Exploring Critical Approaches of Evolutionary
  Computation}, pp.~20--36, IGI Global, 2019.

\bibitem{li2018toward}
H.~Li, P.~Wang, and C.~Shen, ``Toward end-to-end car license plate detection
  and recognition with deep neural networks,'' {\em IEEE Transactions on
  Intelligent Transportation Systems}, no.~99, pp.~1--11, 2018.

\bibitem{qian2018fast}
J.~Qian and B.~Qu, ``Fast license plate recognition method based on competitive
  neural network,'' in {\em 2018 3rd International Conference on
  Communications, Information Management and Network Security (CIMNS 2018)},
  Atlantis Press, 2018.

\bibitem{laroca2018robust}
R.~Laroca, E.~Severo, L.~A. Zanlorensi, L.~S. Oliveira, G.~R. Gon{\c{c}}alves,
  W.~R. Schwartz, and D.~Menotti, ``A robust real-time automatic license plate
  recognition based on the yolo detector,'' in {\em 2018 International Joint
  Conference on Neural Networks (IJCNN)}, pp.~1--10, IEEE, 2018.

\bibitem{nair2018survey}
A.~S. Nair, S.~Raju, K.~Harikrishnan, and A.~Mathew, ``A survey of techniques
  for license plate detection and recognition,'' {\em i-manager's Journal on
  Image Processing}, vol.~5, no.~1, p.~25, 2018.

\bibitem{lu2019l3}
W.~Lu, Y.~Zhou, G.~Wan, S.~Hou, S.~Song, and B.~A. D. B.~U. ADU, ``L3-net:
  Towards learning based lidar localization for autonomous driving,'' in {\em
  Proceedings of the IEEE Conference on Computer Vision and Pattern
  Recognition}, pp.~6389--6398, 2019.

\bibitem{song2018apollocar3d}
X.~Song, P.~Wang, D.~Zhou, R.~Zhu, C.~Guan, Y.~Dai, H.~Su, H.~Li, and R.~Yang,
  ``Apollocar3d: A large 3d car instance understanding benchmark for autonomous
  driving,'' {\em arXiv preprint arXiv:1811.12222}, 2018.

\bibitem{banerjee2018online}
K.~Banerjee, D.~Notz, J.~Windelen, S.~Gavarraju, and M.~He, ``Online camera
  lidar fusion and object detection on hybrid data for autonomous driving,'' in
  {\em 2018 IEEE Intelligent Vehicles Symposium (IV)}, pp.~1632--1638, IEEE,
  2018.

\bibitem{arnold2019survey}
E.~Arnold, O.~Y. Al-Jarrah, M.~Dianati, S.~Fallah, D.~Oxtoby, and
  A.~Mouzakitis, ``A survey on 3d object detection methods for autonomous
  driving applications,'' {\em IEEE Transactions on Intelligent Transportation
  Systems}, 2019.

\bibitem{li2018real}
J.~Li and Z.~Wang, ``Real-time traffic sign recognition based on efficient cnns
  in the wild,'' {\em IEEE Transactions on Intelligent Transportation Systems},
  no.~99, pp.~1--10, 2018.

\bibitem{moritani2018traffic}
T.~Moritani, Y.~Otsubo, and T.~Arinaga, ``Traffic sign recognition system,''
  Jan.~9 2018.
\newblock US Patent 9,865,165.

\bibitem{khalid2018automatic}
S.~Khalid, N.~Muhammad, and M.~Sharif, ``Automatic measurement of the traffic
  sign with digital segmentation and recognition,'' {\em IET Intelligent
  Transport Systems}, vol.~13, no.~2, pp.~269--279, 2018.

\bibitem{arcos2018deep}
{\'A}.~Arcos-Garc{\'\i}a, J.~A. {\'A}lvarez-Garc{\'\i}a, and L.~M.
  Soria-Morillo, ``Deep neural network for traffic sign recognition systems: An
  analysis of spatial transformers and stochastic optimisation methods,'' {\em
  Neural Networks}, vol.~99, pp.~158--165, 2018.

\bibitem{li2018deepsign}
D.~Li, D.~Zhao, Y.~Chen, and Q.~Zhang, ``Deepsign: Deep learning based traffic
  sign recognition,'' in {\em 2018 international joint conference on neural
  networks (IJCNN)}, pp.~1--6, IEEE, 2018.

\bibitem{wu2018traffic}
B.-X. Wu, P.-Y. Wang, Y.-T. Yang, and J.-I. Guo, ``Traffic sign recognition
  with light convolutional networks,'' in {\em 2018 IEEE International
  Conference on Consumer Electronics-Taiwan (ICCE-TW)}, pp.~1--2, IEEE, 2018.

\bibitem{zhou2018improved}
S.~Zhou, W.~Liang, J.~Li, and J.-U. Kim, ``Improved vgg model for road traffic
  sign recognition,'' {\em Computers, Materials and Continua}, vol.~57,
  pp.~11--24, 2018.

\bibitem{li2019clu}
Z.~Li, M.~Dong, S.~Wen, X.~Hu, P.~Zhou, and Z.~Zeng, ``Clu-cnns: Object
  detection for medical images,'' {\em Neurocomputing}, vol.~350, pp.~53--59,
  2019.

\bibitem{naturecommunications2}
Q.~Liu, L.~Fang, G.~Yu, D.~Wang, C.-L. Xiao, and K.~Wang, ``Detection of dna
  base modifications by deep recurrent neural network on oxford nanopore
  sequencing data,'' {\em Nature Communications}, vol.~10, no.~1, p.~2449,
  2019.

\bibitem{naturecommunications3}
P.~J. Schubert, S.~Dorkenwald, M.~Januszewski, V.~Jain, and J.~Kornfeld,
  ``Learning cellular morphology with neural networks,'' {\em Nature
  Communications}, vol.~10, no.~1, p.~2736, 2019.

\bibitem{codella2018skin}
N.~C. Codella, D.~Gutman, M.~E. Celebi, B.~Helba, M.~A. Marchetti, S.~W. Dusza,
  A.~Kalloo, K.~Liopyris, N.~Mishra, H.~Kittler, {\em et~al.}, ``Skin lesion
  analysis toward melanoma detection: A challenge at the 2017 international
  symposium on biomedical imaging (isbi), hosted by the international skin
  imaging collaboration (isic),'' in {\em 2018 IEEE 15th International
  Symposium on Biomedical Imaging (ISBI 2018)}, pp.~168--172, IEEE, 2018.

\bibitem{naji2018survey}
S.~Naji, H.~A. Jalab, and S.~A. Kareem, ``A survey on skin detection in colored
  images,'' {\em Artificial Intelligence Review}, pp.~1--47, 2018.

\bibitem{altaf2019going}
F.~Altaf, S.~Islam, N.~Akhtar, and N.~K. Janjua, ``Going deep in medical image
  analysis: Concepts, methods, challenges and future directions,'' {\em arXiv
  preprint arXiv:1902.05655}, 2019.

\bibitem{DBLP:journals/corr/abs-1904-00853}
E.~Goldman, R.~Herzig, A.~Eisenschtat, O.~Ratzon, I.~Levi, J.~Goldberger, and
  T.~Hassner, ``Precise detection in densely packed scenes,'' {\em CoRR},
  vol.~abs/1904.00853, 2019.

\bibitem{8618422}
Z.~{Yang}, Q.~{Li}, L.~{Wenyin}, and J.~{Lv}, ``Shared multi-view data
  representation for multi-domain event detection,'' {\em IEEE Transactions on
  Pattern Analysis and Machine Intelligence}, pp.~1--1, 2019.

\bibitem{wang2012social}
Y.~Wang, H.~Sundaram, and L.~Xie, ``Social event detection with interaction
  graph modeling,'' in {\em Proceedings of the 20th ACM international
  conference on Multimedia}, pp.~865--868, ACM, 2012.

\bibitem{schinas2015multimodal}
M.~Schinas, S.~Papadopoulos, G.~Petkos, Y.~Kompatsiaris, and P.~A. Mitkas,
  ``Multimodal graph-based event detection and summarization in social media
  streams,'' in {\em Proceedings of the 23rd ACM international conference on
  Multimedia}, pp.~189--192, ACM, 2015.

\bibitem{hasan2018survey}
M.~Hasan, M.~A. Orgun, and R.~Schwitter, ``A survey on real-time event
  detection from the twitter data stream,'' {\em Journal of Information
  Science}, vol.~44, no.~4, pp.~443--463, 2018.

\bibitem{teboul2011shape}
O.~Teboul, I.~Kokkinos, L.~Simon, P.~Koutsourakis, and N.~Paragios, ``Shape
  grammar parsing via reinforcement learning,'' in {\em CVPR 2011},
  pp.~2273--2280, IEEE, 2011.

\bibitem{zhao2010rectilinear}
P.~Zhao, T.~Fang, J.~Xiao, H.~Zhang, Q.~Zhao, and L.~Quan, ``Rectilinear
  parsing of architecture in urban environment,'' in {\em 2010 IEEE Computer
  Society Conference on Computer Vision and Pattern Recognition}, pp.~342--349,
  IEEE, 2010.

\bibitem{friedman2013online}
S.~Friedman and I.~Stamos, ``Online detection of repeated structures in point
  clouds of urban scenes for compression and registration,'' {\em International
  journal of computer vision}, vol.~102, no.~1-3, pp.~112--128, 2013.

\bibitem{shen2011adaptive}
C.-H. Shen, S.-S. Huang, H.~Fu, and S.-M. Hu, ``Adaptive partitioning of urban
  facades,'' in {\em ACM Transactions on Graphics (TOG)}, vol.~30, p.~184, ACM,
  2011.

\bibitem{schindler2008detecting}
G.~Schindler, P.~Krishnamurthy, R.~Lublinerman, Y.~Liu, and F.~Dellaert,
  ``Detecting and matching repeated patterns for automatic geo-tagging in urban
  environments,'' in {\em 2008 IEEE Conference on Computer Vision and Pattern
  Recognition}, pp.~1--7, IEEE, 2008.

\bibitem{wu2010detecting}
C.~Wu, J.-M. Frahm, and M.~Pollefeys, ``Detecting large repetitive structures
  with salient boundaries,'' in {\em European conference on computer vision},
  pp.~142--155, Springer, 2010.

\bibitem{muller2007image}
P.~M{\"u}ller, G.~Zeng, P.~Wonka, and L.~Van~Gool, ``Image-based procedural
  modeling of facades,'' in {\em ACM Transactions on Graphics (TOG)}, vol.~26,
  p.~85, ACM, 2007.

\bibitem{barinova2010geometric}
O.~Barinova, V.~Lempitsky, E.~Tretiak, and P.~Kohli, ``Geometric image parsing
  in man-made environments,'' in {\em European conference on computer vision},
  pp.~57--70, Springer, 2010.

\bibitem{kozinski2015mrf}
M.~Kozinski, R.~Gadde, S.~Zagoruyko, G.~Obozinski, and R.~Marlet, ``A mrf shape
  prior for facade parsing with occlusions,'' in {\em Proceedings of the IEEE
  Conference on Computer Vision and Pattern Recognition}, pp.~2820--2828, 2015.

\bibitem{cohen2014efficient}
A.~Cohen, A.~G. Schwing, and M.~Pollefeys, ``Efficient structured parsing of
  facades using dynamic programming,'' in {\em Proceedings of the IEEE
  Conference on Computer Vision and Pattern Recognition}, pp.~3206--3213, 2014.

\bibitem{gandy2011tensor}
S.~Gandy, B.~Recht, and I.~Yamada, ``Tensor completion and low-n-rank tensor
  recovery via convex optimization,'' {\em Inverse Problems}, vol.~27, no.~2,
  p.~025010, 2011.

\bibitem{candes2011robust}
E.~J. Cand{\`e}s, X.~Li, Y.~Ma, and J.~Wright, ``Robust principal component
  analysis?,'' {\em Journal of the ACM (JACM)}, vol.~58, no.~3, p.~11, 2011.

\bibitem{liu2012tensor}
J.~Liu, P.~Musialski, P.~Wonka, and J.~Ye, ``Tensor completion for estimating
  missing values in visual data,'' {\em IEEE transactions on pattern analysis
  and machine intelligence}, vol.~35, no.~1, pp.~208--220, 2012.

\bibitem{8417901}
J.~{Liu}, E.~{Psarakis}, Y.~{Feng}, and I.~{Stamos}, ``A kronecker product
  model for repeated pattern detection on 2d urban images,'' {\em IEEE
  Transactions on Pattern Analysis and Machine Intelligence}, pp.~1--1, 2018.

\bibitem{anderson2018bottom}
P.~Anderson, X.~He, C.~Buehler, D.~Teney, M.~Johnson, S.~Gould, and L.~Zhang,
  ``Bottom-up and top-down attention for image captioning and visual question
  answering,'' in {\em Proceedings of the IEEE Conference on Computer Vision
  and Pattern Recognition}, pp.~6077--6086, 2018.

\bibitem{vinyals2015show}
O.~Vinyals, A.~Toshev, S.~Bengio, and D.~Erhan, ``Show and tell: A neural image
  caption generator,'' in {\em Proceedings of the IEEE conference on computer
  vision and pattern recognition}, pp.~3156--3164, 2015.

\bibitem{gu2018stack}
J.~Gu, J.~Cai, G.~Wang, and T.~Chen, ``Stack-captioning: Coarse-to-fine
  learning for image captioning,'' in {\em Thirty-Second AAAI Conference on
  Artificial Intelligence}, 2018.

\bibitem{yao2018exploring}
T.~Yao, Y.~Pan, Y.~Li, and T.~Mei, ``Exploring visual relationship for image
  captioning,'' in {\em Proceedings of the European Conference on Computer
  Vision (ECCV)}, pp.~684--699, 2018.

\bibitem{aneja2018convolutional}
J.~Aneja, A.~Deshpande, and A.~G. Schwing, ``Convolutional image captioning,''
  in {\em Proceedings of the IEEE Conference on Computer Vision and Pattern
  Recognition}, pp.~5561--5570, 2018.

\bibitem{bai2018survey}
S.~Bai and S.~An, ``A survey on automatic image caption generation,'' {\em
  Neurocomputing}, vol.~311, pp.~291--304, 2018.

\bibitem{8627954}
W.~{Yang}, R.~T. {Tan}, J.~{Feng}, J.~{Liu}, S.~{Yan}, and Z.~{Guo}, ``Joint
  rain detection and removal from a single image with contextualized deep
  networks,'' {\em IEEE Transactions on Pattern Analysis and Machine
  Intelligence}, pp.~1--1, 2019.

\bibitem{8723605}
X.~{Hu}, C.~{Fu}, L.~{Zhu}, J.~{Qin}, and P.~{Heng}, ``Direction-aware spatial
  context features for shadow detection and removal,'' {\em IEEE Transactions
  on Pattern Analysis and Machine Intelligence}, pp.~1--1, 2019.

\bibitem{waldchen2018machine}
J.~W{\"a}ldchen and P.~M{\"a}der, ``Machine learning for image based species
  identification,'' {\em Methods in Ecology and Evolution}, vol.~9, no.~11,
  pp.~2216--2225, 2018.

\bibitem{bilen2016weakly}
H.~Bilen and A.~Vedaldi, ``Weakly supervised deep detection networks,'' in {\em
  Proceedings of the IEEE Conference on Computer Vision and Pattern
  Recognition}, pp.~2846--2854, 2016.

\bibitem{kantorov2016contextlocnet}
V.~Kantorov, M.~Oquab, M.~Cho, and I.~Laptev, ``Contextlocnet: Context-aware
  deep network models for weakly supervised localization,'' in {\em European
  Conference on Computer Vision}, pp.~350--365, Springer, 2016.

\bibitem{tang2017multiple}
P.~Tang, X.~Wang, X.~Bai, and W.~Liu, ``Multiple instance detection network
  with online instance classifier refinement,'' in {\em Proceedings of the IEEE
  Conference on Computer Vision and Pattern Recognition}, pp.~2843--2851, 2017.

\bibitem{diba2017weakly}
A.~Diba, V.~Sharma, A.~Pazandeh, H.~Pirsiavash, and L.~Van~Gool, ``Weakly
  supervised cascaded convolutional networks,'' in {\em Proceedings of the IEEE
  conference on computer vision and pattern recognition}, pp.~914--922, 2017.

\bibitem{li2016image}
Y.~Li, L.~Liu, C.~Shen, and A.~van~den Hengel, ``Image co-localization by
  mimicking a good detector’s confidence score distribution,'' in {\em
  European Conference on Computer Vision}, pp.~19--34, Springer, 2016.

\bibitem{yang2019activity}
Z.~Yang, D.~Mahajan, D.~Ghadiyaram, R.~Nevatia, and V.~Ramanathan, ``Activity
  driven weakly supervised object detection,'' {\em arXiv preprint
  arXiv:1904.01665}, 2019.

\bibitem{8640243}
F.~{Wan}, P.~{Wei}, Z.~{Han}, J.~{Jiao}, and Q.~{Ye}, ``Min-entropy latent
  model for weakly supervised object detection,'' {\em IEEE Transactions on
  Pattern Analysis and Machine Intelligence}, pp.~1--1, 2019.

\bibitem{8493315}
P.~{Tang}, X.~{Wang}, S.~{Bai}, W.~{Shen}, X.~{Bai}, W.~{Liu}, and A.~L.
  {Yuille}, ``Pcl: Proposal cluster learning for weakly supervised object
  detection,'' {\em IEEE Transactions on Pattern Analysis and Machine
  Intelligence}, pp.~1--1, 2018.

\bibitem{8370896}
C.~{Cao}, Y.~{Huang}, Y.~{Yang}, L.~{Wang}, Z.~{Wang}, and T.~{Tan}, ``Feedback
  convolutional neural network for visual localization and segmentation,'' {\em
  IEEE Transactions on Pattern Analysis and Machine Intelligence}, vol.~41,
  pp.~1627--1640, July 2019.

\bibitem{wan2019c}
F.~Wan, C.~Liu, W.~Ke, X.~Ji, J.~Jiao, and Q.~Ye, ``C-mil: Continuation
  multiple instance learning for weakly supervised object detection,'' {\em
  arXiv preprint arXiv:1904.05647}, 2019.

\bibitem{wu2019cascaded}
Z.~Wu, L.~Su, and Q.~Huang, ``Cascaded partial decoder for fast and accurate
  salient object detection,'' {\em arXiv preprint arXiv:1904.08739}, 2019.

\bibitem{liu2019simple}
J.-J. Liu, Q.~Hou, M.-M. Cheng, J.~Feng, and J.~Jiang, ``A simple pooling-based
  design for real-time salient object detection,'' {\em arXiv preprint
  arXiv:1904.09569}, 2019.

\bibitem{8668551}
W.~{Wang}, J.~{Shen}, X.~{Dong}, A.~{Borji}, and R.~{Yang}, ``Inferring salient
  objects from human fixations,'' {\em IEEE Transactions on Pattern Analysis
  and Machine Intelligence}, pp.~1--1, 2019.

\bibitem{8382302}
L.~{Wang}, L.~{Wang}, H.~{Lu}, P.~{Zhang}, and X.~{Ruan}, ``Salient object
  detection with recurrent fully convolutional networks,'' {\em IEEE
  Transactions on Pattern Analysis and Machine Intelligence}, vol.~41,
  pp.~1734--1746, July 2019.

\bibitem{feng2019attentive}
M.~Feng, H.~Lu, and E.~Ding, ``Attentive feedback network for boundary-aware
  salient object detection,'' in {\em Proceedings of the IEEE Conference on
  Computer Vision and Pattern Recognition}, pp.~1623--1632, 2019.

\bibitem{fan2019shifting}
D.-P. Fan, W.~Wang, M.-M. Cheng, and J.~Shen, ``Shifting more attention to
  video salient object detection,'' in {\em Proceedings of the IEEE Conference
  on Computer Vision and Pattern Recognition}, pp.~8554--8564, 2019.

\bibitem{kim2015spatiotemporal}
H.~Kim, Y.~Kim, J.-Y. Sim, and C.-S. Kim, ``Spatiotemporal saliency detection
  for video sequences based on random walk with restart,'' {\em IEEE
  Transactions on Image Processing}, vol.~24, no.~8, pp.~2552--2564, 2015.

\bibitem{li2013video}
F.~Li, T.~Kim, A.~Humayun, D.~Tsai, and J.~M. Rehg, ``Video segmentation by
  tracking many figure-ground segments,'' in {\em Proceedings of the IEEE
  International Conference on Computer Vision}, pp.~2192--2199, 2013.

\bibitem{li2017benchmark}
J.~Li, C.~Xia, and X.~Chen, ``A benchmark dataset and saliency-guided stacked
  autoencoders for video-based salient object detection,'' {\em IEEE
  Transactions on Image Processing}, vol.~27, no.~1, pp.~349--364, 2017.

\bibitem{liu2016saliency}
Z.~Liu, J.~Li, L.~Ye, G.~Sun, and L.~Shen, ``Saliency detection for
  unconstrained videos using superpixel-level graph and spatiotemporal
  propagation,'' {\em IEEE transactions on circuits and systems for video
  technology}, vol.~27, no.~12, pp.~2527--2542, 2016.

\bibitem{ochs2013segmentation}
P.~Ochs, J.~Malik, and T.~Brox, ``Segmentation of moving objects by long term
  video analysis,'' {\em IEEE transactions on pattern analysis and machine
  intelligence}, vol.~36, no.~6, pp.~1187--1200, 2013.

\bibitem{wang2015consistent}
W.~Wang, J.~Shen, and L.~Shao, ``Consistent video saliency using local gradient
  flow optimization and global refinement,'' {\em IEEE Transactions on Image
  Processing}, vol.~24, no.~11, pp.~4185--4196, 2015.

\bibitem{chen2017video}
C.~Chen, S.~Li, Y.~Wang, H.~Qin, and A.~Hao, ``Video saliency detection via
  spatial-temporal fusion and low-rank coherency diffusion,'' {\em IEEE
  Transactions on Image Processing}, vol.~26, no.~7, pp.~3156--3170, 2017.

\bibitem{chen2018scom}
Y.~Chen, W.~Zou, Y.~Tang, X.~Li, C.~Xu, and N.~Komodakis, ``Scom:
  Spatiotemporal constrained optimization for salient object detection,'' {\em
  IEEE Transactions on Image Processing}, vol.~27, no.~7, pp.~3345--3357, 2018.

\bibitem{li2018unsupervised}
S.~Li, B.~Seybold, A.~Vorobyov, X.~Lei, and C.-C. Jay~Kuo, ``Unsupervised video
  object segmentation with motion-based bilateral networks,'' in {\em
  Proceedings of the European Conference on Computer Vision (ECCV)},
  pp.~207--223, 2018.

\bibitem{liu2014superpixel}
Z.~Liu, X.~Zhang, S.~Luo, and O.~Le~Meur, ``Superpixel-based spatiotemporal
  saliency detection,'' {\em IEEE transactions on circuits and systems for
  video technology}, vol.~24, no.~9, pp.~1522--1540, 2014.

\bibitem{song2018pyramid}
H.~Song, W.~Wang, S.~Zhao, J.~Shen, and K.-M. Lam, ``Pyramid dilated deeper
  convlstm for video salient object detection,'' in {\em Proceedings of the
  European Conference on Computer Vision (ECCV)}, pp.~715--731, 2018.

\bibitem{tang2018weakly}
Y.~Tang, W.~Zou, Z.~Jin, Y.~Chen, Y.~Hua, and X.~Li, ``Weakly supervised
  salient object detection with spatiotemporal cascade neural networks,'' {\em
  IEEE Transactions on Circuits and Systems for Video Technology}, 2018.

\bibitem{wang2017video}
W.~Wang, J.~Shen, and L.~Shao, ``Video salient object detection via fully
  convolutional networks,'' {\em IEEE Transactions on Image Processing},
  vol.~27, no.~1, pp.~38--49, 2017.

\bibitem{tu2016real}
W.-C. Tu, S.~He, Q.~Yang, and S.-Y. Chien, ``Real-time salient object detection
  with a minimum spanning tree,'' in {\em Proceedings of the IEEE Conference on
  Computer Vision and Pattern Recognition}, pp.~2334--2342, 2016.

\bibitem{wang2015saliency}
W.~Wang, J.~Shen, and F.~Porikli, ``Saliency-aware geodesic video object
  segmentation,'' in {\em Proceedings of the IEEE conference on computer vision
  and pattern recognition}, pp.~3395--3402, 2015.

\bibitem{xi2016salient}
T.~Xi, W.~Zhao, H.~Wang, and W.~Lin, ``Salient object detection with
  spatiotemporal background priors for video,'' {\em IEEE Transactions on Image
  Processing}, vol.~26, no.~7, pp.~3425--3436, 2016.

\bibitem{zhang2015minimum}
J.~Zhang, S.~Sclaroff, Z.~Lin, X.~Shen, B.~Price, and R.~Mech, ``Minimum
  barrier salient object detection at 80 fps,'' in {\em Proceedings of the IEEE
  international conference on computer vision}, pp.~1404--1412, 2015.

\bibitem{zhou2014time}
F.~Zhou, S.~Bing~Kang, and M.~F. Cohen, ``Time-mapping using space-time
  saliency,'' in {\em Proceedings of the IEEE Conference on Computer Vision and
  Pattern Recognition}, pp.~3358--3365, 2014.

\bibitem{sun2014ranking}
M.~Sun, A.~Farhadi, and S.~Seitz, ``Ranking domain-specific highlights by
  analyzing edited videos,'' in {\em European conference on computer vision},
  pp.~787--802, Springer, 2014.

\bibitem{yao2016highlight}
T.~Yao, T.~Mei, and Y.~Rui, ``Highlight detection with pairwise deep ranking
  for first-person video summarization,'' in {\em Proceedings of the IEEE
  conference on computer vision and pattern recognition}, pp.~982--990, 2016.

\bibitem{yang2015unsupervised}
H.~Yang, B.~Wang, S.~Lin, D.~Wipf, M.~Guo, and B.~Guo, ``Unsupervised
  extraction of video highlights via robust recurrent auto-encoders,'' in {\em
  Proceedings of the IEEE international conference on computer vision},
  pp.~4633--4641, 2015.

\bibitem{liu2015multi}
W.~Liu, T.~Mei, Y.~Zhang, C.~Che, and J.~Luo, ``Multi-task deep visual-semantic
  embedding for video thumbnail selection,'' in {\em Proceedings of the IEEE
  Conference on Computer Vision and Pattern Recognition}, pp.~3707--3715, 2015.

\bibitem{panda2017weakly}
R.~Panda, A.~Das, Z.~Wu, J.~Ernst, and A.~K. Roy-Chowdhury, ``Weakly supervised
  summarization of web videos,'' in {\em Proceedings of the IEEE International
  Conference on Computer Vision}, pp.~3657--3666, 2017.

\bibitem{potapov2014category}
D.~Potapov, M.~Douze, Z.~Harchaoui, and C.~Schmid, ``Category-specific video
  summarization,'' in {\em European conference on computer vision},
  pp.~540--555, Springer, 2014.

\bibitem{xiong2019less}
B.~Xiong, Y.~Kalantidis, D.~Ghadiyaram, and K.~Grauman, ``Less is more:
  Learning highlight detection from video duration,'' {\em arXiv preprint
  arXiv:1903.00859}, 2019.

\bibitem{he2019bi}
J.~He, S.~Zhang, M.~Yang, Y.~Shan, and T.~Huang, ``Bi-directional cascade
  network for perceptual edge detection,'' {\em arXiv preprint
  arXiv:1902.10903}, 2019.

\bibitem{8516362}
Y.~{Liu}, M.~{Cheng}, X.~{Hu}, J.~{Bian}, L.~{Zhang}, X.~{Bai}, and J.~{Tang},
  ``Richer convolutional features for edge detection,'' {\em IEEE Transactions
  on Pattern Analysis and Machine Intelligence}, pp.~1--1, 2018.

\bibitem{ren2016convolutional}
X.~Ren, Y.~Zhou, J.~He, K.~Chen, X.~Yang, and J.~Sun, ``A convolutional neural
  network-based chinese text detection algorithm via text structure modeling,''
  {\em IEEE Transactions on Multimedia}, vol.~19, no.~3, pp.~506--518, 2016.

\bibitem{liao2017textboxes}
M.~Liao, B.~Shi, X.~Bai, X.~Wang, and W.~Liu, ``Textboxes: A fast text detector
  with a single deep neural network,'' in {\em Thirty-First AAAI Conference on
  Artificial Intelligence}, 2017.

\bibitem{bazazian2017improving}
D.~Bazazian, R.~Gomez, A.~Nicolaou, L.~Gomez, D.~Karatzas, and A.~D. Bagdanov,
  ``Improving text proposals for scene images with fully convolutional
  networks,'' {\em arXiv preprint arXiv:1702.05089}, 2017.

\bibitem{zhang2016multi}
Z.~Zhang, C.~Zhang, W.~Shen, C.~Yao, W.~Liu, and X.~Bai, ``Multi-oriented text
  detection with fully convolutional networks,'' in {\em Proceedings of the
  IEEE Conference on Computer Vision and Pattern Recognition}, pp.~4159--4167,
  2016.

\bibitem{yao2016scene}
C.~Yao, X.~Bai, N.~Sang, X.~Zhou, S.~Zhou, and Z.~Cao, ``Scene text detection
  via holistic, multi-channel prediction,'' {\em arXiv preprint
  arXiv:1606.09002}, 2016.

\bibitem{he2016accurate}
T.~He, W.~Huang, Y.~Qiao, and J.~Yao, ``Accurate text localization in natural
  image with cascaded convolutional text network,'' {\em arXiv preprint
  arXiv:1603.09423}, 2016.

\bibitem{lyu2018multi}
P.~Lyu, C.~Yao, W.~Wu, S.~Yan, and X.~Bai, ``Multi-oriented scene text
  detection via corner localization and region segmentation,'' in {\em
  Proceedings of the IEEE Conference on Computer Vision and Pattern
  Recognition}, pp.~7553--7563, 2018.

\bibitem{ma2018arbitrary}
J.~Ma, W.~Shao, H.~Ye, L.~Wang, H.~Wang, Y.~Zheng, and X.~Xue,
  ``Arbitrary-oriented scene text detection via rotation proposals,'' {\em IEEE
  Transactions on Multimedia}, vol.~20, no.~11, pp.~3111--3122, 2018.

\bibitem{wang2019towards}
X.~Wang, Z.~Cai, D.~Gao, and N.~Vasconcelos, ``Towards universal object
  detection by domain attention,'' {\em arXiv preprint arXiv:1904.04402}, 2019.

\bibitem{bilen2017universal}
H.~Bilen and A.~Vedaldi, ``Universal representations: The missing link between
  faces, text, planktons, and cat breeds,'' {\em arXiv preprint
  arXiv:1701.07275}, 2017.

\bibitem{rebuffi2017learning}
S.-A. Rebuffi, H.~Bilen, and A.~Vedaldi, ``Learning multiple visual domains
  with residual adapters,'' in {\em Advances in Neural Information Processing
  Systems}, pp.~506--516, 2017.

\bibitem{rebuffi2018efficient}
S.-A. Rebuffi, H.~Bilen, and A.~Vedaldi, ``Efficient parametrization of
  multi-domain deep neural networks,'' in {\em Proceedings of the IEEE
  Conference on Computer Vision and Pattern Recognition}, pp.~8119--8127, 2018.

\bibitem{chen2018domain}
Y.~Chen, W.~Li, C.~Sakaridis, D.~Dai, and L.~Van~Gool, ``Domain adaptive faster
  r-cnn for object detection in the wild,'' in {\em Proceedings of the IEEE
  Conference on Computer Vision and Pattern Recognition}, pp.~3339--3348, 2018.

\bibitem{saito2018strong}
K.~Saito, Y.~Ushiku, T.~Harada, and K.~Saenko, ``Strong-weak distribution
  alignment for adaptive object detection,'' {\em arXiv preprint
  arXiv:1812.04798}, 2018.

\bibitem{haupmann2019contrastive}
A.~Haupmann, G.~Kang, L.~Jiang, and Y.~Yang, ``Contrastive adaptation network
  for unsupervised domain adaptation,'' 2019.

\bibitem{han2016seq}
W.~Han, P.~Khorrami, T.~L. Paine, P.~Ramachandran, M.~Babaeizadeh, H.~Shi,
  J.~Li, S.~Yan, and T.~S. Huang, ``Seq-nms for video object detection,'' {\em
  arXiv preprint arXiv:1602.08465}, 2016.

\bibitem{feichtenhofer2017detect}
C.~Feichtenhofer, A.~Pinz, and A.~Zisserman, ``Detect to track and track to
  detect,'' in {\em Proceedings of the IEEE International Conference on
  Computer Vision}, pp.~3038--3046, 2017.

\bibitem{kang2016object}
K.~Kang, W.~Ouyang, H.~Li, and X.~Wang, ``Object detection from video tubelets
  with convolutional neural networks,'' in {\em Proceedings of the IEEE
  conference on computer vision and pattern recognition}, pp.~817--825, 2016.

\bibitem{kang2017t}
K.~Kang, H.~Li, J.~Yan, X.~Zeng, B.~Yang, T.~Xiao, C.~Zhang, Z.~Wang, R.~Wang,
  X.~Wang, {\em et~al.}, ``T-cnn: Tubelets with convolutional neural networks
  for object detection from videos,'' {\em IEEE Transactions on Circuits and
  Systems for Video Technology}, vol.~28, no.~10, pp.~2896--2907, 2017.

\bibitem{kang2017object}
K.~Kang, H.~Li, T.~Xiao, W.~Ouyang, J.~Yan, X.~Liu, and X.~Wang, ``Object
  detection in videos with tubelet proposal networks,'' in {\em Proceedings of
  the IEEE Conference on Computer Vision and Pattern Recognition},
  pp.~727--735, 2017.

\bibitem{zhu2017deep}
X.~Zhu, Y.~Xiong, J.~Dai, L.~Yuan, and Y.~Wei, ``Deep feature flow for video
  recognition,'' in {\em Proceedings of the IEEE Conference on Computer Vision
  and Pattern Recognition}, pp.~2349--2358, 2017.

\bibitem{zhu2017flow}
X.~Zhu, Y.~Wang, J.~Dai, L.~Yuan, and Y.~Wei, ``Flow-guided feature aggregation
  for video object detection,'' in {\em Proceedings of the IEEE International
  Conference on Computer Vision}, pp.~408--417, 2017.

\bibitem{wang2015visual}
L.~Wang, W.~Ouyang, X.~Wang, and H.~Lu, ``Visual tracking with fully
  convolutional networks,'' in {\em Proceedings of the IEEE international
  conference on computer vision}, pp.~3119--3127, 2015.

\bibitem{bertasius2018object}
G.~Bertasius, L.~Torresani, and J.~Shi, ``Object detection in video with
  spatiotemporal sampling networks,'' in {\em Proceedings of the European
  Conference on Computer Vision (ECCV)}, pp.~331--346, 2018.

\bibitem{xiao2018video}
F.~Xiao and Y.~Jae~Lee, ``Video object detection with an aligned
  spatial-temporal memory,'' in {\em Proceedings of the European Conference on
  Computer Vision (ECCV)}, pp.~485--501, 2018.

\bibitem{8686124}
P.~{Tang}, C.~{Wang}, X.~{Wang}, W.~{Liu}, W.~{Zeng}, and J.~{Wang}, ``Object
  detection in videos by high quality object linking,'' {\em IEEE Transactions
  on Pattern Analysis and Machine Intelligence}, pp.~1--1, 2019.

\bibitem{engelcke2017vote3deep}
M.~Engelcke, D.~Rao, D.~Z. Wang, C.~H. Tong, and I.~Posner, ``Vote3deep: Fast
  object detection in 3d point clouds using efficient convolutional neural
  networks,'' in {\em 2017 IEEE International Conference on Robotics and
  Automation (ICRA)}, pp.~1355--1361, IEEE, 2017.

\bibitem{qi2017pointnet}
C.~R. Qi, H.~Su, K.~Mo, and L.~J. Guibas, ``Pointnet: Deep learning on point
  sets for 3d classification and segmentation,'' in {\em Proceedings of the
  IEEE Conference on Computer Vision and Pattern Recognition}, pp.~652--660,
  2017.

\bibitem{qi2017pointnet++}
C.~R. Qi, L.~Yi, H.~Su, and L.~J. Guibas, ``Pointnet++: Deep hierarchical
  feature learning on point sets in a metric space,'' in {\em Advances in
  Neural Information Processing Systems}, pp.~5099--5108, 2017.

\bibitem{zhou2018voxelnet}
Y.~Zhou and O.~Tuzel, ``Voxelnet: End-to-end learning for point cloud based 3d
  object detection,'' in {\em Proceedings of the IEEE Conference on Computer
  Vision and Pattern Recognition}, pp.~4490--4499, 2018.

\bibitem{chen2016monocular}
X.~Chen, K.~Kundu, Z.~Zhang, H.~Ma, S.~Fidler, and R.~Urtasun, ``Monocular 3d
  object detection for autonomous driving,'' in {\em Proceedings of the IEEE
  Conference on Computer Vision and Pattern Recognition}, pp.~2147--2156, 2016.

\bibitem{chen2017multi}
X.~Chen, H.~Ma, J.~Wan, B.~Li, and T.~Xia, ``Multi-view 3d object detection
  network for autonomous driving,'' in {\em Proceedings of the IEEE Conference
  on Computer Vision and Pattern Recognition}, pp.~1907--1915, 2017.

\bibitem{sindagi2019mvx}
V.~A. Sindagi, Y.~Zhou, and O.~Tuzel, ``Mvx-net: Multimodal voxelnet for 3d
  object detection,'' {\em arXiv preprint arXiv:1904.01649}, 2019.

\bibitem{cao2017realtime}
Z.~Cao, T.~Simon, S.-E. Wei, and Y.~Sheikh, ``Realtime multi-person 2d pose
  estimation using part affinity fields,'' in {\em Proceedings of the IEEE
  Conference on Computer Vision and Pattern Recognition}, pp.~7291--7299, 2017.

\bibitem{bulat2016human}
A.~Bulat and G.~Tzimiropoulos, ``Human pose estimation via convolutional part
  heatmap regression,'' in {\em European Conference on Computer Vision},
  pp.~717--732, Springer, 2016.

\bibitem{newell2016stacked}
A.~Newell, K.~Yang, and J.~Deng, ``Stacked hourglass networks for human pose
  estimation,'' in {\em European Conference on Computer Vision}, pp.~483--499,
  Springer, 2016.

\bibitem{chen2014articulated}
X.~Chen and A.~L. Yuille, ``Articulated pose estimation by a graphical model
  with image dependent pairwise relations,'' in {\em Advances in neural
  information processing systems}, pp.~1736--1744, 2014.

\bibitem{toshev2014deeppose}
A.~Toshev and C.~Szegedy, ``Deeppose: Human pose estimation via deep neural
  networks,'' in {\em Proceedings of the IEEE conference on computer vision and
  pattern recognition}, pp.~1653--1660, 2014.

\bibitem{fan2015combining}
X.~Fan, K.~Zheng, Y.~Lin, and S.~Wang, ``Combining local appearance and
  holistic view: Dual-source deep neural networks for human pose estimation,''
  in {\em Proceedings of the IEEE conference on computer vision and pattern
  recognition}, pp.~1347--1355, 2015.

\bibitem{ouyang2014multi}
W.~Ouyang, X.~Chu, and X.~Wang, ``Multi-source deep learning for human pose
  estimation,'' in {\em Proceedings of the IEEE Conference on Computer Vision
  and Pattern Recognition}, pp.~2329--2336, 2014.

\bibitem{8611390}
G.~{Rogez}, P.~{Weinzaepfel}, and C.~{Schmid}, ``Lcr-net++: Multi-person 2d and
  3d pose detection in natural images,'' {\em IEEE Transactions on Pattern
  Analysis and Machine Intelligence}, pp.~1--1, 2019.

\bibitem{chen2018cascaded}
Y.~Chen, Z.~Wang, Y.~Peng, Z.~Zhang, G.~Yu, and J.~Sun, ``Cascaded pyramid
  network for multi-person pose estimation,'' in {\em Proceedings of the IEEE
  Conference on Computer Vision and Pattern Recognition}, pp.~7103--7112, 2018.

\bibitem{papandreou2017towards}
G.~Papandreou, T.~Zhu, N.~Kanazawa, A.~Toshev, J.~Tompson, C.~Bregler, and
  K.~Murphy, ``Towards accurate multi-person pose estimation in the wild,'' in
  {\em Proceedings of the IEEE Conference on Computer Vision and Pattern
  Recognition}, pp.~4903--4911, 2017.

\bibitem{xiao2018simple}
B.~Xiao, H.~Wu, and Y.~Wei, ``Simple baselines for human pose estimation and
  tracking,'' in {\em Proceedings of the European Conference on Computer Vision
  (ECCV)}, pp.~466--481, 2018.

\bibitem{wei2016convolutional}
S.-E. Wei, V.~Ramakrishna, T.~Kanade, and Y.~Sheikh, ``Convolutional pose
  machines,'' in {\em Proceedings of the IEEE Conference on Computer Vision and
  Pattern Recognition}, pp.~4724--4732, 2016.

\bibitem{li2019rethinking}
W.~Li, Z.~Wang, B.~Yin, Q.~Peng, Y.~Du, T.~Xiao, G.~Yu, H.~Lu, Y.~Wei, and
  J.~Sun, ``Rethinking on multi-stage networks for human pose estimation,''
  {\em arXiv preprint arXiv:1901.00148}, 2019.

\bibitem{Krause_2013_ICCV_Workshops}
J.~Krause, M.~Stark, J.~Deng, and L.~Fei-Fei, ``3d object representations for
  fine-grained categorization,'' in {\em The IEEE International Conference on
  Computer Vision (ICCV) Workshops}, June 2013.

\bibitem{lin2015bilinear}
T.-Y. Lin, A.~RoyChowdhury, and S.~Maji, ``Bilinear cnn models for fine-grained
  visual recognition,'' in {\em Proceedings of the IEEE international
  conference on computer vision}, pp.~1449--1457, 2015.

\bibitem{he2017fine}
X.~He, Y.~Peng, and J.~Zhao, ``Fine-grained discriminative localization via
  saliency-guided faster r-cnn,'' in {\em Proceedings of the 25th ACM
  international conference on Multimedia}, pp.~627--635, ACM, 2017.

\bibitem{he2018fast}
X.~He, Y.~Peng, and J.~Zhao, ``Fast fine-grained image classification via
  weakly supervised discriminative localization,'' {\em IEEE Transactions on
  Circuits and Systems for Video Technology}, vol.~29, no.~5, pp.~1394--1407,
  2018.

\bibitem{khosla2011novel}
A.~Khosla, N.~Jayadevaprakash, B.~Yao, and F.-F. Li, ``Novel dataset for
  fine-grained image categorization: Stanford dogs,'' in {\em Proc. CVPR
  Workshop on Fine-Grained Visual Categorization (FGVC)}, vol.~2, 2011.

\bibitem{maji2013fine}
S.~Maji, E.~Rahtu, J.~Kannala, M.~Blaschko, and A.~Vedaldi, ``Fine-grained
  visual classification of aircraft,'' {\em arXiv preprint arXiv:1306.5151},
  2013.

\bibitem{zhao2017survey}
B.~Zhao, J.~Feng, X.~Wu, and S.~Yan, ``A survey on deep learning-based
  fine-grained object classification and semantic segmentation,'' {\em
  International Journal of Automation and Computing}, vol.~14, no.~2,
  pp.~119--135, 2017.

\end{thebibliography}
%%\begin{IEEEbiographynophoto}{John Doe}
%%Biography text here.
%%\end{IEEEbiographynophoto}

%\newpage

%%\begin{IEEEbiographynophoto}{Jane Doe}
%%Biography text here.
%%\end{IEEEbiographynophoto}


%\vfill


\end{document}